\documentclass{article}

\PassOptionsToPackage{numbers, compress}{natbib}

\usepackage[final]{neurips_2025}

\usepackage[dvipsnames]{xcolor}
\usepackage[T1]{fontenc}
\usepackage{tikz}
\usepackage{dsfont}
\usepackage{pgfplots}
\usepgfplotslibrary{fillbetween}
\usetikzlibrary{patterns}
\usetikzlibrary{arrows.meta, positioning}

\usepackage{graphicx} 
\usepackage{hyperref} 

\usepackage{xspace}

\usepackage{algorithm}

\usepackage{makecell}
\usepackage{orcidlink}

\usepackage{enumitem}

\usepackage{graphicx}
\usepackage{amsmath, amssymb}

\usepackage{lua-visual-debug}
\usepackage{wrapfig}
\usepackage{mdframed}
\usepackage{hyperref}
\hypersetup{%
	colorlinks=true,           
	allcolors=brown!70!black,   
	pdfstartview=Fit,          
	breaklinks=true,
	pdfauthor={},
	pdftitle={{}}
}

\usepackage{cleveref}
\usepackage{color}

\usepackage{stmaryrd}
\usepackage{amsfonts}
\usepackage{marvosym}
\usepackage{multirow}

\usepackage{bussproofs}
\usepackage{booktabs}
\usepackage{colortbl}

\usepackage{listings}

\usepackage{subcaption}
\usepackage{todonotes}
\usepackage{graphicx}

\usepackage{stmaryrd}

\usepackage{ltl}

\definecolor{gray}{rgb}{0.5,0.5,0.5}
\definecolor{darkgreen}{rgb}{0,0.4,0}
\definecolor{darkblue}{rgb}{0.2, 0.2, 0.6}
\definecolor{niceblue}{rgb}{0.16, 0.32, 0.75}
\definecolor{niceblack}{rgb}{0.0, 0.18, 0.39}
\definecolor{prettyred}{rgb}{1.0, 0.13, 0.32}
\definecolor{prettypink}{rgb}{0.98, 0.38, 0.5}
\definecolor{prettyblue}{rgb}{0.16, 0.32, 0.75}
\definecolor{prettygreen}{rgb}{0.0, 0.5, 0.0}
\definecolor{prettyorange}{rgb}{0.99, 0.53, 0.05}
\definecolor{prettypurple}{rgb}{0.6, 0.4, 0.8}
\definecolor{prettyyellow}{rgb}{0.99, 0.93, 0.0}
\definecolor{airforceblue}{rgb}{0.36, 0.54, 0.66}

\usepackage[many]{tcolorbox}    	
\usepackage{multicol}               

\colorlet{main}{black}

\colorlet{sub}{main!30}

\tcbset{
	sharp corners,
	colback = white,
	before skip = 0.2cm,    
	after skip = 0.5cm      
}                           

\newtcolorbox{boxA}{
	fontupper = \bf,
	boxrule = 1.5pt,
	colframe = black 
}

\newtcolorbox{boxB}{
	fontupper = \bf\color{main}, 
	boxrule = 1.5pt,
	colframe = main,
	rounded corners,
	arc = 5pt   
}

\newtcolorbox{boxC}{
	colback = sub, 
	boxrule = 0pt  
}

\newtcolorbox{boxD}{
	colback = sub, 
	colframe = main, 
	boxrule = 0pt, 
	toprule = 3pt, 
	bottomrule = 3pt 
}

\newtcolorbox{boxE}{
	enhanced, 
	boxrule = 0pt, 
	borderline = {0.75pt}{0pt}{main}, 
	borderline = {0.75pt}{2pt}{sub} 
}

\newtcolorbox{boxF}{
	colback = sub,
	enhanced,
	boxrule = 1.5pt, 
	colframe = white, 
	borderline = {1.5pt}{0pt}{main, dashed} 
}

\newtcolorbox{boxG}{
	enhanced,
	boxrule = 0pt,
	colback = sub,
	borderline west = {1pt}{0pt}{main}, 
	borderline west = {0.75pt}{2pt}{main}, 
	borderline east = {1pt}{0pt}{main}, 
	borderline east = {0.75pt}{2pt}{main}
}

\newtcolorbox{boxH}{
	colback = sub, 
	colframe = main, 
	boxrule = 0pt, 
	leftrule = 6pt 
}

\newtcolorbox{boxI}{
	colback = sub, 
	colframe = main, 
	boxrule = 0pt, 
	toprule = 6pt 
}

\newtcolorbox{boxJ}{
	sharpish corners, 
	colback = sub, 
	colframe = main, 
	boxrule = 0pt, 
	toprule = 4.5pt, 
	enhanced,
	fuzzy shadow = {0pt}{-2pt}{-0.5pt}{0.5pt}{black!35} 
}

\newtcolorbox{boxK}{
	sharpish corners, 
	colframe =prettyblue!50,
	boxrule = 0pt,
	toprule = 4.5pt, 
	enhanced,
	fuzzy shadow = {0pt}{-2pt}{-0.5pt}{0.5pt}{prettyblue!50} 
}

\newtcolorbox{boxL}{
	fontupper = \color{main},
	rounded corners,
	arc = 6pt,
	colback = sub, 
	colframe = main!50, 
	boxrule = 0pt, 
	bottomrule = 4.5pt 
}

\newtcolorbox{boxM}{
	fontupper = \color{white},
	rounded corners,
	arc = 6pt,
	colback = main!80, 
	colframe = main, 
	boxrule = 0pt, 
	bottomrule = 4.5pt,
	enhanced,
	fuzzy shadow = {0pt}{-3pt}{-0.5pt}{0.5pt}{black!35}
}

{}

{}

{}

\definecolor{c1}{HTML}{602273}
\definecolor{c2}{HTML}{F39F5A}
\definecolor{c3}{HTML}{AE445A}
\definecolor{c4}{HTML}{6e4909}

\colorlet{statementColor}{c3}

\newtheorem{theorem}{Theorem}

\usepackage{scalefnt}

\colorlet{command-color}{black!70}
\definecolor{dkcyan}{rgb}{0.1, 0.3, 0.3}
\definecolor{dkgreen}{rgb}{0,0.3,0}
\definecolor{dkblue}{rgb}{0,0.3,1.0}
\colorlet{comment-color}{black!50}

\lstdefinelanguage{code-lang}{
	keywords={def, repeat, return, if, then, else,or},
	keywordstyle=[1]\color{command-color},
	morekeywords=[2]{faultLocalization,instrument, symbolicExecution,SyGuS},
	keywordstyle=[2]\color{dkgreen},
	morekeywords=[3]{iterativeRepair},
	keywordstyle=[3]\color{dkcyan},
	comment=[l][\color{comment-color}]{//},
	literate=%
	{=}{{{\color{command-color}=}}}1
	{|}{{{\color{command-color}|}}}1
	{:}{{{\color{command-color}:}}}1
	{:=}{{{\color{command-color}:=}}}1
	{@}{ }1
}

\lstdefinestyle{code-style}{
	escapeinside={(*}{*)},
	basicstyle=\ttfamily\fontsize{9}{11}\selectfont,
	columns=fullflexible,
	commentstyle=\sffamily\color{black!50!white},
	framexleftmargin=1em,
	framexrightmargin=1ex,
	keepspaces=true,
	mathescape,
	numbers=left,
	numberblanklines=false,
	numbersep=0.5em,
	numberstyle=\relscale{0.65}\color{gray}\ttfamily,
	showstringspaces=true,
	stepnumber=1,
	xleftmargin=1.2em,
}

\lstdefinelanguage{example-lang}{
	keywords={while,do,if, then,else, observe,or},
	keywordstyle=[1]{\color{c3}},
	morekeywords=[2]{string,int,bool},
	keywordstyle=[2]\color{black!60},
	morekeywords=[3]{phase,title,session,decision,print,password,attack,request,username,date,userPassword,info,LOG,credentials,reviewA,reviewB, reviewerAid,reviewerBid,order,notification,withdraw,balance,amount,ErrorLog,TransactionLog},
	keywordstyle=[3]\color{c1},
	morekeywords=[4]{366,1,0,true},
	keywordstyle=[4]\color{c4},
	morestring=[b]",
	stringstyle=\color{c4},
	comment=[l][\color{comment-color}]{//},
	literate=%
	{=}{{{\color{c3}=}}}1
	{!=}{{{\color{c3}!=}}}1
	{>}{{{\color{c3}>}}}1
	{<}{{{\color{c3}<}}}1
	{+}{{{\color{c3}+}}}1
	{-}{{{\color{c3}-}}}1
	{*}{{{\color{c3}*}}}1
	{@}{ }1
}

\lstdefinestyle{example-style}{
	escapeinside={(*}{*)},
	basicstyle=\ttfamily\fontsize{8}{10}\selectfont,
	columns=fullflexible,
	commentstyle=\sffamily\color{black!50!white},
	framexleftmargin=0em,
	framexrightmargin=0ex,
	keepspaces=true,
	mathescape,
	numbers=left,
	numberblanklines=false,
	numbersep=1.0em,
	numberstyle=\relscale{0.65}\color{gray}\ttfamily,
	showstringspaces=true,
	stepnumber=1,
	xleftmargin=1.2em
}

\lstdefinestyle{example-style-large}{
	escapeinside={(*}{*)},
	basicstyle=\ttfamily\fontsize{10}{12}\selectfont,
	columns=fullflexible,
	commentstyle=\sffamily\color{black!50!white},
	framexleftmargin=0em,
	framexrightmargin=0ex,
	keepspaces=true,
	mathescape,
	numbers=left,
	numberblanklines=false,
	numbersep=0.5em,
	numberstyle=\relscale{0.65}\color{gray}\ttfamily,
	showstringspaces=true,
	stepnumber=1
}

\lstnewenvironment{code}[1][]
{\small
	\lstset{
		style=code-style, 
		language=code-lang,
		#1
	}
}
{}

\lstnewenvironment{exampleCode}[1][]
{\small
	\lstset{
		style=example-style, 
		language=example-lang,
		#1
	}
}
{}

\newcommand{\AP}{\mathsf{AP}}

\newcommand{\realnum}{\mathbb{R}}

\newcommand{\integers}{\mathbb{Z}}
\newcommand{\infinite}{{\omega}}
\newcommand{\finite}{*}

\newcommand{\definedas}{\triangleq}

\newcommand{\MDP}{\mathcal{M}}

\newcommand{\MDPtuple}{\tupleof{ \states, \inits, \actions, \trans, \AP, 
\labels }}
\newcommand{\states}{\mathit{S}}
\newcommand{\state}{\mathit{s}}
\newcommand{\labels}{\mathit{L}}
\newcommand{\labelsof}[1]{\mathit{L}(#1)}

\newcommand{\actions}{\mathit{A}}
\newcommand{\action}{\mathit{a}}

\newcommand{\trans}{\mathbf{P}}

\newcommand{\inits}{{\state^0}}

\newcommand{\edgeof}[1]{\xrightarrow{#1}}
\newcommand{\prob}{\mathbb{P}}

\newcommand{\optimal}{\star}

\newcommand{\maxpolicy}{\pi^{\optimal}}
\newcommand{\policy}{\pi}
\newcommand{\nnpolicy}{\mathcal{NN}}
\newcommand{\optnnpolicy}{\nnpolicy^{\optimal}}
\newcommand{\optnnpolicyof}[1]{\optnnpolicy_{#1}}

\newcommand{\traj}{\zeta}
\newcommand{\trajs}{\mathcal{Z}}

\newcommand{\tracevar}{\tau}

\newcommand{\HyperLTL}{{HyperLTL}\xspace}

\newcommand{\LTL}{{LTL}\xspace}
\newcommand{\hyperltl}{{\HyperLTL\xspace}}

\newcommand{\trajfromto}[3]{{#1}_{\bracketof{{#2}:{#3}}}}

\newcommand{\predicatefunc}{\mathit{f}}
\newcommand{\predicatefuncof}[1]{\predicatefunc\big({#1}\big)}
\newcommand{\constant}{\mathit{c}}

\newcommand{\tr}{\mathit{T}}
\newcommand{\Tr}{\mathcal{T}}
\newcommand{\trace}{\mathit{t}}
\newcommand{\traceof}[1]{\mathsf{Tr}(#1)}
\newcommand{\tracesof}[1]{\mathsf{Traces}(#1)}

\newcommand{\OR}{\vee}

\newcommand{\always}{\G}
\newcommand{\until}{\mathbin{\mathcal{U}}}

\newcommand{\tru}{\mathsf{true}}
\newcommand{\fals}{\mathsf{false}}
\newcommand{\G}{\LTLsquare}
\newcommand{\F}{\LTLdiamond}
\newcommand{\X}{\LTLcircle}
\newcommand{\U}{\mathbin{\mathcal{U}}}

\newcommand{\alphabet}{\mathrm{\Sigma}}

\newcommand\modified[1]{#1}

\newcommand{\traceVars}{\mathit{Vars(\varphi)}}
\newcommand{\traceVarsOf}[1]{\mathit{Vars(#1)}}

\newcommand{\var}{p}

\newcommand{\quant}{\mathds{Q}}

\newcommand{\rbvalue}{{\rho}}
\newcommand{\rbvalueof}[1]{\rbvalue\big({#1}\big)}

\newcommand{\maximum}{\textsf{max}}
\newcommand{\minimum}{\textsf{min}}
\newcommand{\maxamong}[1]{\underset{#1}{\textsf{max}}}
\newcommand{\minamong}[1]{\underset{#1}{\textsf{min}}}
\newcommand{\distrubition}{\mathcal{D}}
\newcommand{\distributionof}[1]{\distrubition_{#1}}

\newcommand{\setof}[1]{ \{#1\} }
\newcommand{\skolemized}[1]{\mathbf{Skolem}(#1)}

\newcommand{\skolemfunc}{\mathbf{f}}

\newcommand{\allforall}{\mathds{\quant}}
\newcommand{\allexists}{\mathds{\quant}}
\newcommand{\forallsof}[1]{{\allforall^\forall_{#1}}}
\newcommand{\existsof}[1]{{\allexists^\exists_{#1}}}
\newcommand{\sizeof}[1]{{|#1|}}

\newcommand{\traceassignment}{\Pi}

\newcommand{\zip}{\textsf{zip}}
\newcommand{\zipof}[1]{\textsf{zip}(#1)}

\newcommand{\discount}{\gamma}

\newcommand{\leafindex}{k}

\newcommand{\reward}{\mathit{R}}
\newcommand{\expectedreward}{\mathbb{E}}
\newcommand{\Qvalue}{\mathit{Q}}

\newcommand{\rewardof}[2]{{\reward}({#1}, {#2})}
\newcommand{\expectedrewardof}[2]{{\expectedreward}({#1}, {#2})}

\newcommand{\tupleof}[1]{\langle #1 \rangle}

\newcommand{\traceone}{{\tracevar_1}}
\newcommand{\tracetwo}{{\tracevar_2}}

\newcommand{\ap}[1]{\textsf{#1}} 
\newcommand{\prop}[1]{\textsf{#1}} 

\newcommand{\agonesafe}{\textcolor{blue!70}{\ap{A1}}\xspace}
\newcommand{\agtwosafe}{\textcolor{orange}{\ap{A2}}\xspace}
\newcommand{\Tonesafe}{\psi_{\ap{G1}_{\traceone}}}
\newcommand{\Ttwosafe}{\psi_{\ap{G2}_{\tracetwo}}}
\newcommand{\collide}{\psi_{\prop{CA}_{\traceone, \tracetwo}}}

\newcommand{\goalone}{\textcolor{cyan!80}{\ap{G1}}\xspace}
\newcommand{\goaltwo}{\textcolor{orange!50}{\ap{G2}}\xspace}
\newcommand{\tronesafe}{\tracevar_1}
\newcommand{\trtwosafe}{\tracevar_2}

\newcommand{\pcp}{\prop{PCP}}
\newcommand{\semi}{\prop{SemiMatch}}
\newcommand{\match}{\prop{Match}\xspace}
\newcommand{\extend}{\prop{Extend}}

\newcommand{\Dominos}{D}
\newcommand{\dom}{\mathit{dom}}
\newcommand{\topp}{\mathit{top}}
\newcommand{\bottom}{\mathit{bot}}

\newcommand{\spectral}{\text{SPECTRL}\xspace}

\newcommand{\locmark}{\color{gray}\it} 
\newcommand{\droneone}{\textsf{\small FF}\xspace} 
\newcommand{\dronetwo}{\textsf{\small Med}\xspace}

\newcommand{\motifdist}{\textsf{dist}}
\newcommand{\motifsafe}{\textsf{safe}}
\newcommand{\motifputout}{\textsf{fire}}
\newcommand{\motifsave}{\textsf{save}}
\newcommand{\motifloc}{\textsf{\small Location}}
\newcommand{\motifRescue}{\textsf{Rescue}}
\newcommand{\pathone}{\tau_1}
\newcommand{\pathtwo}{\tau_2}
\newcommand{\movevia}[1]{$\xrightarrow{\text{\tiny{#1}}}$}

\newcommand*\spvertund[1]{\vrule width0pt height0pt depth#1\relax}

\newcommand{\orderedunion}{{\cup}_{{\scriptscriptstyle\leq}}}
\newcommand{\orderedsets}{\bowtie}
\newcommand{\bracketof}[1]{ [ #1 ] }

\newcommand{\hptracevar}{{\tau}}

\newcommand{\nnpolicyprob}[2]{{\nnpolicy}({#1}\mid{#2})}

\newcommand{\method}{\textsc{HypRL}\xspace}
\newcommand{\cav}[1]{{\color{black}#1}}
\newcommand{\cavcomments}[1]{}

\newcommand{\converge}{\overset{\optimal}{\rightarrow}}

\newcommand{\borzoored}[1]{\textcolor{black}{#1}}

\newcommand{\rebuttal}[1]{\textcolor{black}{#1}}

\newcommand{\camera}[1]{\textcolor{black}{#1}}

\newcommand{\familyofsamples}{\mathcal{S}}
\newcommand{\imageof}[1]{\mathit{Img}({#1})}
\usepackage[utf8]{inputenc} 
\usepackage[T1]{fontenc}    
\usepackage{hyperref}       
\usepackage{url}            
\usepackage{booktabs}       
\usepackage{amsfonts}       
\usepackage{nicefrac}       
\usepackage{microtype}      

\title{\textsc{HypRL}: Reinforcement Learning of Control Policies for Hyperproperties}

\author{%
  Tzu-Han Hsu\thanks{Equal contribution.} \quad
  Arshia Rafieioskouei\footnotemark[1] \quad
  Borzoo Bonakdarpour \\
  Michigan State University \\
  \texttt{\{tzuhan, rafieios, borzoo\}@msu.edu}
}

\begin{document}

\maketitle

\vspace{-7mm}

\begin{abstract}
	{Reward shaping} in multi-agent reinforcement learning (MARL) for complex 
	tasks remains a significant challenge. 
	Existing approaches often fail to find optimal solutions or cannot efficiently handle such tasks.
	We propose \method, a specification-guided reinforcement learning 
	framework that learns 
	control policies w.r.t. {\em hyperproperties} expressed in \HyperLTL.
	Hyperproperties constitute a powerful formalism for specifying 
	objectives and constraints over sets of execution traces across agents.
	To learn policies that maximize the satisfaction of a \HyperLTL formula $\varphi$, we apply 
	Skolemization to manage quantifier alternations and define quantitative robustness functions to 
	shape rewards over execution traces of a Markov decision process with unknown transitions.
	A suitable RL algorithm is then used to learn policies that collectively maximize the expected 
	reward and, consequently, increase the probability of satisfying $\varphi$.
	We evaluate \method on {a} diverse {set of} benchmarks, including 
	safety-aware planning, Deep Sea Treasure, and the Post Correspondence Problem.
	We also compare with specification-driven baselines to demonstrate the effectiveness and efficiency of \method.
\end{abstract}


\vspace{-3mm}
\section{Introduction} \label{sec:intro}
\begin{figure}[htb!]
	\begin{subfigure}[c]{.33 \columnwidth}
		$\diamond$~($O_1$): \droneone extinguishes all fires.\\[0.3em]
		$\diamond$~($O_2$): \dronetwo rescues all victims.\\[0.3em]
		$\diamond$~($C_1$): \droneone and \dronetwo always stay within 2-cells of each other.\\[0.3em]
		$\diamond$~($C_2$): \dronetwo must not passing any fire zone at all time.
	\end{subfigure}	
	\hfill
	\begin{subfigure}[c]{.2 \columnwidth}
		\centering
		\includegraphics[scale=.18]{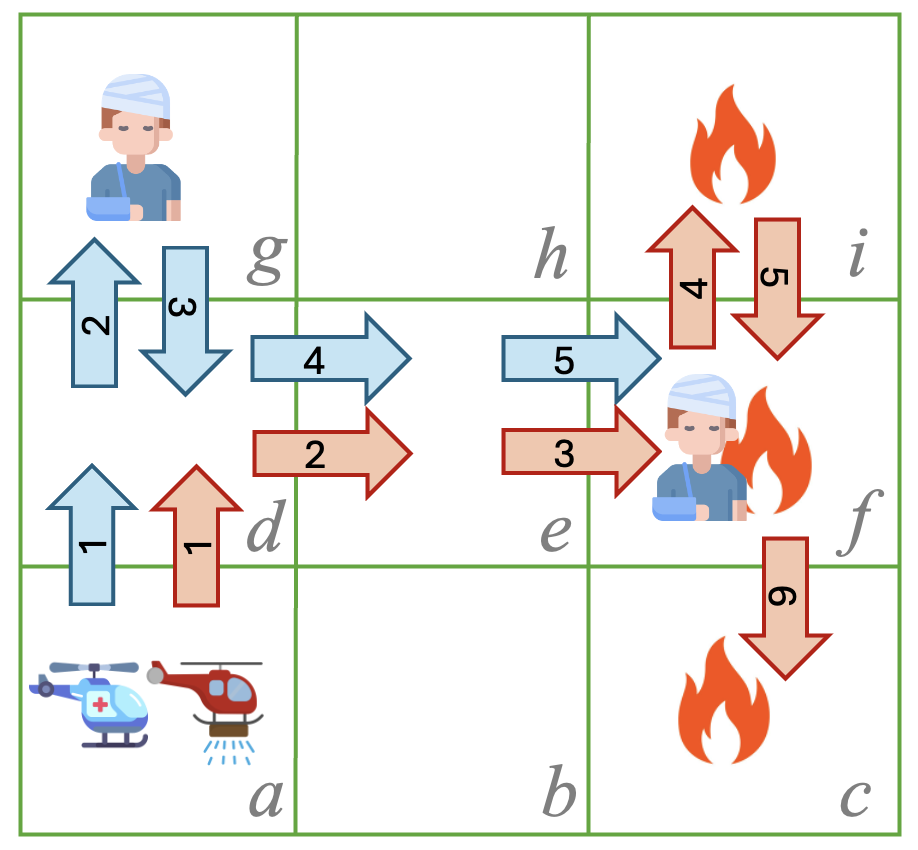}
		\caption{Optimal policies.}\label{fig:motif-optimal}

	\end{subfigure}
	\hfill
	\begin{subfigure}[c]{.22 \columnwidth}
		\centering
		\includegraphics[scale=.18]{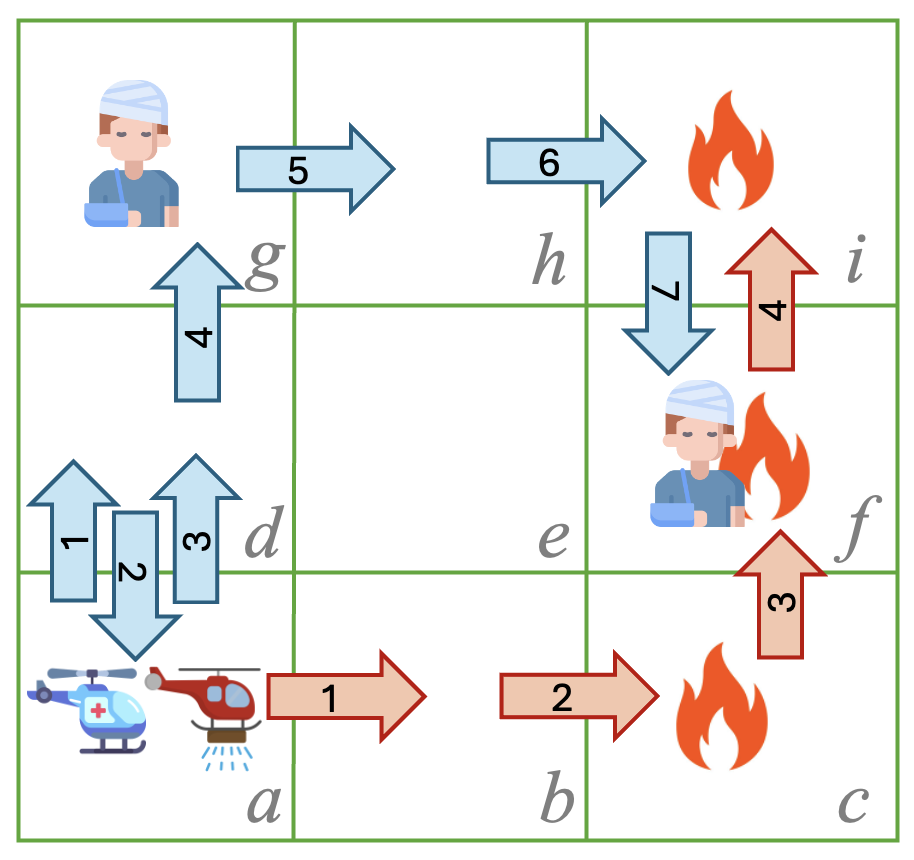}
		\caption{\dronetwo paces in $a-d$.}\label{fig:motif-repeat}

	\end{subfigure}
	\hfill
	\begin{subfigure}[c]{.2 \columnwidth}
		\centering
		\includegraphics[scale=.18]{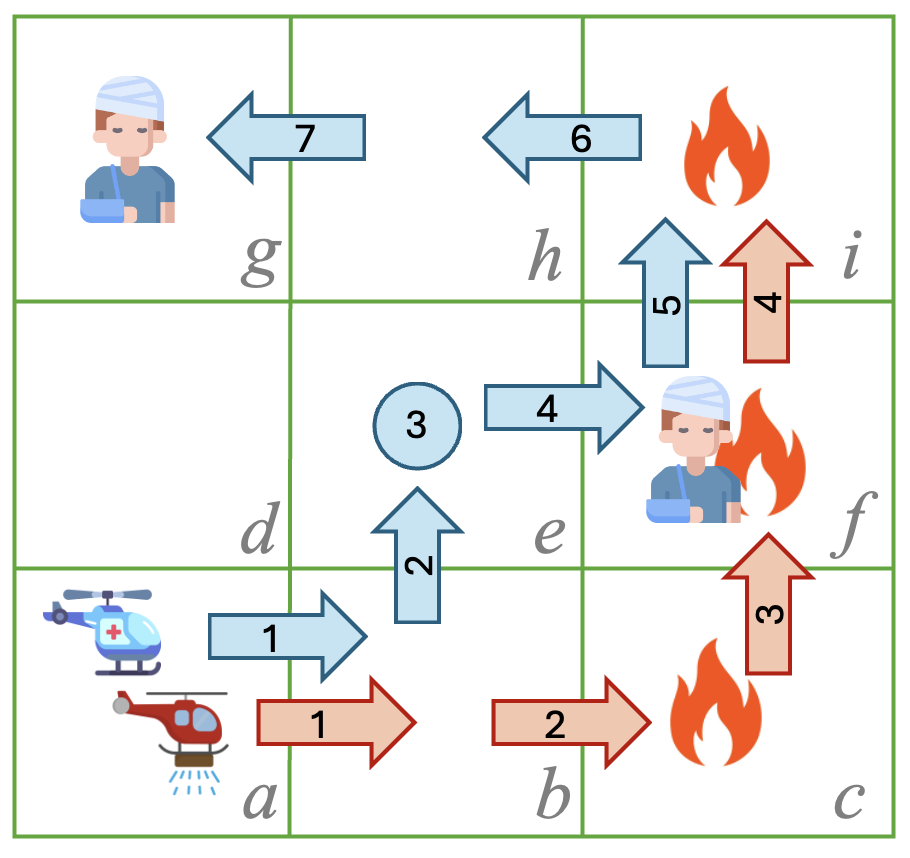}
		\caption{\dronetwo pauses in $e$.}\label{fig:motif-delay}

	\end{subfigure}

\caption{A wildfire scenario with two objectives ($O_1$, $O_2$) and two relational constraints ($C_1$, $C_2$).}
\label{fig:motivation}
\vspace{-5mm}
\end{figure}

Designing reward functions that accurately capture desirable behaviors in 
multi-agent reinforcement learning (MARL) {remains a notorious stumbling block}. 
%
Consider a wildfire scenario in a $3 \times 3$ grid-world with cells labeled $\{{\locmark a}, 
{\locmark b}, \ldots, {\locmark i}\}$ (see \Cref{fig:motivation}). 
%
Locations $\{{\locmark i}, {\locmark f}, {\locmark c}\}$ are on fire, and two victims are located at $\{{\locmark f}, {\locmark g}\}$.
Now, two autonomous agents are deployed from $\{{\locmark a}\}$ with two objectives;
$\boldsymbol{O_1}$: the firefighter agent (\droneone) must extinguish all fire zones, 
and $\boldsymbol{O_2}$: the medical agent (\dronetwo) aims to rescue all victims. 
The agents also have to satisfy two constraints; $\boldsymbol{C_1}$: they must 
always remain within a 2-cell communication range, and $\boldsymbol{C_2}$: \dronetwo cannot 
pass any fire zone before \droneone extinguished the fire in that zone.  
The goal is to learn optimal policies for \droneone and \dronetwo that maximize 
the probability of satisfying all above requirements, as shown by the agent paths 
in~\Cref{fig:motif-optimal}.

Now, suppose we use an existing RL approach by assigning the following rewards; 
extinguish fire: $+50$, rescue a victim: $+10$, 
agent out of range: $-100$, and \dronetwo in fire 
zone: $-100$. 
These approaches would guide \droneone to complete $O_1$ optimally by path 
{\locmark a} \movevia{R} {\locmark b} \movevia{R} {\locmark c} \movevia{U} {\locmark f} 
\movevia{U} {\locmark i}, but due to $C_1$, it forces \dronetwo to delay 
the rescue of the victim in $\{{\locmark g}\}$ with redundant moves: {\locmark a} \movevia{U} {\locmark d} \movevia{D} {\locmark a}  \movevia{U} {\locmark 
d} \movevia{U} {\locmark g}$\cdots$ (see~\Cref{fig:motif-repeat}).  
Furthermore, the victim in $\{{\locmark f}\}$ is trapped in fire, so
%
satisfying $O_2$ depends on the progress of $O_1$, while respecting $C_1$ (safety) 
and $C_2$ (dependency).
Such complex requirements make reward design in MARL particularly challenging, as policies must 
account for both individual objectives and relational constraints between agents with dependencies.
{In fact, existing RL approaches often fail to find optimal policies because agent paths are implicitly \emph{universally} quantified, which is overly restrictive and limits the ability of RL to explore more optimal solutions.
That is, 
the set of requirements is of the form $\forall \tau.\forall \tau'.\psi$, where $\tau$ 
and $\tau'$ are agents' paths and $\psi$ specifies constrains and objectives.}

Temporal logics~\cite{p77} have shown to be both expressive and effective to address objectives 
and constraints in single-agent 
settings~\cite{li2017reinforcement,jothimurugan2021compositional,aksaray2016q,brafman2018ltlf,de2019foundations,hasanbeig2018logically,hasanbeig2019reinforcement,kuo2020encoding,xu2019transfer,zun2019modular}.
However, these methods cannot handle our motivating example as it involves multiple agents. 
Recent works have been extended to 
MARL~\cite{liu2024guidingmultiagentmultitaskreinforcement,leon2020extendedmarkovgameslearn,hammond2021multiagentreinforcementlearningtemporal,elsayedaly2022logicbasedrewardshapingmultiagent},
 but these approaches either fail to capture relational dependencies between agents or can handle 
only a subset of temporal properties, such as \emph{co-safty}~\cite{KupfermanV99,thomas1988safety}.
%
{Moreover, specifications in temporal logics that formalize the behavior of individual 
traces (e.g., \LTL) are again inherently universally quantified and, hence, too restrictive.}

In this paper, we propose a specification-guided RL framework that leverages the 
{powerful framework of {\em hyperproperties}~\cite{cs10} and, in particular, the temporal 
logic} \HyperLTL~\cite{cfkmrs14}, as its formal foundation.
Hyperproperties characterize requirements over sets of execution traces, allowing the specification 
of relational and dependent behaviors that involve multiple agents or joint system executions, 
rather {than individual behaviors of agents}.
This expressiveness is particularly well-suited for capturing relational, quantified, and multi-agent 
objectives.
%
Building on this foundation, we propose \underline{\bf Hyp}erproperties for \underline{\bf 
R}einforcement \underline{\bf L}earning (\method), a framework that learns a collection of control 
policies which maximizes the probability of satisfying a hyperproperty expressed as a \HyperLTL 
formula.
{In our motivating example, \HyperLTL allows a specification of the form $\forall\tau. 
\exists\tau'.\psi$, which enables \method to search for more optimal policies and identify the one 
in~\Cref{fig:motif-optimal}.
%
}
{The main challenge here is to deal with \HyperLTL formulas with quantifier alternation, 
which has not been studied prior to \method.}
%
Our work represents a novel step towards the automated synthesis of reward mechanisms that 
capture complex objectives and constraints in multi-agent setting. 


\begin{tcolorbox}[width=\textwidth,
                  colback=cyan!10,
                  colframe=cyan!10,
                  boxsep=3pt,
                  left=0pt,
                  right=0pt,
                  top=1pt]
    \paragraph{Contributions:}  
    \begin{enumerate}
        \item We formulate the control policies synthesis problem for a multi-agent system as a 
        learning problem, where the goal is to maximize the probability of satisfying a \HyperLTL 
        formula that expresses a set of objectives and constraints (\Cref{sec:problem-statement}).
        \item We address the challenge of reasoning about quantifier alternation in a \HyperLTL 
        formula due to expressing dependencies among agents by Skolemization 
        (\Cref{sec:skolemization}).
		\item We introduce quantitative semantics for \HyperLTL that compute robustness values as 
		rewards, providing a solution of reward shaping for optimizing hyperproperty satisfaction 
		(\Cref{sec:quantsemantics}). 
		Finally, we use off-the-shelf RL algorithms to learn an optimal collection of policies (\Cref{sec:step3}).
        \item We evaluate \method on a diverse set of learning tasks, including safety-aware multi-objective planning, the Post Correspondence Problem, the famous Deep Sea Treasure benchmark, and the wildfire scenario. 
		Our results show that \method is more efficient and effective in handling complex requirements compared to selected baselines (\Cref{sec:Experiments}).
    \end{enumerate}
\end{tcolorbox}
\vspace{-6mm}

\section{Related Work} 
\label{sec:related}

\vspace{-2mm}
\paragraph{Logic-based Single-agent RL.}
\textsc{DiRL}~\cite{jothimurugan2021compositional} is a framework that synthesizes an optimal policy which the specification is expressed in the language \spectral~\cite{jothimurugan2019composable}.
However, \spectral does not capture the full expressiveness of temporal logics as it only supports a 
fragment of \LTL.
In~\cite{li2017reinforcement}, the authors introduce {\em truncated} \LTL for quantitative reasoning over \LTL formulas, but it is limited to single-agent task specifications (similar reasonings are also presented in~\cite{xrwks22,aksaray2016q,brafman2018ltlf}). 
In~\cite{kuo2020encoding}, \LTL formulas are encoded by replacing operators and predicates with recurrent neural networks connected according to the parse tree, forming a structure that captures the formula's semantics. 
In~\cite{de2019foundations,hasanbeig2018logically,xu2019transfer,zun2019modular,NEURIPS2024_d4857f72}, the authors compose an MDP with a  Limit Deterministic
Büchi Automaton (LDBA), which represents the \LTL specification, and solve the compositional model as a control problem.

\paragraph{Logic-based Multi-Agent Reinforcement Learning.}
In~\cite{jothimurugan2022specification}, 
the authors extend the \spectral language~\cite{jothimurugan2019composable} to non-cooperative multi-agent systems 
and train joint policies that form a Nash Equilibrium, which requires relational reasoning.
%
However, their algorithm follows an ``enumerate-and-verify'' approach by exhaustively searching RL policies to identify one that has higher probability to form such equilibrium, which leads to substantial learning overhead as the number of agents increases.
Moreover, the extended language in~\cite{jothimurugan2022specification} cannot handle arbitrary HyperLTL formulas.
%
In~\cite{elsayedaly2022logicbasedrewardshapingmultiagent}, the authors propose a reward shaping method by composing an MDP with a single centralized LDBA derived from a global \LTL specification. 
However, this approach cannot capture inter-agent dependencies, and the product construction of LDBA may suffer from state-space explosion.
More recent efforts in~\cite{liu2024guidingmultiagentmultitaskreinforcement,leon2020extendedmarkovgameslearn,hammond2021multiagentreinforcementlearningtemporal} 
extend quantitative reasoning to formulas with \LTL operators in a multi-agent setting. 
However, in contrast to \method, they only cover {\em co-safety} specifications and do not support the full logic.


\vspace{-3mm}
\paragraph{Shield Synthesis.} Shield synthesis for RL is a technique that asks an agent to 
propose an action in each learning step, 
and a {\em shield} (i.e., a safety guard) evaluates whether such action is safe~\cite{alshiekh2018safe,konighofer2020shield,konighofer2023online}. 
In~\cite{melcer2022shield}, the authors apply shield synthesis in a decentralized multi-agent setting, 
where the learning targets are specified by deterministic finite automata.
However, the specifications are limited to universal (i.e., a $\forall^*. \psi$ formula)
and cannot handle properties such as planning tasks that involve dependencies (which is a $\forall\exists$ hyperproperty).
Furthermore, the authors in~\cite{elsayed2021safe} proposed {\em factored} shielding, which 
can learn multiple policies by a factorization of joint state space 
(i.e., decomposing one shield into multiple sub-shields). 
However, the main contribution is the improvement on RL scalability,
but the limitation on universal-only properties remains.

\vspace{-2mm}
\section{Preliminaries}
\label{sec:problem}

\vspace{-3mm}
\paragraph{Markov Decision Processes (MDP).} 
\label{sec:mdp}

%
\begin{wrapfigure}{r}{0.3\textwidth}
	\centering
	 \vspace{-3mm} 
	\resizebox{0.35\textwidth}{!}{%
		\begin{tikzpicture}[->, auto, thick,node distance=3cm]
			\tikzstyle{state}=[rectangle,draw=black, text=black,minimum size=0.5cm,inner sep=0.1cm]
			\tikzstyle{action}=[circle, fill=black, draw=black,  minimum size=0.2cm,inner sep=0]

			\node[state]    (a)             {$\langle 2,0 \rangle$};
			
			\node[action]    (a1)          [yshift=2cm,xshift=1.5cm,below of=a]      {};
			\node[action]    (a0)            [left of=a1]    {};
			
			\node[state] (b) 				[yshift=2cm,below of=a0]	{$\langle 2,1 \rangle$};
			\node[state](d)  [right of=b]	{$\langle 1,0 \rangle$};

			\node[action](b3) [below of = b, yshift=2cm, xshift = -1.5cm] {};
			
			\node [action ](b0) [right of = b3, xshift=-1.5cm]{};
			
			\node [action] (d0) [below of =d, yshift=2cm]{};
			
			\node[action] (d1) 				[right of=d0, xshift=-2cm]	{};

			\node[state] (e) 				[yshift=2cm,below of=b3]	{$\langle 1,1 \rangle$};
			\node[state] (c) 				[right of=e]	{$\langle 2,2 \rangle$};
			
			\node[state] (g)  [yshift=2cm,below of=d1]	{$\langle 0,0 \rangle$};

            \node [action](g0) [right of = g, xshift=-2cm]{};

			\node [action](e3) [below of = e, yshift=2cm]{};
                \node [action](e0) [right of = e3, xshift = -1.5cm]{};
                
                \node [action](c1) [below of = c, yshift=2cm]{};

                \node[state] (h) 				[yshift=2cm,below of=e3]	{$\langle 0,1 \rangle$};
			\node[state] (f) 				[yshift=2cm,below of=c1]	{$\langle 1,2 \rangle$};

                \node [action](h1) [below of = h, yshift=2cm]{};
                \node [action](f3) [below of = f, yshift=2cm]{};

                \node[state] (i) 				[yshift=3cm,below of=h1, xshift=1.5cm]	{$\langle 0,2 \rangle$};

			\path
			(a)
			edge [] node[swap] {R} (a0)
			edge [] node {U} (a1)

                (a0)
			edge [] node[swap] {1} (b)
			
			(a1)
			edge [] node[swap] {0.5} (b)
			edge [] node {0.5} (d)

                (b)
			edge [] node[swap] {L} (b3)
			edge [] node {R} (b0)
			
			(d)
			edge [] node[swap] {R} (d0)
			edge [] node {U} (d1)

                (b3)
			edge [] node[swap] {1} (e)

            (b0)
			edge [] node[swap] {0.5} (e)
			edge [] node {0.5} (c)

            (e)
			edge [] node[swap] {L} (e3)
			edge [] node {R} (e0)

            (c)
			edge [] node[swap] {U} (c1)

            (e3)
			edge [] node[swap] {1} (h)

            (e0)
			edge [] node[swap] {1} (f)

            (c1)
			edge [] node[] {1} (f)

            (h)
			edge [] node[swap] {U} (h1)

            (f)
			edge [] node[swap] {L} (f3)

            (h1)
			edge [] node[swap] {1} (i)

            (f3)
			edge [] node[swap] {1} (i)

            (d0)
			edge[out=180, in=180]  node {1} (d)

            (d1)
			edge [] node[swap] {1} (g)

            (g)
			edge [] node[swap] {R} (g0)

            (g0)
			edge[out=0, in=0]  node {1} (d);

                \node[left of=a, xshift=2.2cm] {$a$};
                \node[left of=b, xshift=2.2cm] {$b$};
                \node[left of=d, xshift=2.2cm] {$d$};
                \node[left of=c, xshift=2.2cm] {$c$};
                \node[left of=e, xshift=2.2cm] {$e$};
                \node[left of=g, xshift=2.2cm] {$g$};
                 \node[left of=h, xshift=2.2cm] {$h$};
                  \node[left of=f, xshift=2.2cm] {$f$};
                  \node[left of=i, xshift=2.3cm, yshift=0.2cm] {$i$};			
		\end{tikzpicture}
	}
	\caption{The MDP of The grid-world in~\Cref{fig:motivation}.}
	\vspace{-5mm}
	\label{fig:mdpexmp}
\end{wrapfigure} 
We model our RL problem for hyperproperties using MDP
$\MDP = \MDPtuple$, 
where $\states$ is a finite set of {\em states}, $s^0 \in \states$ is the {\em initial state}, 
and $\actions$ is a finite set of {\em actions}.  
The {\em transition probability function} 
$\trans: \states \times \actions \times \states \rightarrow [0,1]$ assigns a probability to each 
transition, ensuring that for any state $\state$ and action $\action$, the sum of outgoing 
probabilities satisfies $\sum_{\state’ \in \states} \trans(\state, \action, \state’) = 1$.
Additionally, $\AP$ denotes a finite set of {\em atomic propositions}, and $\labels: \states \rightarrow 
2^\AP$ is the {\em labeling} function. 
For example,~\Cref{fig:mdpexmp} represents the MDP of~\Cref{fig:motivation}, where each state 
can be represented by $\langle x,y\rangle$ with $x,y \in\lbrace 0,1,2\rbrace $, labeled by $\lbrace 
a,b,...,i\rbrace$, and actions $\actions = \lbrace U,D,L,R\rbrace$ (i.e., four 
moving directions).
A {\em path} in $\MDP$ is a sequence $\traj = \state_0 \edgeof{\action_0} \state_1 
\edgeof{\action_1} \state_2 \edgeof{\action_2} \cdots$, where $\state_i \in \states$ and $\action_i \in 
\actions$ for all $i \geq 0$. 
A {\em sub-path} $\traj_{\bracketof{\ell:k}}$ is a segment $\state_\ell \xrightarrow{\action_\ell} \cdots \xrightarrow{\action_{k-1}} \state_k$, for $0 \leq \ell < k < \sizeof{\traj}$. 
We write $\trajs^\finite$ and $\trajs^\infinite$ to denote the sets of all finite and infinite paths, respectively.
A (deterministic) \emph{policy} 
$ \policy: \trajs^* \rightarrow \actions $
maps each finite path to a (fixed) action $\action \in \actions$.
The {\em trace} $\trace$ of a path $\traj$ is the sequence of labels 
$\traceof{\traj} = \trace(0)\trace(1)\trace(2)\cdots$, where $\trace(i) = 
\labels(\state_i)$ for all $i \geq 0$. 
Slightly abusing notation, we use $\tracesof{\trajs^\finite}$ and $\tracesof{\trajs^\infinite}$ to denote the sets of all finite and infinite traces, respectively.

\vspace{-3mm}
\paragraph{Finite Semantics for HyperLTL.}\label{sec:hyperltl}
The syntax of \HyperLTL~\cite{cfkmrs14} is given by the following grammar:
\[
\varphi ::= \exists \tracevar . \varphi \mid \forall \tracevar. \varphi \mid \psi \quad
~~~~~~~~\psi ::= \var_\tracevar \mid \lnot \psi \mid \psi \OR \psi \mid \X \psi \mid \psi \until \psi,
\]
where $\var \in \AP$ is an atomic proposition, $\tracevar$ is a {\em trace variable}, and the second 
rule (for unquantified formulas) produces LTL formulas. The Boolean connectives $\lnot$ and 
$\OR$ have their standard meanings, while $\X$ and $\until$ denote the ``next'' and ``until’’ 
temporal operators, respectively.
 Other Boolean and temporal operators are derived as syntactic sugar: $\tru \triangleq 
 \var_\tracevar  \vee  \neg  \var_\tracevar$, $\fals \triangleq \neg \tru$, $\psi_1 \rightarrow \psi_2 
 \triangleq \neg \psi_1 \vee  \psi_2$, $\F\psi \triangleq \tru\,\until \psi$, and $\always\psi \triangleq 
 \neg\F\neg\psi$, where `$\F$' and `$\always$' are the \emph{eventually} and {\em always} 
 temporal operators.
In a quantified formula, $\exists \tracevar$ means ``along some trace $\tracevar$'', and $\forall \tracevar$ means ``along all traces $\tracevar$''. 
We write $\traceVars$ for the set of trace variables in a formula $\varphi$. 
A formula is {\em closed} if all $\tracevar \in \traceVars$ are quantified, with no variable quantified 
twice. 
Since RL algorithms operate on finite samples, we adopt the {\em finite semantics} of 
\HyperLTL~\cite{bsb17} over finite trace assignments, where a partial mapping $\Pi: \traceVars 
\rightharpoonup (2^\AP)^\finite$ assigns each $\tracevar \in \traceVars$ to a finite trace. 
Given $\Pi$, a trace variable $\tracevar$, and a finite trace $\trace \in (2^\AP)^\finite$, we write 
$\Pi[\tracevar \rightarrow \trace]$ for the assignment identical to $\Pi$, except that $\tracevar$ is 
mapped to $\trace$.
We denote by $\Pi_\emptyset$ the empty trace assignment. 
For $\trace \in \Pi$, we refer to traces in the image of $\Pi$. 
Throughout this paper, we abbreviate tuples $\tupleof{x_1, \ldots, x_n}$ as $\tupleof{x_i}_{i \in {1, \ldots, n}}$.
An {\em interpretation} of a \HyperLTL formula $\varphi = 
\quant_1 \tracevar_1. \ldots \quant_n \tracevar_n.~ \psi$, denoted by $\Tr 
=\tupleof{\tr_{\tracevar_i}}_{i\in\setof{1,\ldots,\sizeof{\traceVarsOf{\varphi}}}}$,
 is a tuple of sets of traces,
where we have one set $\tr_{\tracevar_i}$ per trace variable $\tracevar_i$, denoting the set of traces that can be assigned to $\tracevar_i$.
Let $\distributionof{\policy_i}$ be the distribution over a set of paths 
induced by a policy $\policy_i$, and we write $\trajs_{\tracevar_i} \sim 
\distributionof{\policy_i}$ to denote a set of paths $\trajs_{\tracevar_i}$ 
sampled from $\distributionof{\policy_i}$, such that $\policy_i$ is 
the policy 
associated with the trace variable $\tracevar_i$, for each 
${i\in\setof{1,\ldots,\sizeof{\traceVarsOf{\varphi}}}}$  (each trace variable 
ranges 
over the possible behaviors of an agent).
We also define a \emph{family} of sampled sets $\familyofsamples = 
\tupleof{\trajs_{\tracevar_i} \sim 
\distributionof{\policy_i}}_{i\in\setof{1,\ldots,\sizeof{\traceVarsOf{\varphi}}}}$. 
That is, for each $\tracevar_i$ in $\traceVarsOf{\varphi}$, $\tr_{\tracevar_i} = 
\tracesof{\trajs_{\tracevar_i} \sim \distributionof{\policy_i}}$ is the set of 
traces that $\tracevar_i$ can range over, which comes from the sampled 
paths from the associated policy $\policy_i$.
Abusing notation, we write $\Tr = \tracesof{\familyofsamples}$ as the tuple of sets of sampled traces. 
%
%
%
%
The satisfaction relation $\models$ maps a formula $\varphi$ to a model $(\Tr, \Pi, i)$, where $i \in \integers_{\geq 0}$ indicates the current evaluation position.
Formally: 
%
	\[
	\begin{array}{ll@{\hspace{1.5em}}l@{\hspace{1.5em}}l}
		\renewcommand{\arraystretch}{1.5}
		(\Tr, \Pi,0) & \models \exists \tracevar.\ \modified{\psi} & \text{iff} & \text{there is a  } t \in 
		\tr_{\tracevar}, \text{ such that } (\Tr,\Pi[\tracevar \rightarrow t],0) \models \psi\\
		(\Tr, \Pi, 0) & \models \forall \tracevar.\ \modified{\psi} & \text{iff} & \text{for all } t \in 
		\tr_{\tracevar}, \text{ such that } (\Tr,\Pi[\tracevar\rightarrow t],0) \models \psi\\
		(\Tr, \Pi, i) & \models \var_\tracevar & \text{iff} & \var \in \Pi(\tracevar)(i)\\
		(\Tr, \Pi,i) &\models \neg \psi & \text{iff} & (\Tr, \Pi, i) \not\models \psi\\ 
		%
		(\Tr, \Pi,i) &\models \psi_1 \OR \psi_2 & \text{iff} & (\Tr, \Pi,i) \models 
		\psi_1\text{ or }	(\Tr, \Pi,i) \models \psi_2\\
		%
		%
		(\Tr, \Pi,i) &\models \X \psi & \mbox{iff} & (\Tr,\Pi,i+1)\models\psi
		\text{ and } \text{for all } \trace \in \Pi. |t| \geq i+1\\
		(\Tr, \Pi, i) &\models \psi_1 \until \psi_2 & \text{iff} &  \text{there exists } j \geq i \text{ with } j < 
		\min_{\trace \in \Pi} |\trace|, \text{such that } (\Tr,\Pi, j) \models \psi_2  \\ 
		&&& \text{ and for all } k \in [i, j), (\Tr,\Pi, k)\models \psi_1.
	\end{array}
	\]
%
We say that an interpretation $\Tr$ satisfies a \HyperLTL formula $\varphi$, written as $\Tr \models 
\varphi$, if $(\Tr, \Pi_\emptyset, 0) \models \varphi$. 
Likewise, a family of samples $\familyofsamples$ (induced by each 
$\policy_{\tau_i}$ associated with each $\tracevar_i \in \traceVars$), satisfies a 
sentence $\varphi$ if $\tupleof{\tracesof{\trajs_{\tracevar_i} \sim 
\distributionof{\policy_i}}}_{{i\in\setof{1,\ldots,\sizeof{\traceVarsOf{\varphi}}}}} \models 
\varphi$. 
%

{
\vspace{-3mm}
\paragraph{Example.}
The following \HyperLTL formula captures the objectives and constraints of our motivating example 
described in~\Cref{fig:motivation}, where $\pathone$ 
is the path for $\droneone$ and $\pathtwo$ is the path for 
$\dronetwo$:
%
\vspace{-4mm}
\begin{center}
	\small
	$
	\textit{Specification}:
	~\varphi_{\textsf{Rescue}}  \triangleq \forall \pathone. \exists \pathtwo.
	(\psi_{\motifputout} \land \psi_{\motifsave} \land \psi_{\motifdist}  \land \psi_{\motifsafe})
	$
		\begin{align*}
			O_1&: 
			~\psi_{\motifputout}  \definedas
			\F ({\locmark i}_{\pathone})  \land 
			\F ({\locmark f}_{\pathone}) \land 
			\F ({\locmark c}_{\pathone})
			~~~
			&C_1&: 
			~\psi_{\motifdist}  \definedas
			\G (| \motifloc_{\pathone} - \motifloc_{\pathtwo} | < 3)
			\\
			O_2&:
			~\psi_{\motifsave}  \definedas 
			\F ({\locmark g}_{\pathtwo})
			\land
			\F ({\locmark f}_{\pathtwo})
			&C_2&:
			~\psi_{\motifsafe}  \definedas 
			(\neg {\locmark i}_{\pathtwo} \U {\locmark i}_{\pathone}) \land
			(\neg {\locmark f}_{\pathtwo} \U {\locmark f}_{\pathone}) \land
			(\neg {\locmark c}_{\pathtwo} \U {\locmark c}_{\pathone})
		\end{align*}
\end{center} 
%
For instance, the dependency constraint $C_2$ for ``\dronetwo cannot enter any fire zone until 
\droneone extinguished the fire in that zone'' is expressed using the conjunction of temporal 
\emph{until} operators.
%
%
Notice that $\varphi_\motifRescue$ features $\forall\exists$ quantifier alternation, which 
increases the complexity of reasoning about hyperproperties~\cite{bf18}.
%
We emphasize that most RL approaches assume purely universal forms (i.e., $\forall^*$), which 
cannot capture agent dependencies and often yield sub-optimal solutions. 
For instance, if \droneone ignores that \dronetwo must wait until the fire is extinguished to rescue a victim, \droneone may follow its own optimal path that causes unnecessary delay for \dronetwo (see~\Cref{fig:motif-delay}).  
}

\vspace{-2mm}
\section{Problem Statement}
\label{sec:problem-statement}
\vspace{-2mm}
%
%
%
Let us use $\star$ to denote optimality (e.g., $\maxpolicy$ denotes an optimal policy). 
The following optimization problem formulates policy synthesis as a learning problem.
%
\begin{boxK}
	Given an MDP $\MDP$ with unknown transitions
	and a \HyperLTL formula $\varphi$ of the form
	$
	\quant_1 \tracevar_1. \ldots 
	\quant_n \tracevar_n.~ 
	\psi
	$,
	our goal is to identify a tuple of $n$ policies
	$ \tupleof{\maxpolicy_1, \ldots, \maxpolicy_n}$, such that:
	\begin{align*}
		\tupleof{\maxpolicy_i}_{i \in 
			\{1, \ldots,n\}}
		\in
		\bigg[
		\underset{
			\tupleof{
				{
					\policy_i
				}
			}
		}
		{{\arg\max}}
		~{\prob}
		\Big[ 
		\tupleof{ \tracesof{\trajs_{\tau_i} \sim \distributionof{\policy_i}}}
		\models 
		\varphi
		\Big]
		\bigg]_{i \in \{1, \ldots,n\}}
	\end{align*}
	where $\distributionof{\policy_1},\ldots,\distributionof{\policy_n}$ 
	are the distributions over set of paths 
	generated by policy spaces $\policy_1,\ldots, \policy_n$. 
	%
	That is, $\tupleof{\maxpolicy_1, \ldots, \maxpolicy_n}$ 
	maximizes the probability $\prob$ 
	such that the generated tuple of sets of traces
	$\tupleof{ \tracesof{\trajs_{\tau_1} \sim \distributionof{\policy_1}}, \ldots, \tracesof{\trajs_{\tau_n} 
	\sim \distributionof{\policy_n}}}$ from $\MDP$ satisfies $\varphi$. 
\end{boxK}
\vspace{-4mm}
\paragraph{Example.} 
Consider the MDP in~\Cref{fig:mdpexmp} and the following \hyperltl formula:
$$\varphi _ {\mathsf{exp}}\triangleq\forall\tau _ 1.\exists\tau _ 2.\Big(\F i _ {\tau _ 1}\wedge\G dist(\langle x,y\rangle_ {\tau _ 1},\langle x, y\rangle_ {\tau _ 2})<3\Big)$$
Suppose agent $\droneone$ with $\pi _ 1$ draws samples $\mathcal{Z} _ {\tau _ 1} =\lbrace\zeta^1_\droneone,\zeta^2_\droneone\rbrace$ from the MDP:
\small
$$\zeta_\droneone^1:\underbrace{\langle 2,0\rangle} _ 
{a}\overset{\text{R}}{\rightarrow}\underbrace{\langle2,1\rangle} _ {b} 
\overset{\text{R}}{\rightarrow}\underbrace{\langle 2,2\rangle} _ 
{c}\overset{\text{U}}{\rightarrow}\underbrace{\langle 1,2\rangle} _ 
{f}\overset{\text{L}}{\rightarrow}\underbrace{\langle0,2\rangle} _ {i} 
~~~~~~~~~~\zeta_\droneone^2:\underbrace{\langle 2,0\rangle} _ 
{a}\overset{\text{R}}{\rightarrow}\underbrace{\langle2,1\rangle} _ 
{b}\overset{\text{R}}{\rightarrow}\underbrace{\langle 1,1\rangle} _ 
{e}\overset{\text{L}}{\rightarrow}\underbrace{\langle 0,1\rangle} _ 
{h}\overset{\text{U}}{\rightarrow}\underbrace{\langle0,2\rangle} _ {i}$$
Agent $\dronetwo$ with policy $\pi _ 2$ draws $\mathcal{Z} _ 
{\tau _ 2} =\lbrace \zeta^1_\dronetwo,\zeta^2_\dronetwo\rbrace$:
\small
$$\zeta_\dronetwo^1:\underbrace{\langle 2,0\rangle} _ 
{a}\overset{\text{U}}{\rightarrow}\underbrace{\langle1,0\rangle} _ 
{d}\overset{\text{U}}{\rightarrow}\underbrace{\langle 0,0\rangle} _ 
{g}\overset{\text{R}}{\rightarrow}\underbrace{\langle 1,0\rangle} _ 
{d}\overset{\text{R}}{\rightarrow}\underbrace{\langle1,0\rangle} _ {d}
~~~~~~~~~~\zeta_\dronetwo^2:\underbrace{\langle 2,0\rangle} _ 
{a}\overset{\text{U}}{\rightarrow}\underbrace{\langle2,1\rangle} _ 
{b}\overset{\text{L}}{\rightarrow}\underbrace{\langle 1,1\rangle} _ {e} 
\overset{\text{R}}{\rightarrow}\underbrace{\langle 1,2\rangle} _ {f} 
\overset{\text{L}}{\rightarrow}\underbrace{\langle0,2\rangle} _ {i}$$ 
\normalsize
Notice that, the number of samples can be more than two.
We now calculate the probability of satisfying $\varphi $ using $\mathcal{Z}_{\tau_1}$ and $\mathcal{Z} _ {\tau _ 2}$ (obtained by $\pi _ 1$ and $\pi _ 2$) as follows: 
$$
\mathsf{Traces}(\langle\lbrace \zeta_\droneone^1\rbrace,\mathcal{Z} _ {\tau _ 2}\rangle)\models\varphi _ {\mathsf{exp}}
\quad
\mathsf{Traces}(\langle\lbrace \zeta_\droneone^2\rbrace,\mathcal{Z} _ {\tau _ 2}\rangle)\models\varphi _ {\mathsf{exp}},
$$
where $\zeta_\dronetwo^2$ is the witness to $\tau_2$ (existentially quantified) for both satisfaction relations.  
%
Hence, given $\mathcal{Z}_1$ and $\mathcal{Z}_2$, the probability of satisfying $\varphi_\mathsf{exp}$ is evaluated as:
$$\mathbb{P} _ {\langle \pi _ 1, \pi _ 2\rangle}\Big[ \mathsf{Traces}(\langle\mathcal{Z} _ {\tau _ 1},\mathcal{Z} _ {\tau _ 2}\rangle)\models\varphi _ {\mathsf{exp}}\Big]=1$$
Now, if $\varphi _ {\mathsf{exp}}$ had the form $\forall\forall$, 
the evaluation has to go over all combinations between $\mathcal{Z} _ {\tau _ 1}$ and $\mathcal{Z} _ {\tau _ 2}$:
$$\mathsf{Traces}(\langle\lbrace\zeta_\droneone^1\rbrace,\lbrace\zeta_\dronetwo^1\rbrace\rangle)\not\models\varphi _ {\mathsf{exp}} 
\quad
\mathsf{Traces}(\langle\lbrace\zeta_\droneone^1\rbrace,\lbrace\zeta_\dronetwo^2\rbrace\rangle)\models\varphi _ {\mathsf{exp}}$$ 
$$\mathsf{Traces}(\langle\lbrace\zeta_\droneone^2\rbrace,\lbrace\zeta_\dronetwo^1\rbrace\rangle)\not\models\varphi_{\mathsf{exp}} 
\quad\quad
\mathsf{Traces}(\langle\lbrace\zeta_\droneone^2\rbrace,\lbrace\zeta_\dronetwo^2\rbrace\rangle)\models\varphi_{\mathsf{exp}}$$

Thus, the satisfaction probability of $\forall\forall\psi$ would be 0.5. This example demonstrates that the probability of satisfying a HyperLTL formula crucially depends on its quantifier structure.

\vspace{-3mm}
\section{Algorithmic Details of \method} 
\label{sec:rlhp}
\vspace{-2mm}
To solve the problem formally stated in~\Cref{sec:problem}, 
our algorithm proceeds in the following three steps. 
We first Skolemize $\varphi$~\cite{skolem1920} to eliminate quantifier alternations and simplify the learning task. 
We then define quantitative semantics for \HyperLTL, converting satisfaction checking into robustness value optimization. 
Finally, we train a neural network using these robustness signals to learn optimal policies that solve the original learning problem.
\subsection{Step 1: HyperLTL Skolemization}
\label{sec:skolemization}

%
Let a \HyperLTL formula be of the form
$
\varphi =
\quant_1 \tracevar_1. \quant_2 \tracevar_2. \ldots \quant_n. \tracevar_n.~
\psi(\tracevar_1, \tracevar_2, \ldots, \tracevar_n),
$
where for $1\leq\ell\leq n$, each $\quant_\ell \in \{\forall, \exists\}$ 
quantifies a trace variable $\tracevar_\ell$, and $\psi$ is a quantifier-free \LTL formula.
%
We first Skolemize $\varphi$, producing $\skolemized{\varphi}$, to eliminate quantifier alternations.
%
We define $\existsof{} = \{i \mid \quant_i = \exists\}$ and
$\forallsof{} = \{j \mid \quant_j = \forall\}$ 
as the sets of existential and universal quantifier indices, respectively.
%
For each $i \in \existsof{}$, we denote $\forallsof{i} = \{j < i \mid \quant_j = \forall\}$ as the index set of all universal quantifiers preceding $\quant_i$.
A {\em Skolem function} for each $i \in \existsof{}$ is defined as
$
\skolemfunc_i: \Tr^{|\forallsof{i}|} \rightarrow \Tr,
$
and reduces to a constant function when $\forallsof{i} = \emptyset$. 
A trace assignment $\traceassignment$
is {\em consistent} with $\skolemfunc_i$, if $\traceassignment(\tracevar_{i_j}) \in \Tr$ 
\cav{for all $j \in \forallsof{i}$}, and 
$\traceassignment(\tracevar_i) = 
\skolemfunc_i
\big(
\traceassignment(\tracevar_{i_1}), 
\traceassignment(\tracevar_{i_2}), \ldots,
\traceassignment({\tracevar_i}_{\sizeof{\forallsof{i}}} )
\big)
$ 
for all $i \in \existsof{}$, where 
$\forallsof{i} = \setof{i_1 < i_2 < \cdots < i_{\sizeof{\forallsof{i}}}}$.
\cavcomments{for all $i \in \forallsof{i}$ should be for all $j \in \forallsof{i}$}
If $(\Tr, \traceassignment, 0) \models \varphi$ for every trace assignment 
$\traceassignment$ consistent with all $\skolemfunc_i$, 
then each $\skolemfunc_i$ is said to \emph{witness} the satisfaction of $\varphi$~\cite{wz24}.
%
%
For the inner \LTL formula $\psi$ (i.e., obtaining $\skolemized{\psi}$), we replace each proposition $p_{\tau_i}$ with $p_{\skolemfunc_i}$ for all $p \in \AP$ and $i \in \existsof{}$, thereby instantiating variables of the existentially quantified paths via their Skolem witnesses. 
%
In general, a Skolemized $\varphi$ is of the following form:
\vspace{-1mm}
\begin{align}
\label{eq:skolem}
\skolemized{\varphi} = 
\underbrace{\exists \skolemfunc_{i}(\tau_{i_1}, \ldots, \tau_{i_{|\forallsof{i}|}})}_{\text{for each } i \in 
\existsof{}}.~
\underbrace{\spvertund{1.5ex}\forall \tracevar_{j}.}_{\text{for each }j \in \forallsof{}}\skolemized{\psi}
\end{align}
\vspace{-4mm}

Based on this transformation, 
we re-write the problem statement from~\Cref{sec:problem} as follows. 
We first define the \emph{image} of $\skolemfunc_i$: 
$$
\imageof{\skolemfunc_i}
\definedas
\{ \skolemfunc_i(\trace_{i_1}, \ldots, \trace_{i_{|\forallsof{i}|}}) \mid 
t_{i_{j}} \in \tracesof{			{
				\trajs_{\tau_{i_j}} \sim \distributionof{\policy_{i_{j}}}
		}},\ 
j \in \forallsof{i} \}
$$
That is, $\imageof{\skolemfunc_i}$ is the set of mapped traces (which 
$\tau_{i_j}$ for each ${i \in \existsof{}}$ ranges over) from all possible 
preceding $\forall$-quantified $\tau_{i_j}$, where each $\trace_{i_j}$ is from its 
own sampled trace set $\trajs_{\tau_{i_j}}$.
%
Now, let us use $\orderedsets$ as a notation to ensure the collection of trace sets are 
ordered w.r.t. their path indices.
Given two tuples of sets of traces $\Tr_1$ and $\Tr_2$, 
we define $\Tr_1 \orderedsets \Tr_2 \definedas \tupleof{\tracesof{\trajs_{\tau_x}}}_{x \in \{1 \cdots 
n\}}$, where each $\tracesof{\trajs_{\tau_x}}$ is either in $\Tr_1$ or $\Tr_2$ (and not both).
%
Given an MDP $\MDP$ and a \HyperLTL specification $\varphi$
of the form
$
\quant_1 \tracevar_1. \quant_2 \tracevar_2. \ldots 
\quant_n \tracevar_n.~ 
\psi
$,
our goal is to compute (1) a tuple of Skolem witnesses
$ 
\tupleof{\skolemfunc_{i}}_{i \in \existsof{}}
$,
and (2) a tuple of policies 
$ 
\tupleof{\maxpolicy_{j}}_{j \in \forallsof{}}
$, such that:
\begin{align*}
	& \tupleof{\maxpolicy_j}_{j \in \forallsof{}}
	\in
	\bigg[
	\underset{
		\tupleof{
			{
				{\policy_{{j}}}
			}
		}
	}
	{{\arg\max}} 
	~\prob
	\Big[ 
	\tupleof{ 
		\imageof{\skolemfunc_i}
	}
	\orderedsets
	\tupleof{ 
		\tracesof{			{
				\trajs_{\tau_j} \sim \distributionof{\policy_{{j}}}
		}}
	}
	\models 
	\skolemized{\varphi}
	\Big]
	\bigg]_{{i \in \existsof{}}, {j \in \forallsof{}}}
\end{align*}
That is, the tuple of policies $\tupleof{\maxpolicy_j}$ maximizes the probability that
the ordered collection of (1) the generated traces of all universal quantifiers
\rebuttal{$\tupleof{ \tracesof{\trajs_{\tau_j}\sim \distributionof{\policy_{{j}}} }}_{j \in \forallsof{}}$}
and (2) \rebuttal{the sets of mapped traces (i.e., the image) of each Skolem witness for all existential quantifiers
$\tupleof{\imageof{\skolemfunc_i}}_{i \in \existsof{}}$}
together satisfies $\skolemized{\varphi}$.
Notice that in the updated problem statement, we compute policies only for universally quantified traces, while Skolem functions are learned for existentially quantified traces to witness the optimality.


\vspace{-2mm}
\subsection{Step 2: Policy Learning with Quantitative Semantics}
\label{sec:quantsemantics}
\vspace{-1mm}
To transform the satisfaction checking problem (i.e., determining $\models$) into an optimization task, we define quantitative semantics for \HyperLTL, extended from~\cite{li2017reinforcement}. 
In particular, we evaluate the Skolemized \HyperLTL formula $\skolemized{\varphi}$ over tuples of sampled paths $\tupleof{\traj_1, \traj_2, \ldots, \traj_n}$ from $\MDP$. 
%
%
\begin{figure}[t]
	\begin{center}
		\renewcommand{\arraystretch}{1.5}
		\begin{boxK}
				\vspace{-2mm}
				\input{robustnessLTL}
				\vspace{-3mm}
		\end{boxK}	
	\end{center}
	\vspace{-1.5mm}
	\caption{Quantitative semantics for \LTL. }
	\label{fig:quantsemantics}
	\vspace{-5.5mm}
\end{figure}

%
\vspace{-2mm}
\paragraph{Robustness for a Single Trace.}
Let $\realnum$ be the set of real numbers and $\Psi$ the set of all \LTL formulas.
We define a valuation function $\predicatefunc: 2^\AP \rightarrow \realnum$ that assigns a real value to a set of atomic propositions, provided as part of the input.
Given a state $\state \in \states$ of an MDP $\MDP$, 
the quantitative semantics are defined over predicates in the form of 
$\predicatefuncof{\labelsof{\state}} < \constant$, where 
$\constant$ is a user-specified threshold.
%
Next, we define a {\em robustness function} 
$\rbvalue: \tracesof{\trajs^\finite}\times \Psi \rightarrow \realnum$ 
that assigns a robustness value to a finite trace for an \LTL formula.
Intuitively, the robustness value evaluates ``how far'' the given finite trace 
is from satisfying $\psi$.
The complete quantitative semantics is shown in~\Cref{fig:quantsemantics}. 
We use constants $\rbvalue_{\mathit{max}}$ and $\rbvalue_{\mathit{min}}$ 
for the maximum and minimum robustness values, respectively.  
Given a trace, a higher $\rbvalue$ value implies the trace has higher robustness to satisfy $\psi$, 
and a lower $\rbvalue$ value means the trace is less likely to satisfy $\psi$ (e.g., a potential violation). 
%
%
%

%

Formally, given an \LTL formula $\psi$ and an MDP $\MDP$, 
a path $\traj$ with a higher robustness value indicates that it has higher probability to satisfy $\psi$.  
%
The optimization task of seeking a single policy $\maxpolicy$ is:
\[
\maxpolicy \in 
\underset{
	\policy
}
{{\arg\max}}~
\underset{\traj \sim \distributionof{\policy}}
{\prob}
\Big[ 
\rbvalueof{\traceof{\trajfromto{\traj}{0}{k}}, \psi} \converge \rbvalue_{\mathit{max}} 
\Big]
\]
Here, for simplicity, we use the notation $\converge$ to represent convergence.
That is, $\maxpolicy$ maximizes the probability of satisfying $\psi$ over the distribution of paths generated by policy $\policy$.

\vspace{-3mm}
\paragraph{Robustness for a Tuple of Traces.}
\cav{
To evaluate the robustness value over multiple traces, we first define a $\zip$ function that pointwise bundles a tuple of traces.
Given a tuple of finite traces $\tupleof{\trace_1, \ldots, \trace_n}$, we derive a zipped trace 
$
\zipof{\tupleof{\trace_1, \ldots, \trace_n}} 
$
and for all $i \geq 0$, 
$\zipof{\tupleof{\trace_1, \ldots, \trace_n}} (i) 
\definedas
\tupleof{\trace_1(i),\ldots,\trace_n(i)}
$. 
%
%
Given an \LTL formula, 
a tuple of paths $\tupleof{\traj_1, \traj_2, \ldots, \traj_n}$ has higher probability of satisfying $\psi$ if the robustness value of
$
\zip\big(\tupleof{\traceof{\trajfromto{\traj_1}{0}{k_1}}, \traceof{\trajfromto{\traj_2}{0}{k_2}}, \ldots, \traceof{\trajfromto{\traj_n}{0}{k_n}}}\big)
$
converges to $\rbvalue_{\mathit{max}}$ w.r.t. $\psi$ for some $k_1, \ldots, k_n$, where $0 \leq 
k_\ell \leq \sizeof{\traj_\ell}$ for each $1\leq\ell\leq n$.} 
Thus, the optimization task of computing a tuple of policies 
$\tupleof{\maxpolicy_1, \maxpolicy_2, \ldots, \maxpolicy_n}$ 
that maximizes the robustness becomes: 
\[
\tupleof{\maxpolicy_\ell}_{\ell \in \setof{1,...n}}
\in
\bigg[
\underset{
	\tupleof{
		\policy_{{\ell}}
	}
}
{{\arg\max}}~
{\prob}
\Big[
\rbvalueof{
	\zip
	\big(
	\tupleof{
	\traceof{\trajfromto{\traj_\ell}{0}{k_\ell} \sim \distributionof{\policy_{{\ell}}}}
	}
	\big), 
	\psi}
\converge 
\rbvalue_{\mathit{max}}
\Big]
\bigg]_{\ell \in \setof{1,\dots,n}}
\]
This formulation reflects that \LTL formulas are implicitly universally quantified as we emphasized in~\Cref{sec:intro}. For instance, model checking $\forall^\finite. \psi$ reduces to checking $\forall.\psi$ via self-composition~\cite{barthe2011secure,bf18}.
\vspace{-6mm}
\paragraph{Robustness for Skolemized HyperLTL.}
Optimizing an alternating \HyperLTL formula requires that policies for universally quantified paths simultaneously optimize the existentially quantified paths. 
%
%
\rebuttal{
In order to preserve ordering, we use $\orderedunion$ to make sure the union of trace tuples are in order w.r.t. their path indicies. 
That is, given two tuple of traces $\tupleof{\cdot}_1$ and $\tupleof{\cdot}_2$, 
we define $\tupleof{\cdot}_1 \orderedunion \tupleof{\cdot}_1 \definedas \tupleof{\traceof{\traj_{x}}}_{x \in \{1 \cdots n\}}$, 
where each $\traceof{\traj_{x}}$ is either from $\tupleof{\cdot}_1$ or $\tupleof{\cdot}_2$.      
%
}
Given a Skolemized \HyperLTL formula, the inner LTL formula is denoted as $\skolemized{\psi}$, as defined in~\eqref{eq:skolem}.
For each $i \in \existsof{}$ and $j \in \forallsof{}$, we say a tuple of paths
$\tupleof{\traj_1, \traj_2,\ldots,\traj_n}$
satisfies $\psi$ if and only if: 
\[
\Big[
\rbvalue
\Big(
	\zip
	\big(
	\tupleof{\traceof{\trajfromto{\traj_i}{0}{k_i}}}~
	\orderedunion
	\tupleof{\traceof{\trajfromto{\traj_j}{0}{k_j}}}
	\big), 
	\skolemized{\psi}
\Big)
\Big]_{{i \in \existsof{}},{j \in \forallsof{}}}
\converge \rbvalue_{\mathit{max}}
\]
Essentially, the set
$
\tupleof{\traceof{\trajfromto{\traj_i}{0}{k_i}}}_{i \in \existsof{}}
$ captures the tuple of traces produced by each Skolem function for each existential quantifiers \rebuttal{(i.e., a trace from $\imageof{\skolemfunc_i}$) for each ${i \in \existsof{}}$)}, 
while 
$
\tupleof{\traceof{\trajfromto{\traj_j}{0}{k_j}}}_{j \in \forallsof{}}
$
corresponds to the tuple of traces for each ${j \in \forallsof{}}$.	
%
%
That is, we can formalize the optimization task for a Skolemized HyperLTL as follows: 
%
\vspace{-1mm}
\rebuttal{
\begin{align}
\bigg[
\tupleof{\maxpolicy_i}
\orderedunion
\tupleof{\maxpolicy_j}
\in 
\underset{
	\tupleof{\policy_i}
	\orderedunion
	\tupleof{\policy_j}
}
{{\arg\max}}
{\prob}
\Big[
\rbvalue
\Big(
\zip
\big(
\tupleof{\traceof{\trajfromto{\traj_i}{0}{k_i}& \sim \distributionof{\policy_{{i}}}}\big}
\orderedunion
\tupleof{\traceof{\trajfromto{\traj_j}{0}{k_j} \sim \distributionof{\policy_{{j}}}}\big}
\big),
\nonumber
\\ &
\skolemized{\psi}
\Big)
\converge 
\rbvalue_{\mathit{max}}
\Big]
\bigg]_{i \in \existsof{}, j \in \forallsof{}} 
\label{eq:opt}	
\end{align}
Notice that, for each $i \in \existsof{}$ (i.e., a Skolem function), 
the evaluation of $\rbvalue$ dependes on whether the 
sampled trace $\traj_i \sim \distributionof{\policy_{{i}}}$ \emph{serves} 
as a witness for its preceding ${\tupleof{\traj_{i_j}}}_{{i_j} \in \forallsof{i}}$. 
That is, 
$
\rbvalueof{\skolemfunc_{i}, \psi} \definedas \rbvalueof{\skolemfunc_{i}(\traceof{\traj_{i_1}}, \ldots, \traceof{\traj_{i_{{|\forallsof{i}|}}}}), \psi}
$.
Hence, the optimization of $\tupleof{\maxpolicy_i}$ depends on $\tupleof{\maxpolicy_j}$, which is necessary to capture the semantics of a HyperLTL formula with quantifier alternation (we provide a comprehensive example in~\Cref{appendix:running-example}).
}
Finally, we show that the set of optimal policies obtained by~\eqref{eq:opt} 
also solves the original optimization problem in~\Cref{sec:problem}
(detailed proof in~\Cref{proof:theorems}).

\newcommand{\thrmone}{
	Given an MDP $\MDP$ and a \HyperLTL formula $\varphi$,  
	an optimal tuple of policies
	$\tupleof{\maxpolicy_i}_{i \in \existsof{}} 
	\orderedunion	
	\tupleof{\maxpolicy_j}_{j \in \forallsof{}}$
	for $\skolemized{\varphi}$ 
	is also an optimal tuple of policies for $\varphi$, 
	that optimizes the probability of satisfying $\varphi$ in $\MDP$ (the problem statement 
	in~\Cref{sec:problem}).}
\begin{boxK}
\begin{theorem}
\thrmone
\label{theorem:skolem}
\end{theorem}
\end{boxK}

\subsection{Step 3: Reinforcement Learning for \HyperLTL}
\label{sec:step3}


%
Now, we solve the optimization task in~\eqref{eq:opt} using RL, where the reward signal is defined by the robustness values from~\Cref{sec:quantsemantics}.
As our MDP model lacks a built-in reward function, we use robustness values to guide learning.
To avoid confusion with standard RL terminology, we refer to expected robustness value when mentioning expected rewards.
For simplicity, we assume all sampled paths have equal length.

\vspace{-3mm}
\paragraph{Optimizing Expected Reward.}
The goal of this step is to learn an optimal neural network $\optnnpolicy$, a parameterized compositional function that synthesizes the policies solving~\eqref{eq:opt}.
%
%
The construction of learning constraints is inspired by the well-known {\em Bellman Equation}~\cite{barron1989bellman}, which characterizes the optimal expected reward of a decision process as a recursive function of immediate reward and the value of future states.
%
Let us use $\state_\leafindex$ to denoted the state of a zipped path at position $\leafindex$ 
(i.e., $\state_\leafindex \definedas 
\zipof{\tupleof{\trace_1, \ldots, \trace_n}} (\leafindex)$).
%
We define the \emph{immediate reward} for a state-action pair $(\state_\leafindex, \action_\leafindex)$ based on the robustness value of the zipped trace from~\eqref{eq:opt}:
\newcommand{\eqrewardfunction}{
\reward(\state_\leafindex, \action_\leafindex)
\definedas
\Big[
\rbvalue
\Big(
{
	\zip
	\big(
	\tupleof{
	\traceof{\trajfromto{\traj_i}{0}{\leafindex}}
	}
	\orderedunion
	\tupleof{
	\traceof{\trajfromto{\traj_j}{0}{\leafindex}}
	}
	\big), 
	\skolemized{\psi}
}
\big)
\Big]_{{i \in \existsof{}},{j \in \forallsof{}}}
}
\begin{equation*}
\eqrewardfunction
\end{equation*}
%
%
%
The {\em expected reward} $\expectedreward(\state_\leafindex, \action_\leafindex)$ follows the classic Bellman formulation, combining immediate reward and future reward based on some discount factor $\discount$ (see~\Cref{appendix:expected_reward} for details). 
%
%
\newcommand{\moving}{\small\textsf{move}}
Intuitively, $\expectedreward(\state_\leafindex, \action_\leafindex)$ quantifies the long-term utility of taking action $\action_\leafindex$ at state $\state_\leafindex$.
Finally, for any arbitrary state $s$ and action $a$, the Bellman Equation then defines the $Q$-value recursively for each $(\state, \action) \in \states \times \actions$, denoted as $\Qvalue^\nnpolicy(\state, \action)$:
\begin{equation*}
\Qvalue^\nnpolicy(\state, \action)
~\definedas~
\sum_{\state' \in \states}
{\trans(\state, \action, \state')}
\bigg[
{\rewardof{\state}{\action}}
+
{\discount}
\sum_{\action' \in \actions}
{
	\nnpolicy({\action'}\mid{\state'}) 
	~\expectedrewardof{\state'}{\action'}
}
\bigg],
\end{equation*}
where $\nnpolicy({\action'}\mid{\state'})$ indicates
that $\nnpolicy$ takes action $\action'$ on state $\state'$.
%
Consequently, the optimal action-value function for each $(\state,\action)$ is:
\begin{equation}
\Qvalue^{\optnnpolicy}(\state, \action) 
\definedas
\underset{\tiny\nnpolicy}{\maximum}~
\Qvalue^{\nnpolicy}(\state, \action)
\label{eq:optQvalue}
\end{equation}
Intuitively, $\Qvalue^{\optnnpolicy}(\state, \action)$ captures the maximum expected reward achievable from a state $\state$ by taking action $\action$.
%
We remark that, our framework samples all paths for each quantifier simultaneously.
%
%
That is, the learned neural network $\optnnpolicy$ induces a set of $n$ functions $\setof{\optnnpolicyof{1}, \ldots, \optnnpolicyof{n}}$, where $n = \sizeof{\traceVarsOf{\varphi}}$, and for all $1\leq\ell \leq n$, $\optnnpolicyof{\ell}$ maps a state to an optimal actions for a path $\traj_\ell$.
%
%

%
%

\paragraph{Constructing Policies.}
Based on the learned $\optnnpolicy$, we now construct the tuple of policies 
$\tupleof{\maxpolicy_i}_{i \in \existsof{}}$ 
and $\tupleof{\maxpolicy_j}_{j \in \forallsof{}}$
that solves~\eqref{eq:opt} for each $k$-step ranging over the sample size as follows. 

\begin{itemize}
	\vspace{-3mm}
	\item For each ${j \in \forallsof{}}$, we inductively 
	construct the policies:
	\[
	\maxpolicy_j (\trajfromto{\traj_j}{0}{k})
	\definedas
	\optnnpolicyof{j}(s_k)
	\]
	\item For each $i \in \existsof{}$, we construct a Skolem witness as follows:
	%
	\begin{align*}
	\maxpolicy_i (\trajfromto{\traj_i}{0}{k})
	\definedas
	\optnnpolicyof{i}
	\Big(
	\skolemfunc_i
	\big(
	\traceof{\trajfromto{\traj_{i_1}}{0}{k}}, 
	\ldots, 
	\traceof{\traj_{i_\sizeof{\forallsof{i}} \bracketof{0:k} }}
	\big)
	\Big)
	\end{align*}
\end{itemize}

%
%
That is, the optimal policy for 
a finite prefix $\traj_i$ with $i \in \existsof{}$ 
depends on the optimal actions taken along the preceding universal paths $\traj_{i_1}, \ldots, \traj_{i_\sizeof{\forallsof{i}}}$, illustrating how our framework captures the dependency of an existential path on the universally quantified ones.
%
To this end, from the learned neural network
$\optnnpolicy$, we succesfully derive two tuples $\tupleof{\maxpolicy_i}_{i \in \existsof{}}$ 
and
$\tupleof{\maxpolicy_j}_{j \in \forallsof{}}$ 
for~\Cref{eq:opt}.
%

\newcommand{\thrmtwo}{Given an MDP $\MDP$ and a \HyperLTL formula $\varphi$, 
	the optimal neural network function $\optnnpolicy$
	derives 
	a tuple of Skolem function witnesses 
	$\tupleof{\skolemfunc_{i}}_{i \in \existsof{}}$
	and a tuple of optimal policies 
	$\tupleof{\maxpolicy_{j}}_{j \in \forallsof{}}$ 
	that optimize the satisfaction of $\skolemized{\varphi}$.
}
\begin{boxK}
\begin{theorem}
	\thrmtwo
	\label{theorem:Qlearning}
\end{theorem}
\end{boxK}
\vspace{-2mm}

\Cref{theorem:Qlearning} gives the premise of~\Cref{theorem:skolem}, which, in turn, solves the 
original problem stated in~\Cref{sec:problem} (detailed proof in~\Cref{proof:theorems}).

\begin{figure}[t!]
	\begin{minipage}{.42 \textwidth}
		\centering
			\scalebox{.85}{
				\input{figs/approach}
			}
			\captionof{figure}{Overview of \method.
			}\label{fig:approach}
		\end{minipage}
		\hfill
		\begin{minipage}{.52 \textwidth}
				\centering
				\captionof{table}{\HyperLTL specifications of case studies.}\label{tab:cases_formula}
				\vspace{-3mm}
				\renewcommand{\arraystretch}{0.1}  
\begin{tabular}{p{0.4cm}ll}
    \toprule
    {\small (SRL)}& 
    \scalebox{0.7}{
      \thead[l]{
      $
      \forall \tronesafe . \exists \trtwosafe.~ 
      \always \tupleof{x_{\tau_1}, y_{\tau_1}} \neq \tupleof{x_{\tau_2}, y_{\tau_2}}  
      \land 
      \F \tupleof{x_{\tau_1}, y_{\tau_1}} = \tupleof{x_{G1}, y_{G1}}  
      \land$ \\
      ~~~~~~~~~~ $\F \tupleof{x_{\tau_2}, y_{\tau_2}} = \tupleof{x_{G2}, y_{G2}}
      $
      }
    }
    \\
    \cmidrule(lr{0.25em}){1-2}
    {\small (DST)} &
    \scalebox{0.7}{
      \thead[l]{
        $\forall \traceone. \exists \tracetwo.  
        \F (\textit{T1}_\traceone \land 
          \F (\textit{T2}_\traceone \land 
            \F (\textit{T3}_\traceone) \dots))
        ~\land$~ \\
        ~~~~~~~~~~$\G~(\textit{step}_{\tracetwo} < \delta) \land \G~(|\textit{pos}_\traceone - \textit{pos}_\tracetwo| < 1)$\\
      }
    }
    \\
    \midrule
  {\small (PCP)} &
    \scalebox{0.7}{
        \thead[l]{
        $
        \forall \traceone. \exists \tracetwo.~ 
        \psi_{\semi_{\traceone}} ~\until~
        \big( \psi_{\extend_{\traceone, \tracetwo}} \land
        \underset{p \in \AP}{\bigwedge} (p_{\topp_{\tracetwo}}  
        \leftrightarrow 
        p_{\bottom_{\tracetwo}})
        \big)
        $\\
        ~~$\psi_{\semi_{\traceone}} \triangleq \Big[\underset{p \in \AP}{\bigwedge} (p_{\topp_{\traceone}} 
        \leftrightarrow p_{\bottom_{\traceone}})\Big]~\until (\#_{\topp_{\traceone}} \oplus 
        \#_{\bottom_{\traceone}})$\\
        ~~$\varphi_{\extend_{\traceone, \tracetwo}} \triangleq \Big[\underset{p \in \AP}{\bigwedge} \big(  (p_{\topp_{\traceone}} \leftrightarrow p_{\topp_{\tracetwo}})  \land  (p_{\bottom_{\traceone}} \leftrightarrow p_{\bottom_{\tracetwo}})   \big) \Big] ~ \until ~$ \\
        \hspace{30mm}$\big( (\#_{\topp_{\traceone}} \lor \#_{\bottom_{\traceone}} ) \land (\neg \#_{\topp_{\tracetwo}} \land \neg \#_{\bottom_{\tracetwo}})\big)$
        }
    }
    \\
    \bottomrule
  \end{tabular}
				
	\end{minipage}
	\vspace{-3mm}
\end{figure}

\vspace{-2mm}
	\section{Implementation and Experiments}\label{sec:Implementation}\label{sec:Experiments}
	\vspace{-2mm}
	\method is fully implemented (see~\Cref{fig:approach}).
	Given a \HyperLTL formula $\varphi$ with $n$ quantifiers, \method first constructs its 
	Skolemized form $\skolemized{\varphi}$, as prescribed 
	in~\Cref{sec:skolemization}.
	Next, at each step $t$, for each $i\in\{1,\ldots,n\}$, a policy $\policy_i$ observes a finite path $\trajfromto{{\traj_i}}{0}{t}$, and selects an action $\policy_i (\trajfromto{\traj_i}{0}{t})$, yielding an updated path $\trajfromto{{\traj_i}}{0}{t+1}$.
	The environment computes feedback using robustness function $\rho$ 
	to guide learning (\Cref{sec:quantsemantics}).
	The optimal tuple of policies $\tupleof{\maxpolicy_1, \ldots, \maxpolicy_n}$ are learned iteratively using a selected RL algorithm such as DQN~\cite{mnih_human-level_2015}, PPO~\cite{schulman2017proximalpolicyoptimizationalgorithms}, or CQ-Learning~\cite{10.5555/1838206.1838301} (\Cref{sec:step3}).
	We summarize all our case studies and their corresponding 
	HyperLTL specifications in~\Cref{tab:case_studies} and~\Cref{tab:cases_formula}, respectively.
	We also investigate the wildfire scenario introduced 
	in~\Cref{sec:intro} with larger grid-worlds.
	%
	All experiments are ran on an Apple M1 Max (10-core CPU, 24-core GPU).
	%
	%
	Additional details of experimental setups are provided in~\Cref{appendix:experiments}.

	\begin{table}[t]
		\vspace{-3mm}
		\centering
		\caption{Descriptions of the case studies.}
		\renewcommand{\arraystretch}{0.6} 
		\scalebox{0.85}{
			\begin{tabular}{cp{10.3cm}}
			\toprule
			\begin{tabular}[l]{@{}c@{}} Safe RL in Grid Worlds\\(SRL) \end{tabular}
			& 
			\begin{tabular}[l]{@{}l@{}} Two agents collaboratively learn policies $\tupleof{\maxpolicy_1,\maxpolicy_2}$ to reach \\ their respective goals while avoiding collisions in a grid world. \end{tabular}
			\\
			\midrule
			\begin{tabular}[l]{@{}c@{}} Deep Sea Treasure\\(DST) \end{tabular} 
			& 
			\begin{tabular}[l]{@{}l@{}} Two agents (a driver and a treasure collector) navigate among treasures.\\ Learn policies $\tupleof{\maxpolicy_1,\maxpolicy_2}$ such that, for all ways the collector maximizes \\ treasures, a safe route exists for the driver to exit within the time limit $\delta$. \end{tabular}
			\\
			\midrule
			\begin{tabular}[l]{@{}c@{}} Post Correspondence Problem\\(PCP) \end{tabular}
			& 
			\begin{tabular}[l]{@{}l@{}} Learn policies $\tupleof{\maxpolicy_1,\maxpolicy_2}$ for PCP: for any semi-matching sequence of \\ dominos, there exists an extension into full matches. \end{tabular}
			\\
			\bottomrule
			\end{tabular}
		}
		\label{tab:case_studies}
		\vspace{-5mm}
	\end{table}

\begin{table}[b!]
    \vspace{-7mm}
    \caption{SRL, avg. and std. error of 10 evaluations after 1k training episodes, 10 runs, where S/B stands for step-bound which is 100 steps for all cases.}
    \label{tab:GW}
    \centering
    \resizebox{.88\textwidth}{!}{%
    \begin{tabular}{lccccccc}
        \toprule
        & 
        & \multicolumn{2}{c}{CQ} 
        & \multicolumn{2}{c}{CQ + Shield~\cite{elsayed2021safe}} 
        & \multicolumn{2}{c}{CQ + \method} \\
        \cmidrule(lr){3-4} \cmidrule(lr){5-6} \cmidrule(lr){7-8}
        Maps & No. Agents  & Steps & Collisions & Steps & Collisions & Steps & Collisions \\
        \midrule
        ISR       & \multirow{4}{*}{2} & 27.95$\pm7.4$ & 0.19$\pm0.1$  & 17.40$\pm2.2$ & \textbf{0.00}$\pm0.0$ & \textbf{7.58}$\pm0.3$ & 0.25$\pm0.2$ \\
        Pentagon  &                    & 36.46$\pm7.7$ & 0.28$\pm0.1$  & 75.20$\pm12.6$ & \textbf{0.00}$\pm0.0$ & \textbf{11.90}$\pm4.6$ & 0.53$\pm0.5$ \\
        SUNY      &                    & 11.99$\pm0.5$ & 0.01$\pm0.0$  & \textbf{11.50}$\pm0.3$ & \textbf{0.00}$\pm0.0$ & 12.48$\pm0.6$ & \textbf{0.00}$\pm0.0$ \\
        MIT       &                    & 41.28$\pm8.5$ & 0.20$\pm0.1$  & 33.46$\pm3.4$ & \textbf{0.00}$\pm0.0$ & \textbf{23.20}$\pm0.5$ & \textbf{0.00}$\pm0.0$ \\
        \midrule
			ISR & \multirow{4}{*}{3}  & 98.79$\pm$0.8 & 12.68$\pm$3.8  & S/B & \textbf{0.00}$\pm$0.0  & \textbf{74.18}$\pm$5.1 & 7.78$\pm$1.0 \\
			Pentagon & & 97.15$\pm$2.4 & 16.46$\pm$7.2 & S/B & \textbf{0.00}$\pm$0.0 & \textbf{78.82}$\pm$1.7 & 10.92$\pm$1.4 \\
			SUNY & & 84.89$\pm$7.9  & 0.63$\pm$0.2 & 82.35$\pm$4.1& \textbf{0.00}$\pm$0.0  & \textbf{44.95}$\pm$8.3 &  0.71$\pm$0.4 \\
			MIT & &	96.96$\pm$1.8 & 2.83$\pm$1.3 & S/B & \textbf{0.00}$\pm$0.0& \textbf{71.53}$\pm$7.7 & 1.58$\pm$0.7 \\
			\bottomrule
    \end{tabular}
    }
\end{table}


	\vspace{-3mm}
	\paragraph{Baseline.} 
	We experiment all cases using the setup from~\cite{li2017reinforcement} by comparing against 
	manually designed reward functions.
	To ensure a fair comparison, for DST, we use the reward function from~\cite{vptrb2011}; for the rest of the cases, we construct several reward functions and report the best-performing one (details are provided in~\Cref{appendix:experiments}). 
	%
	%
	For SRL, we also compare with the method called ``shield 
	synthesis''~\cite{elsayed2021safe}, where a 
	pre-synthesized {\em shield} enforces a safety property during 
	learning and, in this specific case, can reason about 
	inter-agent dependencies.

	\vspace{-2mm}
	\paragraph{Results and Analysis.} First, for SRL, we use the maps 
	from~\cite{10.5555/1558109.1558118} and compare 
	CQ-learning+\method against shield 
	synthesis~\cite{elsayed2021safe} (detailed map descriptions are in\Cref{appendix:experiments}).
	%
	%
	%
	It is worth noting that for~\cite{elsayed2021safe}, a shield has to be 
	explicitly designed to avoid unsafe states, whereas \method achieves 
	better safe and 
	goal-directed behavior purely 
	through the robustness values derived from a \HyperLTL formula.
	\borzoored{Our analysis for SRL is presented in~\Cref{tab:GW}. 
	\camera{In a two-agent setting}, the result shows that the policies learned by \method  
	outperform the shielded agent in all of the environments by requiring 
	fewer steps to reach the goal, except in SUNY, where their performance is quite similar.}
	Furthermore, in both SUNY and MIT maps, \method successfully avoids collisions as what we 
	replicated from~\cite{elsayed2021safe}.
	\camera{In a three-agent setting (where the HyperLTL formula is expanded to $\forall\forall\exists$), the shielding method successfully avoids all collisions; however, it failed to reach the goals within the bound of 100 steps in ISR, Pentagon, and MIT. 
	In contrast, \method was able to guide the agents to reach their goals significantly faster than both baselines and achieved fewer collisions compared to CQ-learning.}
	We emphasize that although shield synthesis can avoid collision in all 
	cases, it comes with the cost of spending a significant number of steps 
	to reach the goal.
	In contrast, \method maintains low collision rates in ISR and Pentagon, and learns policies that reach the goal with substantially fewer steps in ISR, Pentagon, and MIT maps.
	
	Second, for SRL, we compare DQN+\method with DQN and present the results in~\Cref{fig:result-saferl}. 
	The top four figures demonstrate that \method always achieves a higher number of successful 
	goal completions under the same number of learning episodes. 
	The bottom four figures show that \method requires fewer episodes to learn collision avoidance. 
	For DST, we compare PPO+\method with PPO, and DQN+\method with 
	DQN, using the reward functions from~\cite{vptrb2011} in both cases.
	We evaluate using two metrics: (1) the total amount of collected treasures ($\sum\raisebox{-1ex}{\includegraphics[height=2.5ex]{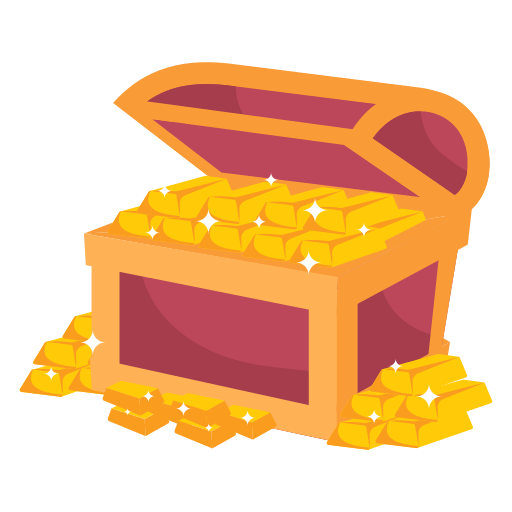}}$), and 
	(2) the average amount of collected treasures per step 
	($\sum\!\raisebox{-0.9ex}{\includegraphics[height=2ex]{figs/treasure-chest.png}}/\mathit{steps}$).
	As shown in~\Cref{tab:dst}, policies learned by \method outperform those learned by the baseline, regardless of the number of training episodes or the choice of the RL algorithm.

	\begin{wrapfigure}{r}{0.34\textwidth}
		\centering
		\vspace{-6mm}
		\resizebox{0.35\textwidth}{!}{%





\begin{tikzpicture}
    \begin{axis}[
        xlabel={\huge\bf Episodes},
        ylabel={\huge\bf \# of successful match},
        width=1.3\textwidth,
        height=0.8\textwidth,
        legend style={at={(0.25,0.67)},anchor=south,draw=black, font=\huge, inner sep=0pt},
        xtick={200,400,600,800,1000},
        xticklabel style={font=\huge},
        yticklabel style={font=\huge},
        ylabel style={yshift=15pt},
        xmin=0,
        xmax=1000
    ]
    \addplot[color=Blue,  ultra thick, solid , line width=4pt] table [x=episode, y=mean, col sep=comma] {figs/dom5.csv}; \addlegendentry{ \huge DQN+\method $(k=5)$}
    \addplot[color=Red,  ultra thick, solid ,line width=4pt] table [x=episode, y=mean, col sep=comma] {figs/dom6.csv};\addlegendentry{\huge DQN+\method $(k=6)$}
    \addplot[color=Green,  ultra thick, solid , line width=4pt] table [x=episode, y=mean, col sep=comma] 
    {figs/dom5_base_1.csv};\addlegendentry{\huge DQN $(k=5)$}
    \addplot[color=Goldenrod,  ultra thick, solid , line width=4pt] table [x=episode, y=tot_done, col sep=comma] 
    {figs/dom6_base.csv};\addlegendentry{\huge DQN 
    $(k=6)$}
    \addplot[name path=upper, draw=none] table [x=episode, y=upper, col sep=comma] {figs/dom5.csv};
    \addplot[name path=lower, draw=none] table [x=episode, y=lower, col sep=comma] {figs/dom5.csv};
    \addplot[fill=blue, fill opacity=0.15] fill between[of=upper and lower];
    \addplot[name path=upper_dqn6, draw=none] table [x=episode, y=upper, col sep=comma] {figs/dom6.csv};
    \addplot[name path=lower_dqn6, draw=none] table [x=episode, y=lower, col sep=comma] {figs/dom6.csv};
    \addplot[Red, opacity=0.2] fill between[of=upper_dqn6 and lower_dqn6];

    \addplot[name path=upper_dqn5base, draw=none] table [x=episode, y=upper, col sep=comma] {figs/dom5_base_1.csv};
    \addplot[name path=lower_dqn5base, draw=none] table [x=episode, y=lower, col sep=comma] {figs/dom5_base_1.csv};

    \addplot[Green, opacity=0.1] fill between[of=upper_dqn5base and lower_dqn5base];

    \end{axis}
    \end{tikzpicture}
		}
		\caption{PCP results, avg. and std. error of total matches, 10 runs.}
		\vspace{-6mm}
		\label{fig:dom_exp}
	\end{wrapfigure} 

	Finding policies to solve the PCP problem is inherently 
	challenging due to its undecidability.
	\method is able to learn optimal policies that solve PCP more efficient, 
	compare to the baseline.
	\Cref{fig:dom_exp} presents the performance comparison of DQN+\method with DQN, showing that \method achieves more successful matches in both 5- and 6-domino settings. 
	
	Lastly, we evaluate \method on the wildfire scenario and compare PPO+\method with PPO.
        \camera{As shown in~\Cref{tab:search}, even when scaling up the grid-world environment from $3^2$ to $5^2$, $8^2$, and $10^2$, \method consistently outperforms the baseline by requiring fewer steps to rescue victims, extinguish all fires, and maintain a safe distance between the two agents. 
        The step-bounds are set to 10k steps for the $3^2$ and $5^2$ environments, 20k steps for $8^2$, and 30k steps for $10^2$. 
        In fact, PPO fails to learn policies capable of rescuing victims ($O_2$) within the step-bounds for $5^2$ and $8^2$, and is unable to extinguish all fires ($O_1$) within the step-bound in the $10^2$ setting.}

	\begin{figure}[t!]
		\centering

		\scalebox{0.9}{
			\input{figs/saferl_comp1.tex} 
		}
		\ \\
		\scalebox{0.9}{
			\input{figs/saferl_comp2.tex}
		}
		\caption{SRL results, avg. and std. error of total number of goal completions by 2 agents (top) and collisions (bottom). DQN+\method (line~\ref{line:hyprl}) and DQN (line~\ref{line:reward}), over 10 runs.}\label{fig:result-saferl}
	
	\end{figure}

	

	%

	\begin{table}[t]
		\vspace{-2mm}
		\centering
		\begin{minipage}{0.57\textwidth}
		\caption{DST results, avg. and std. error over 10 evaluations across, 10 runs, and Epi. stands for ``Episodes''.}
		\label{tab:dst}
		\resizebox{\textwidth}{!}{%
		\begin{tabular}{p{21mm}p{5mm}p{15mm}p{15mm} p{23mm}p{5mm}p{15mm}p{15mm}}
			\toprule
			Method & Epi. & 
			$\sum\!\raisebox{-0.9ex}{\includegraphics[height=2ex]{figs/treasure-chest.png}}$ & 
			$\sum\!\raisebox{-0.9ex}{\includegraphics[height=2ex]{figs/treasure-chest.png}}/steps$ & 
			Method & Epi. & 
			$\sum\!\raisebox{-0.9ex}{\includegraphics[height=2ex]{figs/treasure-chest.png}}$ & 
			$\sum\!\raisebox{-0.9ex}{\includegraphics[height=2ex]{figs/treasure-chest.png}}/steps$ \\
			\midrule
			PPO & 500 & 4.31$\pm1.2$ & 0.17$\pm0.04$ & DQN & 500 & 1.08$\pm0.2$ & 0.05$\pm0.00$  \\
			PPO + \textsc{HypRL} & 500 & \textbf{22.93}$\pm2.2$ & \textbf{0.91}$\pm0.08$ & DQN + \textsc{HypRL} & 500 & \textbf{4.12}$\pm1.4$  & \textbf{0.14}$\pm0.05$ \\
			\midrule
			PPO & 1000 & 3.22$\pm0.8$ & 0.12$\pm0.03$ & DQN & 1000 & 1.45$\pm0.2$ & 0.02$\pm0.00$ \\
			PPO + \textsc{HypRL} & 1000 & \textbf{23.97}$\pm2.2$ & \textbf{1.12}$\pm0.08$ & DQN + \textsc{HypRL} & 1000 & \textbf{4.43}$\pm0.8$  & \textbf{0.21}$\pm0.03$ \\
			\bottomrule
		\end{tabular}%
		}
		\end{minipage}
		\hfill
		\begin{minipage}{0.42\textwidth}
		\centering
		\caption{Wildfire results, avg. and std. error of 10 trials after 5k episodes, 10 runs.}
		\label{tab:search}
		\resizebox{\textwidth}{!}{
			\renewcommand{\arraystretch}{0.91}
		\begin{tabular}{p{3mm}p{21mm}lll}
			\toprule
			Size & Method &  \texttt{Dist} & Steps $O_1$ & Steps $O_2$ \\
			\midrule
			\multirow{2}{*}{$3^2$} & PPO & 2.5$\pm0.01$  & 33.43$\pm4.1$  & 787.03$\pm31.8$ \\
			& PPO + \method & \textbf{2.30}$\pm0.03$ & \textbf{18.940}$\pm1.1$ & \textbf{143.550}$\pm1.1$ \\
			\midrule
			\multirow{2}{*}{$5^2$} & PPO & 4.2$\pm0.01$ & 62.7$\pm9.7$ & S/B  \\
			& PPO + \method & \textbf{2.1}$\pm0.05$ & \textbf{59.50}$\pm14.3$ & \textbf{8057.5}$\pm121.4$  \\
			\midrule
                \multirow{2}{*}{$8^2$} & PPO & 11.2$\pm$0.03 & 16801.8$\pm$2144.0 & S/B  \\
			& PPO + \method & \textbf{6.94}$\pm$0.07 & \textbf{4149.6}$\pm$1743.1 & \textbf{386.2}$\pm$80.5  \\
                \midrule
                \multirow{2}{*}{$10^2$} & PPO & 10.9$\pm$0.01 & 	S/B & 29023.6$\pm$976.4  \\
			& PPO + \method & \textbf{5.3}$\pm$0.10 & \textbf{21272.8}$\pm$3579.0 & \textbf{570.3}$\pm$52.2  \\
			\bottomrule
		\end{tabular}
		}
		\end{minipage}
		\vspace{-3mm}
		\end{table}

\vspace{-2mm}
\section{Conclusion} 
\label{sec:conclusion}
\vspace{-2mm}
We introduced \method, a reinforcement learning framework that synthesizes control policies for tasks specified as \HyperLTL formulas over MDPs with unknown transitions. 
By leveraging the expressiveness of hyperproperties, \method can handle complex multi-agent objectives and relational constraints, including safe planning and undecidable problems like PCP. 
Our quantitative semantics automatically derives robustness values from the specification and learns control policies that maximize the probability of satisfying the given \HyperLTL property. 
Implemented with off-the-shelf RL algorithms, \method demonstrates superior performance over comparable baselines across a set of diverse case studies, especially in scenarios where reward shaping is non-trivial.

\vspace{-2mm}
\paragraph{Limitations.}
\method currently does not support fully decentralized control policies, where each agent learns its local 
policy without access to global observations and rewards. 
Second, it is possible that environment states are not fully observable. That is, instead of an MDP, 
the environment is a partially observable MDP (POMDP).
We believe these limitations open up exciting future work and can be addressed by extending 
\method.

\vspace{-2mm}
\paragraph{Societal impacts.} 
This work marks a significant step towards extending MARL to complex tasks with relational 
requirements.
By enabling the learning of policies under interdependent objectives, our approach broadens the applicability of MARL to critical domains such as disaster response, swarm drones, smart home automation, and industrial resource management, where coordination among agents is essential for success.

\bibliographystyle{plainnat}
\bibliography{bibliography,references}

\newcommand{\SortNoop}[1]{}
\begin{thebibliography}{47}
\providecommand{\natexlab}[1]{#1}
\providecommand{\url}[1]{\texttt{#1}}
\expandafter\ifx\csname urlstyle\endcsname\relax
  \providecommand{\doi}[1]{doi: #1}\else
  \providecommand{\doi}{doi: \begingroup \urlstyle{rm}\Url}\fi

\bibitem[Agarwal et~al.(2019)Agarwal, Jiang, Kakade, and
  Sun]{agarwal2019reinforcement}
Alekh Agarwal, Nan Jiang, Sham~M Kakade, and Wen Sun.
\newblock Reinforcement learning: Theory and algorithms.
\newblock \emph{CS Dept., UW Seattle, Seattle, WA, USA, Tech. Rep},
  32:\penalty0 96, 2019.

\bibitem[Aksaray et~al.(2016)Aksaray, Jones, Kong, Schwager, and
  Belta]{aksaray2016q}
Derya Aksaray, Austin Jones, Zhaodan Kong, Mac Schwager, and Calin Belta.
\newblock Q-learning for robust satisfaction of signal temporal logic
  specifications.
\newblock In \emph{2016 IEEE 55th Conference on Decision and Control (CDC)},
  pages 6565--6570. IEEE, 2016.

\bibitem[Alshiekh et~al.(2018)Alshiekh, Bloem, Ehlers, K{\"o}nighofer, Niekum,
  and Topcu]{alshiekh2018safe}
Mohammed Alshiekh, Roderick Bloem, R{\"u}diger Ehlers, Bettina K{\"o}nighofer,
  Scott Niekum, and Ufuk Topcu.
\newblock Safe reinforcement learning via shielding.
\newblock In \emph{Proceedings of the AAAI conference on artificial
  intelligence}, volume~32, 2018.

\bibitem[Barron and Ishii(1989)]{barron1989bellman}
EN~Barron and H~Ishii.
\newblock The bellman equation for minimizing the maximum cost.
\newblock \emph{NONLINEAR ANAL. THEORY METHODS APPLIC.}, 13\penalty0
  (9):\penalty0 1067--1090, 1989.

\bibitem[Barthe et~al.(2011)Barthe, D'argenio, and Rezk]{barthe2011secure}
Gilles Barthe, Pedro~R D'argenio, and Tamara Rezk.
\newblock Secure information flow by self-composition.
\newblock \emph{Mathematical Structures in Computer Science}, 21\penalty0
  (6):\penalty0 1207--1252, 2011.

\bibitem[Bellman(1952)]{bellman1952theory}
Richard Bellman.
\newblock On the theory of dynamic programming.
\newblock \emph{Proceedings of the national Academy of Sciences}, 38\penalty0
  (8):\penalty0 716--719, 1952.

\bibitem[Beutner and Finkbeiner(2023)]{lmcs:9209}
Raven Beutner and Bernd Finkbeiner.
\newblock Hyperatl*: A logic for hyperproperties in multi-agent systems.
\newblock \emph{Logical Methods in Computer Science}, Volume 19, Issue
  2:\penalty0 13, May 2023.
\newblock ISSN 1860-5974.

\bibitem[Bonakdarpour and Finkbeiner(2018)]{bf18}
B.~Bonakdarpour and B.~Finkbeiner.
\newblock The complexity of monitoring hyperproperties.
\newblock In \emph{Proceedings of the 31st {IEEE} Computer Security Foundations
  Symposium {CSF}}, pages 162--174, 2018.

\bibitem[Bonakdarpour and Sheinvald(2023)]{bs23}
B.~Bonakdarpour and S.~Sheinvald.
\newblock Finite-word hyperlanguages.
\newblock \emph{Information and Computation}, 295:\penalty0 104944, 2023.

\bibitem[Brafman et~al.(2018)Brafman, De~Giacomo, and Patrizi]{brafman2018ltlf}
Ronen Brafman, Giuseppe De~Giacomo, and Fabio Patrizi.
\newblock Ltlf/ldlf non-markovian rewards.
\newblock In \emph{Proceedings of the AAAI conference on artificial
  intelligence}, volume~32, 2018.

\bibitem[Brett et~al.(2017)Brett, Siddique, and Bonakdarpour]{bsb17}
N.~Brett, U.~Siddique, and B.~Bonakdarpour.
\newblock Rewriting-based runtime verification for alternation-free
  {H}yper{LTL}.
\newblock In \emph{Proceedings of the 23rd International Conference on Tools
  and Algorithms for the Construction and Analysis of Systems (TACAS)}, pages
  77--93, 2017.

\bibitem[Clarkson and Schneider(2010)]{cs10}
M.~R. Clarkson and F.~B. Schneider.
\newblock Hyperproperties.
\newblock \emph{Journal of Computer Security}, 18\penalty0 (6):\penalty0
  1157--1210, 2010.

\bibitem[Clarkson et~al.(2014)Clarkson, Finkbeiner, Koleini, Micinski, Rabe,
  and S{\'{a}}nchez]{cfkmrs14}
M.~R. Clarkson, B.~Finkbeiner, M.~Koleini, K.~K. Micinski, M.~N. Rabe, and
  C.~S{\'{a}}nchez.
\newblock Temporal logics for hyperproperties.
\newblock In \emph{Proceedings of the 3rd Conference on Principles of Security
  and Trust {POST}}, pages 265--284, 2014.

\bibitem[De~Giacomo et~al.(2019)De~Giacomo, Iocchi, Favorito, and
  Patrizi]{de2019foundations}
Giuseppe De~Giacomo, Luca Iocchi, Marco Favorito, and Fabio Patrizi.
\newblock Foundations for restraining bolts: Reinforcement learning with
  ltlf/ldlf restraining specifications.
\newblock In \emph{Proceedings of the international conference on automated
  planning and scheduling}, volume~29, pages 128--136, 2019.

\bibitem[De~Hauwere et~al.(2010)De~Hauwere, Vrancx, and
  Now\'{e}]{10.5555/1838206.1838301}
Yann-Micha\"{e}l De~Hauwere, Peter Vrancx, and Ann Now\'{e}.
\newblock Learning multi-agent state space representations.
\newblock In \emph{Proceedings of the 9th International Conference on
  Autonomous Agents and Multiagent Systems: Volume 1 - Volume 1}, AAMAS '10,
  page 715–722, Richland, SC, 2010. International Foundation for Autonomous
  Agents and Multiagent Systems.
\newblock ISBN 9780982657119.

\bibitem[ElSayed-Aly and
  Feng(2022)]{elsayedaly2022logicbasedrewardshapingmultiagent}
Ingy ElSayed-Aly and Lu~Feng.
\newblock Logic-based reward shaping for multi-agent reinforcement learning,
  2022.

\bibitem[ElSayed-Aly et~al.(2021)ElSayed-Aly, Bharadwaj, Amato, Ehlers, Topcu,
  and Feng]{elsayed2021safe}
Ingy ElSayed-Aly, Suda Bharadwaj, Christopher Amato, R\"{u}diger Ehlers, Ufuk
  Topcu, and Lu~Feng.
\newblock Safe multi-agent reinforcement learning via shielding.
\newblock In \emph{Proceedings of the 20th International Conference on
  Autonomous Agents and MultiAgent Systems}, AAMAS '21, page 483–491,
  Richland, SC, 2021. International Foundation for Autonomous Agents and
  Multiagent Systems.
\newblock ISBN 9781450383073.

\bibitem[Fan(1953)]{fan1953minimax}
Ky~Fan.
\newblock Minimax theorems.
\newblock \emph{Proceedings of the National Academy of Sciences}, 39\penalty0
  (1):\penalty0 42--47, 1953.

\bibitem[Finkbeiner and Hahn(2016)]{fh16}
B.~Finkbeiner and C.~Hahn.
\newblock Deciding hyperproperties.
\newblock In \emph{Proceedings of the 27th International Conference on
  Concurrency Theory (CONCUR)}, pages 13:1--13:14, 2016.

\bibitem[Hammond et~al.(2021)Hammond, Abate, Gutierrez, and
  Wooldridge]{hammond2021multiagentreinforcementlearningtemporal}
Lewis Hammond, Alessandro Abate, Julian Gutierrez, and Michael Wooldridge.
\newblock Multi-agent reinforcement learning with temporal logic
  specifications, 2021.

\bibitem[Hasanbeig et~al.(2018)Hasanbeig, Abate, and
  Kroening]{hasanbeig2018logically}
Mohammadhosein Hasanbeig, Alessandro Abate, and Daniel Kroening.
\newblock Logically-constrained reinforcement learning.
\newblock \emph{arXiv preprint arXiv:1801.08099}, 2018.

\bibitem[Hasanbeig et~al.(2019)Hasanbeig, Kantaros, Abate, Kroening, Pappas,
  and Lee]{hasanbeig2019reinforcement}
Mohammadhosein Hasanbeig, Yiannis Kantaros, Alessandro Abate, Daniel Kroening,
  George~J Pappas, and Insup Lee.
\newblock Reinforcement learning for temporal logic control synthesis with
  probabilistic satisfaction guarantees.
\newblock In \emph{2019 IEEE 58th conference on decision and control (CDC)},
  pages 5338--5343. IEEE, 2019.

\bibitem[Jothimurugan et~al.(2019)Jothimurugan, Alur, and
  Bastani]{jothimurugan2019composable}
Kishor Jothimurugan, Rajeev Alur, and Osbert Bastani.
\newblock A composable specification language for reinforcement learning tasks.
\newblock \emph{Advances in Neural Information Processing Systems}, 32, 2019.

\bibitem[Jothimurugan et~al.(2021)Jothimurugan, Bansal, Bastani, and
  Alur]{jothimurugan2021compositional}
Kishor Jothimurugan, Suguman Bansal, Osbert Bastani, and Rajeev Alur.
\newblock Compositional reinforcement learning from logical specifications.
\newblock \emph{Advances in Neural Information Processing Systems},
  34:\penalty0 10026--10039, 2021.

\bibitem[Jothimurugan et~al.(2022)Jothimurugan, Bansal, Bastani, and
  Alur]{jothimurugan2022specification}
Kishor Jothimurugan, Suguman Bansal, Osbert Bastani, and Rajeev Alur.
\newblock Specification-guided learning of nash equilibria with high social
  welfare.
\newblock In \emph{International Conference on Computer Aided Verification},
  pages 343--363. Springer, 2022.

\bibitem[K{\"o}nighofer et~al.(2020)K{\"o}nighofer, Lorber, Jansen, and
  Bloem]{konighofer2020shield}
Bettina K{\"o}nighofer, Florian Lorber, Nils Jansen, and Roderick Bloem.
\newblock Shield synthesis for reinforcement learning.
\newblock In \emph{Leveraging Applications of Formal Methods, Verification and
  Validation: Verification Principles: 9th International Symposium on
  Leveraging Applications of Formal Methods, ISoLA 2020, Rhodes, Greece,
  October 20--30, 2020, Proceedings, Part I 9}, pages 290--306. Springer, 2020.

\bibitem[K{\"o}nighofer et~al.(2023)K{\"o}nighofer, Rudolf, Palmisano, Tappler,
  and Bloem]{konighofer2023online}
Bettina K{\"o}nighofer, Julian Rudolf, Alexander Palmisano, Martin Tappler, and
  Roderick Bloem.
\newblock Online shielding for reinforcement learning.
\newblock \emph{Innovations in Systems and Software Engineering}, 19\penalty0
  (4):\penalty0 379--394, 2023.

\bibitem[Kuo et~al.(2020)Kuo, Katz, and Barbu]{kuo2020encoding}
Yen-Ling Kuo, Boris Katz, and Andrei Barbu.
\newblock Encoding formulas as deep networks: Reinforcement learning for
  zero-shot execution of ltl formulas.
\newblock In \emph{2020 IEEE/RSJ International Conference on Intelligent Robots
  and Systems (IROS)}, pages 5604--5610. IEEE, 2020.

\bibitem[Kupferman and Vardi(1999)]{KupfermanV99}
Orna Kupferman and Moshe~Y. Vardi.
\newblock Model checking of safety properties.
\newblock In \emph{International Conference on Computer Aided Verification,
  {CAV} 1999}, 1999.

\bibitem[Le et~al.(2024)Le, Wagner, Witzman, Rabinovich, and
  Ong]{NEURIPS2024_d4857f72}
Xuan-Bach Le, Dominik Wagner, Leon Witzman, Alexander Rabinovich, and Luke Ong.
\newblock Reinforcement learning with {LTL} and
  \${\textbackslash}omega\$-regular objectives via optimality-preserving
  translation to average rewards.
\newblock In \emph{The Thirty-eighth Annual Conference on Neural Information
  Processing Systems}, 2024.

\bibitem[León and Belardinelli(2020)]{leon2020extendedmarkovgameslearn}
Borja~G. León and Francesco Belardinelli.
\newblock Extended markov games to learn multiple tasks in multi-agent
  reinforcement learning, 2020.

\bibitem[Li et~al.(2017)Li, Vasile, and Belta]{li2017reinforcement}
Xiao Li, Cristian-Ioan Vasile, and Calin Belta.
\newblock Reinforcement learning with temporal logic rewards.
\newblock In \emph{2017 IEEE/RSJ International Conference on Intelligent Robots
  and Systems (IROS)}, pages 3834--3839. IEEE, 2017.

\bibitem[Liu et~al.(2024)Liu, Cong, Chen, Jin, and
  Zhu]{liu2024guidingmultiagentmultitaskreinforcement}
Chanjuan Liu, Jinmiao Cong, Bingcai Chen, Yaochu Jin, and Enqiang Zhu.
\newblock Guiding multi-agent multi-task reinforcement learning by a
  hierarchical framework with logical reward shaping, 2024.

\bibitem[Melcer et~al.(2022)Melcer, Amato, and Tripakis]{melcer2022shield}
Daniel Melcer, Christopher Amato, and Stavros Tripakis.
\newblock Shield decentralization for safe multi-agent reinforcement learning.
\newblock \emph{Advances in Neural Information Processing Systems},
  35:\penalty0 13367--13379, 2022.

\bibitem[Melo and Veloso(2009)]{10.5555/1558109.1558118}
Francisco~S. Melo and Manuela Veloso.
\newblock Learning of coordination: exploiting sparse interactions in
  multiagent systems.
\newblock In \emph{Proceedings of The 8th International Conference on
  Autonomous Agents and Multiagent Systems - Volume 2}, AAMAS '09, page
  773–780, Richland, SC, 2009. International Foundation for Autonomous Agents
  and Multiagent Systems.
\newblock ISBN 9780981738178.

\bibitem[Mnih et~al.(2015)Mnih, Kavukcuoglu, Silver, Rusu, Veness, Bellemare,
  Graves, Riedmiller, Fidjeland, Ostrovski, Petersen, Beattie, Sadik,
  Antonoglou, King, Kumaran, Wierstra, Legg, and
  Hassabis]{mnih_human-level_2015}
Volodymyr Mnih, Koray Kavukcuoglu, David Silver, Andrei~A. Rusu, Joel Veness,
  Marc~G. Bellemare, Alex Graves, Martin Riedmiller, Andreas~K. Fidjeland,
  Georg Ostrovski, Stig Petersen, Charles Beattie, Amir Sadik, Ioannis
  Antonoglou, Helen King, Dharshan Kumaran, Daan Wierstra, Shane Legg, and
  Demis Hassabis.
\newblock Human-level control through deep reinforcement learning.
\newblock \emph{Nature}, 518\penalty0 (7540):\penalty0 529--533, February 2015.
\newblock ISSN 1476-4687.

\bibitem[Pnueli(1977)]{p77}
A.~Pnueli.
\newblock The temporal logic of programs.
\newblock In \emph{Symposium on Foundations of Computer Science (FOCS)}, pages
  46--57, 1977.

\bibitem[Raffin et~al.(2021)Raffin, Hill, Gleave, Kanervisto, Ernestus, and
  Dormann]{stable-baselines3}
Antonin Raffin, Ashley Hill, Adam Gleave, Anssi Kanervisto, Maximilian
  Ernestus, and Noah Dormann.
\newblock Stable-baselines3: Reliable reinforcement learning implementations.
\newblock \emph{Journal of Machine Learning Research}, 22\penalty0
  (268):\penalty0 1--8, 2021.

\bibitem[Schulman et~al.(2017)Schulman, Wolski, Dhariwal, Radford, and
  Klimov]{schulman2017proximalpolicyoptimizationalgorithms}
John Schulman, Filip Wolski, Prafulla Dhariwal, Alec Radford, and Oleg Klimov.
\newblock Proximal policy optimization algorithms, 2017.

\bibitem[Skolem(1920)]{skolem1920}
Thoralf Skolem.
\newblock Logisch-kombinatorische untersuchungen über die erfüllbarkeit oder
  beweisbarkeit mathematischer sätze nebst einem theoreme über dichte mengen.
\newblock \emph{Videnskapsselskapets Skrifter, I. Matematisk-naturvidenskabelig
  Klasse}, 1920.

\bibitem[Thomas(1988)]{thomas1988safety}
Wolfgang Thomas.
\newblock Safety-and liveness-properties in propositional temporal logic:
  characterizations and decidability.
\newblock \emph{Banach Center Publications}, 21\penalty0 (1):\penalty0
  403--417, 1988.

\bibitem[Vamplew et~al.(2011)Vamplew, Dazeley, Berry, Issabekov, and
  Dekker]{vptrb2011}
Peter Vamplew, Richard Dazeley, Adam Berry, Rustam Issabekov, and Evan Dekker.
\newblock Empirical evaluation methods for multiobjective reinforcement
  learning algorithms.
\newblock \emph{Machine Learning}, 84\penalty0 (1):\penalty0 51--80, 2011.

\bibitem[Winter and Zimmermann(2024)]{wz24}
S.~Winter and M.~Zimmermann.
\newblock Tracy, traces, and transducers: Computable counterexamples and
  explanations for hyperltl model-checking, 2024.

\bibitem[Xu et~al.(2022)Xu, Rawat, Wong, Kankanhalli, and Shah]{xrwks22}
Z.~Xu, Y.~Rawat, Y.~Wong, M.~S. Kankanhalli, and M.~Shah.
\newblock Don\textquotesingle t pour cereal into coffee: Differentiable
  temporal logic for temporal action segmentation.
\newblock In \emph{Advances in Neural Information Processing Systems},
  volume~35, pages 14890--14903, 2022.

\bibitem[Xu and Topcu(2019)]{xu2019transfer}
Zhe Xu and Ufuk Topcu.
\newblock Transfer of temporal logic formulas in reinforcement learning.
\newblock In \emph{IJCAI: proceedings of the conference}, volume~28, page 4010.
  NIH Public Access, 2019.

\bibitem[Yang et~al.(2019)Yang, Sun, and Narasimhan]{NEURIPS2019_4a46fbfc}
Runzhe Yang, Xingyuan Sun, and Karthik Narasimhan.
\newblock A generalized algorithm for multi-objective reinforcement learning
  and policy adaptation.
\newblock In H.~Wallach, H.~Larochelle, A.~Beygelzimer, F.~d\textquotesingle
  Alch\'{e}-Buc, E.~Fox, and R.~Garnett, editors, \emph{Advances in Neural
  Information Processing Systems}, volume~32. Curran Associates, Inc., 2019.

\bibitem[Zun~Yuan et~al.(2019)Zun~Yuan, Hasanbeig, Abate, and
  Kroening]{zun2019modular}
Lim Zun~Yuan, Mohammadhosein Hasanbeig, Alessandro Abate, and Daniel Kroening.
\newblock Modular deep reinforcement learning with temporal logic
  specifications.
\newblock \emph{arXiv e-prints}, pages arXiv--1909, 2019.

\end{thebibliography}
\newpage

\newpage
\appendix



\section{Proofs of Theorems}\label{proof:theorems}

\subsection{Proof of~\Cref{theorem:skolem}}\label{proof:theorem:skolem}


Recall that given a \HyperLTL formula $\varphi$, $\skolemized{\varphi}$ is in the form of:
\begin{align*}
	\skolemized{\varphi} = 
	\underbrace{\exists \skolemfunc_{i}
		(\tau_{i_1}, \ldots, \tau_{i_{|\forallsof{i}|}}) }_{\text{for each } i \in 
		\existsof{}}.~
	\underbrace{\spvertund{1.5ex}\forall \tracevar_{j}.}_{\text{for each }j \in \forallsof{}}\skolemized{\psi}
\end{align*}
Assuming that $
\tupleof{\maxpolicy_i}_{i \in \existsof{}} 
\orderedunion	
\tupleof{\maxpolicy_j}_{j \in \forallsof{}}
$
is the optimal set of policies for the Skolemized $\varphi$, then for every $k \geq 0$, it maximizes the probability of converging to $\rbvalue_{\mathit{max}}$ for the inner LTL sub-formula $\skolemized{\psi}$: 
\[
\prob
\Big[
\rbvalue
\Big(
	\zip
	\big(
	\tupleof{\traceof{\trajfromto{\traj_i}{0}{k_i}}}~
	\orderedunion
	\tupleof{\traceof{\trajfromto{\traj_j}{0}{k_j}}}
	\big), 
	\skolemized{\psi}
\Big)
\Big]_{{i \in \existsof{}},{j \in \forallsof{}}}
\converge \rbvalue_{\mathit{max}}
\]
%
That is, on any step $k$, 
the zipped path derives a set of paths
that instantiate each path variables in the original $\varphi$ as follows:
\begin{itemize}
	\item For all universal quantified paths $\tau_j$, where $j \in \forallsof{}$:
	\begin{align*}
	\bracketof{ \tracevar_j \mapsto \traceof{\trajfromto{\traj_j}{0}{k}} } 
	\end{align*}
	\item For all existential quantified paths $\tau_i$, where $i \in \existsof{}$:
	\begin{align*}
	\bracketof{ \tracevar_{i} \mapsto \skolemfunc_{i}\big(
	\traceof{\trajfromto{\traj_{i_1}}{0}{k}}, 
	\ldots, 
	\traceof{\traj_{i_\sizeof{\forallsof{i}} \bracketof{0:k} }}
	\big) }, \text{ and }~~~~~~~~~~~~~~~~~~~\\ 	
	\text{  for each } \ell \in \{1,\dots,\sizeof{\forallsof{i}}\},
		\text { if }
		i_\ell = j
		\text { then  }
		\traceof{\trajfromto{\traj_{i_\ell}}{0}{k}} 
		=  
		\traceof{\trajfromto{\traj_j}{0}{k}}   
\end{align*}
\end{itemize}
This instaintiation guarantees that the probability of satisfying ${\varphi}$ is maximized up to step $k$. 
%
Next, the zipped path proceeds to step $k+1$ with an optimal action
given by its corresponding optimal policy
$
\tupleof{\maxpolicy_i}_{i \in \existsof{}} 
\orderedunion	
\tupleof{\maxpolicy_j}_{j \in \forallsof{}}. 
$
That is,
the same path mappings of each universally quantified path variable $\traj_j$ 
and existentially quantified Skolem function witnesses $\skolemfunc_{i}$
holds for all $k \geq 0$.
Finally, for the original \HyperLTL formula
$
\varphi = 
\quant_1 \hptracevar_1. \quant_2 \hptracevar_2. \ldots \quant_n. \hptracevar_n.~ \psi, 
$
by instantiating the trace variables in the same fashion for all $\tau_i$ of $i \in \existsof{}$ and all $\tau_j$ of $j \in \forallsof{}$, 
the same optimality immediately follows, 
because the optimal convergence to $\rbvalue_{\mathit{max}}$ means maximizing the probability of satisfying the inner LTL sub-formula.  
That is, the optimal set of paths (derived from the zipped path) also 
optimizes the satisfaction of original $\varphi$ for all steps $k \geq 0$.
%
To this end, we proved that, 
an optimal tuple of policies
$\tupleof{\maxpolicy_i}_{i \in \existsof{}} 
\orderedunion	
\tupleof{\maxpolicy_j}_{j \in \forallsof{}}$
for $\skolemized{\varphi}$ 
is also an optimal set of policies for $\varphi$, 
that optimizes the probability of satisfying $\varphi$ in $\MDP$.

\subsection{Proof of~\Cref{theorem:Qlearning}}\label{proof:theorem:qlearning}

%
%

Bellman's Principle of Optimality~\cite{bellman1952theory}
states that 
``\textit{for an optimal policy, no matter what the initial decision is, the remaining decisions must constitute an optimal policy with regard to the state resulting from the initial decision}''.
An extended lemma (proved in~\cite{agarwal2019reinforcement}) 
states that for a discounted MDP, 
there exists an {\em optimal} policy, 
denoted as $\optnnpolicy$, such that 
for all $(\state, \action) \in \states \times \actions$, 
there exists a maximum Q-value achieved by $\optnnpolicy$ 
(denoted as $\Qvalue^{\optnnpolicy}$) as introduced in~\Cref{eq:optQvalue}:
\[
\Qvalue^{\optnnpolicy}(\state, \action)  
	\definedas
	\underset{\nnpolicy}{\maximum}~
	\Qvalue^{\nnpolicy}(\state, \action)
\]
Let us denote $\nnpolicyprob{\action}{\state}$ as $\nnpolicy$'s decision to take action $\action$ on state $\state$.
%
Our goal is to find an optimized $\nnpolicy$,
such that:
\[
\underset{\nnpolicy}{\maximum}~
\sum_{\state' \in \states}
{\trans(\state, \action, \state')}
\bigg[
{\rewardof{\state}{\action}}
+
{\discount}
\sum_{\action \in \actions}
{
	\nnpolicyprob{\action}{\state} 
	\expectedrewardof{\state'}{\action'}
}
\bigg]
\]
where ${\trans(\state, \action, \state')}$ is the one-step transition probability, 
${\rewardof{\state}{\action}}$ is the reward of taking action $\action$ on state $\state$,
${\discount}$ is a {discount} selected factor, 
and $\expectedreward$ is the expected reward of keep taking an action $\action'$ on a state $\state'$. 
(as defined in~\Cref{sec:step3}).
Recall that in~\Cref{sec:quantsemantics}, 
our immediate reward ${\rewardof{\state}{\action}}$
is associated with a robustness value of a finite prefix by 
evaluation only up to the current seen state $\state$ 
(i.e., independent from the unseen $\state'$ after taking action $\action$).
As a result, maximizing $\nnpolicy$ do not depend on ${\rewardof{\state}{\action}}$, 
so the previous optimization problem is equivalent to:
\[
{\rewardof{\state}{\action}}
+
\underset{\nnpolicy}{\maximum}~
\Big[
\sum_{\state' \in \states}
\trans(\state, \action, \state')
~{\discount}
\sum_{\action \in \actions}
{
	\nnpolicyprob{\action}{\state}
	\expectedrewardof{\state'}{\action'}
}
\Big]
\]
Let us now only focus on the optimization part 
(i.e., the right side of the plus operator in the above formula).
Based on the definition of expected reward defined in~\Cref{sec:step3}, 
the above formula shows that 
an optimal $\optnnpolicy$ is more likely to the take action $\action$ 
(by the learned $\nnpolicyprob{\action}{\state}$)
that has higher probability (decided by $\trans(\state, \action, \state')$)
to transit to an unseen state $\state'$ that lead to higher expected value 
(estimated by $\expectedrewardof{\state'}{\action'}$).
That is, given a state $\state$, $\optnnpolicy$ outputs an optimal action $\action$, such that:
\begin{align*}
{\discount}
\expectedreward_{\state' \sim \trans(\state, \action, \cdot))}
\big[
\reward(\state', \action')
\big]^{\optnnpolicy},
\end{align*}
which, intuitively, represents an optimal {\em one-step} look-ahead.
%
Now, to connect the above formula with the reward function defined in~\Cref{sec:step3}, we have:
\[
{\discount}
\expectedreward_{\state' \sim \trans(\state, \action, \cdot))}
\bigg[
	\rbvalue
	\Big(
	\zip
	\big(
	\tupleof{\traceof{\trajfromto{\traj_i}{0}{k}}}
	\orderedunion
	\tupleof{\traceof{\trajfromto{\traj_j}{0}{k}}}
	\big), 
	\skolemized{\psi}
	\Big)
\bigg]^{\nnpolicy\optimal}_{{i \in \existsof{}},{j \in \forallsof{}}},
\]
where $\state$ is the state on the $k$-th step of the zipped path. 
Recall that $\rbvalue$ is constructed using min-max approach as presented in~\Cref{fig:quantsemantics}, 
so the optimal outcome of $\rbvalue$ can be derived from the 
{\em Minimax Lemma}~\cite{fan1953minimax}.
That is, if each path always considers the ``worst-possible'' scenario 
that other paths will act during learning, it leads to a set of optimal policy among all paths.
Hence, $\rbvalue$ is guaranteed optimal for all steps $k\geq0$, 
which implies the maximum probability of the following: 
$$
{\prob}
\Big[
\rbvalue
\Big(
\zip
\big(
\tupleof{\traceof{\trajfromto{\traj_i}{0}{k_i} \sim \distributionof{\policy_{{i}}}}\big}
\orderedunion
\tupleof{\traceof{\trajfromto{\traj_j}{0}{k_j} \sim \distributionof{\policy_{{j}}}}\big}
\big),
\skolemized{\psi}
\Big)
\converge 
\rbvalue_{\mathit{max}}
\Big]_{{i \in \existsof{}},{j \in \forallsof{}}}
$$
To this end,
we prove that the action chose by $\optnnpolicy$ achieves the maximum expected value $\expectedreward$ for all $(\state, \action) \in \states \times \actions$. 
Finally, 
$\tupleof{\skolemfunc_{i}}_{i \in \existsof{}}$
and
$\tupleof{\maxpolicy_{j}}_{j \in \forallsof{}}$ 
can be inductively constructed from $\optnnpolicy$
(as we elaborated in~\Cref{sec:step3}), 
which is a policies set that optimizes the satisfaction of $\skolemized{\varphi}$.

\section{Additional Algorithmic Details of HypRL}



\paragraph{Expected Reward of a Zipped Trace.}\label{appendix:expected_reward}

The {\em expected reward}
of a state $\state_\leafindex$ after taking an action $\action_\leafindex$ with {\em discount factor}
%
$\discount \in [0,1]$ is:
\newcommand{\eqexpectedreward}{
	\expectedreward(\state_\leafindex, \action_\leafindex)
	\definedas
	\sum_{t=0}^{\infty} 
	\discount^t 
	\times 
	\reward(\state_{\leafindex+t+1}, \action_{\leafindex+t+1})
}
\begin{equation*}
\eqexpectedreward
\end{equation*}
Intuitively, $\expectedreward(\state_\leafindex, \action_\leafindex)$
evaluates ``how good'' of choosing $\action_\leafindex$ on $\state_\leafindex$ in infinite time steps, where each step is \emph{discounted} by $\discount$.
We address that, $\discount$ is often chosen based on the optimization goal. 
For example, for short-term tasks, a lower 
$\discount$ is preferred because the expected robustness value focuses more on immediate 
$\rbvalue$ value.
We report the setup of these hyperparameters in~\Cref{appendix:experiments}.


\section{A Comprehensive Example of HypRL Algorithm}\label{appendix:running-example}

In this section, we present a comprehensive head-to-toe example that aligns with the algorithmic details  presented in~\Cref{sec:rlhp}.

{
\paragraph{Formula Skolemization.}
In~\Cref{sec:problem}, we presented the \HyperLTL formula for our wildfire example as follows:
\begin{center}
	\small
	$
	\textit{Requirements}:
	~\varphi_{\textsf{Rescue}}  \triangleq \forall \pathone. \exists \pathtwo.
	(\psi_{\motifputout} \land \psi_{\motifsave} \land \psi_{\motifdist}  \land \psi_{\motifsafe})
	$
		\begin{align*}
			O_1&: 
			~\psi_{\motifputout}  \definedas
			\F ({\locmark i}_{\pathone})  \land 
			\F ({\locmark f}_{\pathone}) \land 
			\F ({\locmark c}_{\pathone})
			~~
			&C_1&: 
			~\psi_{\motifdist}  \definedas
			\G (| \motifloc_{\pathone} - \motifloc_{\pathtwo} | < 3)
			\\
			O_2&:
			~\psi_{\motifsave}  \definedas 
			\F ({\locmark g}_{\pathtwo})
			\land
			\F ({\locmark f}_{\pathtwo})
			&C_2&:
			~\psi_{\motifsafe}  \definedas 
			(\neg {\locmark i}_{\pathtwo} \U {\locmark i}_{\pathone}) \land
			(\neg {\locmark f}_{\pathtwo} \U {\locmark f}_{\pathone}) \land
			(\neg {\locmark c}_{\pathtwo} \U {\locmark c}_{\pathone})
		\end{align*}
\end{center} 
Following the illustration in~\Cref{sec:skolemization}, the Skolemized form of the formula $\varphi_{\textsf{Rescue}}$ is as follows: 
		\begin{align*}
		\small
			\skolemized{\varphi_{\textsf{Rescue}}} \definedas
			\exists &\skolemfunc_{2}(\tracevar_{1}).
			\forall\tracevar_{1}.~ \\
			& \skolemized{\psi_{\motifputout}} \land
			\skolemized{\psi_{\motifsave}} \land
			\skolemized{\psi_{\motifdist}} \land 
			\skolemized{\psi_{\motifsafe}}
		\end{align*}
		\begin{align*}
		\small
			 \skolemized{\psi_{\motifputout}} 
			&\definedas
			\F ({\locmark i}_{\pathone})  \land 
			\F ({\locmark f}_{\pathone}) \land 
			\F ({\locmark c}_{\pathone})\\
			\skolemized{\psi_{\motifsave}} 
			&\definedas \F ({\locmark g}_{\skolemfunc_{2}}) \land \F ({\locmark f}_{\skolemfunc_{2}})
			\\
			\skolemized{\psi_{\motifdist}} 
			&\definedas
			\G (| \motifloc_{\pathone} - \motifloc_{\skolemfunc_{2}} | < 3)
			\\
			\skolemized{\psi_{\motifsafe}} 
				&\definedas 
				(\neg {\locmark i}_{\skolemfunc_{2}} \U {\locmark i}_{\pathone})
				\land (\neg {\locmark f}_{\skolemfunc_{2}} \U {\locmark f}_{\pathone})
				(\neg {\locmark c}_{\skolemfunc_{2}} \U {\locmark c}_{\pathone})
		\end{align*}
	In this context, there are two quantifiers, $\quant_2=\exists$ so $\existsof{}=\{2\}$, and $\quant_1=\forall$ so $\forallsof{}=\{1\}$. 
	Besides, all propositions of the existential quantified path variable (originally subscripted by $\tau_2$) average now subscripted by $\skolemfunc_2$ in the Skolemized form.
	Furthermore, for each inner LTL sub-formula, the propositions subscripted with $\tau_2$ is substituted with propositions subscripted with $\skolemfunc_2$. 
}

{
}

{
\paragraph{Robustness Optimization.}
Continuing with the Skolemized formula:
\begin{align*}
	\exists\skolemfunc_{2}(\tracevar_{1}).
	\forall\tracevar_{1}.~
	\skolemized{\psi_{\motifputout}} \land
	\skolemized{\psi_{\motifsave}} \land  
	\skolemized{\psi_{\motifdist}} \land 
	\skolemized{\psi_{\motifsafe}}
\end{align*}
Our goal is to optimize the following:

\begin{align*}
	&
	\tupleof{\maxpolicy_1, \maxpolicy_2}
	\in 
	\underset{
		\tupleof{
			\policy_{1}, \policy_{2}
		}
	}
	{{\arg\max}}~  	
	\prob
	\Big[
	\rbvalue
	\big(
	\zip
	(
		\tupleof{
			\traceof{\trajfromto{\traj_1}{0}{k_1}\sim \distributionof{\policy_{1}} },
			\traceof{\trajfromto{\traj_2}{0}{k_2}\sim \distributionof{\policy_{2}}}
		}
	),\\
	&~~~~~~~~~~~~~~~~~~~~
	\skolemized{
	\psi_{\motifputout} \land 
	\psi_{\motifsave} \land 
	\psi_{\motifdist} \land 
	\psi_{\motifsafe}}
	\big) 
	\converge
	\rbvalue_{\mathit{max}}
	\Big]
\end{align*}
Notice that, by applying $\orderedunion$, the order of the quantified traces are in the right order with respect to the original \HyperLTL formula $\varphi$.
}

\section{Implementation and Experimental Details}\label{appendix:experiments}

\subsection{From \hyperltl Formulas to Robustness Functions}
In our implementation, we use the theory developed in \Cref{sec:skolemization} to obtain the Skolemized form of the formula, denoted $\skolemized{\varphi}$. 
We then generate the corresponding robustness function using the approach presented in \Cref{sec:quantsemantics} (see~\cite{li2017reinforcement} for details on translating temporal properties into robustness functions). 
We now elaborate the following term from the quantitative semantics introduced in~\Cref{fig:quantsemantics}:
\[
    \rbvalueof{\traceof{\trajfromto{\traj}{\ell}{k}}, 
	\predicatefuncof{ {\footnotesize \labelsof{\state_\ell}<\constant}} } ~{=}~ \constant - \predicatefuncof{\labelsof{\state_\ell}}
\]
This equation lies at the core of transitioning from logical properties (which yield Boolean outputs) to robustness functions (which produce real-valued outputs). 
To apply this transformation, we must define the function $f$ and the constant $c$ for each property considered in this paper. 
We now provide details explanations on how we define such functions and constants for each formula introduced in this paper. 
\newcommand{\casesrl}{\text{SRL}}
\begin{itemize}
    \item The \HyperLTL formula defined for safe reinforcement learning (SRL), denoted as $\varphi_{\casesrl}$: 
    \begin{itemize}
        \item 
        For the property $\tupleof{x_{\tau_1}, y_{\tau_1}} = \tupleof{x_{G1}, y_{G1}}$, we define $f$ as a distance function and set $c=1$. 
        The distance function we have in this case study follows the well-known Manhattan distance definition. That is: 
        \[
        \texttt{Dist}(\tupleof{x_{\tau_1}, y_{\tau_1}}, \tupleof{x_{G1}, y_{G1}}) \definedas (|x_{\tau_1} - x_{G1}| + |y_{\tau_1} - y_{G1}|)
        \]
        The, by setting $c=1$, this yields the following expression:
        \[
        \texttt{Dist}(\tupleof{x_{\tau_1}, y_{\tau_1}}, \tupleof{x_{G1}, y_{G1}}) < 1,
        \]
        which expresses that the distance from the first agent (i.e., $\tau_1$) to its goal (i.e., $G_1$) is less than one, which implies that the first agent has \emph{indeed} reached the specified goal location. The corresponding equation is:
        \[
        1 - \texttt{Dist}(\tupleof{x_{\tau_1}, y_{\tau_1}}, \tupleof{x_{G1}, y_{G1}})
        \]
        The optimal value is achieved when the agent is exactly at the location of the goal position, giving $1 - 0 = 1$.
        \item For the property $\tupleof{x_{\tau_2}, y_{\tau_2}} = \tupleof{x_{G2}, y_{G2}}$, the same definition for the distance function is similarly defined. That is:
        \[
        \texttt{Dist}(\tupleof{x_{\tau_2}, y_{\tau_2}}, \tupleof{x_{G2}, y_{G2}}) < 1,
        \]
        which expresses that the distance from the second agent to its goal is less than one, so the corresponding equation is:
        \[
        1 - \texttt{Dist}(\tupleof{x_{\tau_2}, y_{\tau_2}}, \tupleof{x_{G2}, y_{G2}})
        \]
        The optimal value is achieved when the second agent is exactly at the goal position, giving $1 - 0 = 1$.
        \item For the property $\tupleof{x_{\tau_1}, y_{\tau_1}} \neq \tupleof{x_{\tau_2}, y_{\tau_2}}$, we define $f$ as a distance function and set $c = -1$. This yields the condition:
        \[
        -\texttt{Dist}(\tupleof{x_{\tau_1}, y_{\tau_1}}, \tupleof{x_{\tau_2}, y_{\tau_2}}) < -1,
        \]
        which expresses that the distance between the first and second agent (i.e., $\tau_1$ and $\tau_2$) is greater than one; i.e., the agents are not in the same location. The corresponding equation is:
        \[
        -1 + \texttt{Dist}(\tupleof{x_{\tau_1}, y_{\tau_1}}, \tupleof{x_{\tau_2}, y_{\tau_2}}).
        \]
        This equation is maximized when the two agents are far apart, and it evaluates to $-1$ when they are in same location.

    \end{itemize}

	\newcommand{\casedst}{\text{DST}}
   \item The \HyperLTL formula defined for deep sea treasure (DST), denoted as $\varphi_{\casedst}$: 
    \begin{itemize}
        \item For the properties \textit{T1}, \textit{T2}, $\cdots$, we define $f$ for each treasure \textit{Ti} a distance function between the position of the submarine driver ($\tau_i$) and the treasure location (\textit{Ti}), and set $c = 1$. This yields the condition:
        \[
        \texttt{Dist}(\tupleof{x_{\tau_1}, y_{\tau_1}}, \tupleof{x_{T1}, y_{T1}}) < 1
        \]
        which expresses that the distance from the first agent to \textit{T1} is less than one. The corresponding equation is:
        \[
        1 - \texttt{Dist}(\tupleof{x_{\tau_1}, y_{\tau_1}}, \tupleof{x_{T1}, y_{T1}})
        \]
        The same approach is applied to \textit{T2}, \textit{T3},\dots and the remaining properties.
        \item For the property $step_{\tau_2} < \delta$, we use the same approach and obtain the following equation:
            \[
            \delta - step_{\tau_2}
            \]
        \item For the property $|\textit{pos}_\traceone - \textit{pos}_\tracetwo| < 1$, we define $f$ as a distance function and set $c = 1$. This yields the condition:
        \[
        \texttt{Dist}(\tupleof{x_{\tau_1}, y_{\tau_1}}, \tupleof{x_{\tau_2}, y_{\tau_2}}) < 1,
        \]
        which expresses that the distance between the first and second agent is less than one. 
		That is, the agents are in the same location. 
		The corresponding equation is:
        \[
        1 - \texttt{Dist}(\tupleof{x_{\tau_1}, y_{\tau_1}}, \tupleof{x_{\tau_2}, y_{\tau_2}})
        \]
        This equation is maximized when the agents are in the same location.
    \end{itemize}
	\newcommand{\pcpmatch}{\texttt{Match}}
	\newcommand{\casepcp}{\text{PCP}}
    \item The formula defined for the Post Correspondence Problem (PCP), denoted as $\varphi_{\casepcp}$:
    \begin{itemize}
        \item To address the properties such as $(p_1 \leftrightarrow p_2)$, where $p$ means a sequence of alphabets (that forms a string), we define $f$ as a function of evaluating whether two strings are matching and set $c = 1$. The matching function we have in this case study is as follows:
        \[
        \pcpmatch(p_1,  p_2){=} \begin{cases}
            0   & \text{if } p_1 \text{ and } p_2 \text{ are equal,}\\
            1  &  \text{otherwise}.
            \end{cases}
        \]
        This yields the condition:
        \[
        \pcpmatch(p_1, p_2) < 1,
        \]
        which expresses whether the two compared alphabets $p_1$ and $p_2$ are equal or not. The corresponding equation is:
        \[
        1 - \pcpmatch(p_1, p_2)
        \]
        \item For the symbol \#, we are checking whether the alphabet under consideration is equal to \#. This is expressed as the logical condition $p \leftrightarrow \#$.

    \end{itemize}
	\item The \HyperLTL formula defined for the wildfire scenario, denoted as $\varphi_{\textsf{Rescue}}$:
    \begin{itemize}
        \item For the properties \textit{V1}, \textit{V2}, $\cdots$ and \textit{F1}, \textit{F2}, $\cdots$, we use the same distance function $\texttt{Dist}$ (as in SRL and DST) to calculate the distance from agents to victims and fire zones, with $c = 1$. 
		That is, the reachability of the two objectives.
        \item For the property $| \motifloc_{\pathone} - \motifloc_{\pathtwo} | < 3$, we use the distance function $\texttt{Dist}$ with $c = 3$. This constraints make sure that for the optimized strategy, the two agents are always staying within the safe communication range. 
    \end{itemize}
\end{itemize}

\subsection{Safe RL (SRL)}

\paragraph{Case Study Setup.}
	
%
In the Safe RL case study (see~\Cref{fig:gridworld}), the blue circle (denoted \agonesafe) and the orange circle (denoted \agtwosafe) are agents that aim to learn optimal policies to navigate from their initial positions to their respective targets: the blue square (\goalone) and the orange square (\goaltwo), while avoiding collisions.
The state space is represented as a tuple $\tupleof{x, y}$, and the action space is defined as $\action = \{ \textit{stay}, \textit{up}, \textit{down}, \textit{left}, \textit{right} \}$.
The following \HyperLTL formula $\varphi_{\text{SRL}}$ specifies the required objectives:
\[
\varphi_{\text{SRL}} \triangleq \forall \tronesafe . \exists \trtwosafe.~ 
\Big(\F \Tonesafe  ~\land~ \F \Ttwosafe ~\land ~  \always \collide\Big)
\]
where the subformulas are defined as:
\[
\Tonesafe \triangleq \tupleof{x_{\tau_1}, y_{\tau_1}} = \tupleof{x_{G1}, y_{G1}},
~~~~~~~~~~~
\Ttwosafe \triangleq \tupleof{x_{\tau_2}, y_{\tau_2}} = \tupleof{x_{G2}, y_{G2}},
\]
\[
\collide \triangleq \tupleof{x_{\tau_1}, y_{\tau_1}} \neq \tupleof{x_{\tau_2}, y_{\tau_2}}.
\]
Here, $\Tonesafe$ and $\Ttwosafe$ express that \agonesafe and \agtwosafe eventually reach \goalone and \goaltwo, respectively. The subformula $\collide$ ensures that the agents avoid collisions while navigating toward their goals.

\begin{figure}[t]
	
	\begin{subfigure}[t]{0.44\textwidth}
		\centering
		\scalebox{0.3}{\begin{tikzpicture}[]

\foreach \x in {0,1,...,8} {
    \foreach \y in {0,1,...,9} {
        \draw[line width=2pt] (\x,\y) rectangle ++(1,1); 
    }
}

\fill[white, draw=none] (2.02,3.02) rectangle ++(4.96,4.96);
\fill[white, draw=none] (-0.2,-0.2) rectangle ++(2.2,2.2);
\fill[white, draw=none] (4,-0.2) rectangle ++(5.3,1.2);
\fill[white, draw=none] (4,-0.2) rectangle ++(5.2,1.08);

\fill[white, draw=none] (-0.1,3.02) rectangle ++(1,0.96);
\fill[white, draw=none] (-0.1,5.02) rectangle ++(1,0.96);
\fill[white, draw=none] (-0.1,7.02) rectangle ++(1,0.96);
\fill[white, draw=none] (-0.1,9.02) rectangle ++(1.08,1.1);

\fill[white, draw=none] (8.01,0.95) rectangle ++(1.1,1.025);
\fill[white, draw=none] (8.03,3.02) rectangle ++(1.1,0.96);
\fill[white, draw=none] (8.03,5.02) rectangle ++(1.1,0.96);
\fill[white, draw=none] (8.03,7.02) rectangle ++(1.1,0.96);
\fill[white, draw=none] (8.02,9.02) rectangle ++(1.1,1.15);

\fill[white, draw=none] (2.02,9.02) rectangle ++(0.96,1.1);
\fill[white, draw=none] (4.02,9.02) rectangle ++(0.96,1.1);
\fill[white, draw=none] (6.02,9.02) rectangle ++(0.96,1.1);

\fill[white, draw=none] (4.02,0.9) rectangle ++(0.96,1.05);
\fill[white, draw=none] (6.02,0.9) rectangle ++(0.96,1.05);

\fill[white, draw=none] (-0.1,7.02) rectangle ++(1,0.96);
\fill[white, draw=none] (-0.1,9.02) rectangle ++(1.08,1.1);

\fill[orange!80, opacity=0.3] (3,3) rectangle ++(1,1); 
\fill[blue!60, opacity=0.8] (1.5,3.5) circle(0.35cm); 

\fill[cyan!20] (3,0) rectangle ++(1,1); 
\fill[orange!70, opacity=1] (0.5,2.5) circle(0.35cm); 

\end{tikzpicture}}
		\caption{ISR}
		\label{fig:isr_grid}
	\end{subfigure}
	\hfill
    \begin{subfigure}[t]{0.44\textwidth}
		\centering
		\scalebox{0.3}{\begin{tikzpicture}[]

\foreach \x in {0,1,...,16} {
    \foreach \y in {0,1,...,6} {
        \draw[line width=2pt] (\x,\y) rectangle ++(1,1); 
    }
}

\fill[white, draw=none] (-0.2,3) rectangle ++(1.2,4.2);
\fill[white, draw=none] (-0.2,-0.2) rectangle ++(1.2,4.13);
\fill[white, draw=none] (16,3) rectangle ++(1.2,4.1);
\fill[white, draw=none] (16,-0.2) rectangle ++(1.2,4.2);
\fill[white, draw=none] (2.02,2.02) rectangle ++(0.96,2.96);
\fill[white, draw=none] (4.01,2.01) rectangle ++(2.98,2.98);
\fill[white, draw=none] (11.01,2.01) rectangle ++(3.98,2.98);
\fill[white, draw=none] (8.1,2.02) rectangle ++(1.8,0.96);
\fill[white, draw=none] (8.1,4.02) rectangle ++(1.8,0.96);
\fill[white, draw=none] (7.02,-0.2) rectangle ++(4.96,1.2);
\fill[white, draw=none] (7.02,6.02) rectangle ++(4.96,1.2);

\fill[white, draw=none] (0.96,6.02) rectangle ++(1.02,1.1);
\fill[white, draw=none] (3.01,6.02) rectangle ++(0.98,1.1);
\fill[white, draw=none] (5.01,6.02) rectangle ++(0.98,1.1);

\fill[white, draw=none] (0.95,-0.1) rectangle ++(1.05,1.08);
\fill[white, draw=none] (3.01,-0.1) rectangle ++(0.98,1.01);
\fill[white, draw=none] (5.01,-0.1) rectangle ++(0.98,1.01);

\fill[white, draw=none] (15.01,6.02) rectangle ++(2.1,1.2);
\fill[white, draw=none] (13.01,6.02) rectangle ++(0.98,1.1);

\fill[white, draw=none] (15.01,-0.1) rectangle ++(1.04,1.08);
\fill[white, draw=none] (13.01,-0.1) rectangle ++(0.98,1.01);

\fill[orange!80, opacity=0.3] (0,3) rectangle ++(1,1); 
\fill[blue!60, opacity=0.8] (0.5,3.5) circle(0.35cm); 

\fill[cyan!20] (16,3) rectangle ++(1,1); 
\fill[orange!70, opacity=1] (16.5,3.5) circle(0.35cm); 

\end{tikzpicture}}
		\caption{MIT}
		\label{fig:mit_grid}
	\end{subfigure}
    \vfill
\begin{subfigure}[b]{0.44\textwidth}
    \centering
    \scalebox{0.3}{\begin{tikzpicture}[]

\foreach \x in {0,1,...,10} {
    \foreach \y in {0,1,...,8} {
        \draw[line width=2pt] (\x,\y) rectangle ++(1,1); 
    }
}

\draw[line width=1.5pt] (5,4) rectangle ++(1,1);

\fill[white, draw=none] (-0.3,1.015) rectangle ++(1.28,8.1);
\fill[white, draw=none] (10.02,5.02) rectangle ++(1.1,4.2);
\fill[white, draw=none] (10.02,-0.1) rectangle ++(1.1,4.08);
\fill[white, draw=none] (2.02,5.01) rectangle ++(1.96,2.97);
\fill[white, draw=none] (3.8,6.01) rectangle ++(1.2,1.97);
\fill[white, draw=none] (6.02,6.01) rectangle ++(2.95,1.97);
\fill[white, draw=none] (8.02,5.01) rectangle ++(0.96,1.1);
\fill[white, draw=none] (7.02,1.01) rectangle ++(1.96,2.97);
\fill[white, draw=none] (6.02,2.03) rectangle ++(1.1,0.96);
\fill[white, draw=none] (2.02,1.03) rectangle ++(1.96,2.95);
\fill[white, draw=none] (3.8,1.03) rectangle ++(1.2,1.95);

\fill[orange!80, opacity=0.3] (10,4) rectangle ++(1,1); 

\fill[blue!60, opacity=0.8] (6.5,1.5) circle(0.35cm); 

\fill[cyan!20] (5,7) rectangle ++(1,1); 

\fill[orange!70, opacity=1] (2.5,4.5) circle(0.35cm); 

\end{tikzpicture}}
    \caption{Pentagon}
    \label{fig:pen_grid}
\end{subfigure}
	\hfill
	\begin{subfigure}[b]{0.44\textwidth}
		\centering
		\scalebox{0.27}{\begin{tikzpicture}[]

\foreach \x in {0,1,...,22} {
    \foreach \y in {0,1,...,9} {
        \draw[line width=2pt] (\x,\y) rectangle ++(1,1); 
    }
}

\fill[white, draw=none] (-0.2,-0.3) rectangle ++(1.2,7.3);
\fill[white, draw=none] (0.7,-0.3) rectangle ++(16.28,2.3);
\fill[white, draw=none] (1.02,9.02) rectangle ++(23,1.3);
\fill[white, draw=none] (3.02,7.02) rectangle ++(14.95,2.1);
\fill[white, draw=none] (2.02,4.02) rectangle ++(0.97,3.95);
\fill[white, draw=none] (2.02,7.02) rectangle ++(1.1,0.95);
\fill[white, draw=none] (2.95,5.02) rectangle ++(4.03,2.1);
\fill[white, draw=none] (17.85,8.02) rectangle ++(5.4,1.1);
\fill[white, draw=none] (8.02,1.9) rectangle ++(6.95,1.07);
\fill[white, draw=none] (22.02,-0.3) rectangle ++(1.2,6.3);
\fill[white, draw=none] (22.02,7.03) rectangle ++(1.2,1.2);
\fill[white, draw=none] (19.02,2.03) rectangle ++(1.95,1.95);
\fill[white, draw=none] (19.02,5.03) rectangle ++(1.95,0.95);
\fill[white, draw=none] (17.02,2.03) rectangle ++(0.95,1.95);
\fill[white, draw=none] (16.02,6.02) rectangle ++(1.95,1.1);
\fill[white, draw=none] (16.02,6.02) rectangle ++(1.95,1.1);
\fill[white, draw=none] (6.02,6.02) rectangle ++(8.95,1.1);

\fill[white, draw=none] (2.03,1.96) rectangle ++(0.95,0.95);
\fill[white, draw=none] (4.03,1.96) rectangle ++(0.95,0.95);
\fill[white, draw=none] (6.03,1.96) rectangle ++(0.95,0.95);

\fill[white, draw=none] (4.03,4.02) rectangle ++(0.95,1.1);
\fill[white, draw=none] (6.03,4.02) rectangle ++(0.95,1.1);

\fill[white, draw=none] (8.03,2.96) rectangle ++(0.95,0.95);
\fill[white, draw=none] (10.03,2.96) rectangle ++(0.95,0.95);
\fill[white, draw=none] (12.03,2.96) rectangle ++(0.95,0.95);
\fill[white, draw=none] (14.03,2.96) rectangle ++(0.95,0.95);
\fill[white, draw=none] (16.04,1.95) rectangle ++(1.04,1.05);

\fill[white, draw=none] (8.03,5.02) rectangle ++(0.95,1.1);
\fill[white, draw=none] (10.03,5.02) rectangle ++(0.95,1.1);
\fill[white, draw=none] (12.03,5.02) rectangle ++(0.95,1.1);
\fill[white, draw=none] (14.03,5.02) rectangle ++(0.95,1.1);
\fill[white, draw=none] (17.03,5.02) rectangle ++(0.95,1.1);

\fill[white, draw=none] (16.95,-0.1) rectangle ++(1.02,1.06);

\fill[orange!80, opacity=0.3] (22,6) rectangle ++(1,1);
\fill[blue!60, opacity=0.8] (22.5,6.5) circle(0.35cm);

\fill[cyan!20] (17,1) rectangle ++(1,1); 

\fill[orange!70, opacity=1] (17.5,1.5) circle(0.35cm); 

\end{tikzpicture}}
		\caption{SUNY}
		\label{fig:suny_grid}
	\end{subfigure}
	\caption{Maps of Grid World~\cite{10.5555/1558109.1558118} benchmarks.}
	\label{fig:gridworld}
\end{figure}


%
\paragraph{First Experminemt Setup.}
In the first experiment, to ensure a fair comparison, we evaluate \method combined with CQ-Learning against the shielding method~\cite{elsayed2021safe} also using CQ-Learning. 
The implementation and hyperparameters of CQ-Learning are taken from~\cite{elsayed2021safe}. 
Since CQ-Learning initially employs a decentralized phase (i.e., a single-agent state space), we modify the formula $\varphi_{\text{SRL}}$ to remove inter-agent dependencies:
\[
\forall \tronesafe . \forall \trtwosafe.~ 
\Big(\F \Tonesafe  ~\land~ \F \Ttwosafe ~\land ~  \always \collide\Big)
\]
In this setting, Skolemization is unnecessary, and the agents act independently. 
When the algorithm transitions into a joint phase, expanding the state space to include inter-agent interactions, we use the original specification $\varphi_{\text{SRL}}$. 
At this stage, \method guides the learning process effectively to achieve both goal completion and collision avoidance.

\paragraph{Second Experminemt Setup.} We employ DQN as our learning algorithm, utilizing a neural network with three hidden layers of 512 nodes and ReLU activation functions.
We set the discount factor to $\gamma = 1.0$, the learning rate to 0.001, the initial $\epsilon$ to 1.0 with a decay rate of 0.995 down to a minimum of 0.01, and use the Adam optimizer.
We set the number of training episodes to 200 for the SUNY, ISR, and Pentagon maps, and to 300 for the MIT map due to its increased complexity. Each episode consists of 300 steps (see~\Cref{fig:gridworld} for the map setups).

\paragraph{Baselines Reward Functions.} The manually designed reward functions have the following form:
\[
R^{\text{SRL}} = 
\begin{cases}
	a & \text{if both agents reach their respective goals,} \\
	b & \text{if one agent reaches its goal,} \\
	c & \text{if the agents collide.}
\end{cases}
\]
The function $R^{\text{SRL}}$ addresses all objectives and safety constraints of the problem. The manually designed reward function is tested with the following values:
\begin{itemize}
    \item $a=10, b=5, c=-5$ ($R_1^{\text{SRL}}$)
    \item $a=2, b=1, c=-1$ ($R_2^{\text{SRL}}$)
    \item $a=20, b=10, c=-10$ ($R_3^{\text{SRL}}$)
    \item $a=100, b=50, c=-10$ ($R_4^{\text{SRL}}$)
    \item $a=10, b=5, c=-10$ ($R_5^{\text{SRL}}$)
\end{itemize}
In the main paper, we report the results obtained using values $a=10, b=5, c=-5$ ($R_1^{\text{SRL}}$) for the baseline reward function.
\Cref{fig:srl_baseline} presents the results corresponding to all different reward functions $R^1$, $R^2$, $R^3$, $R^4$, and $R^5$ we introduced above.

\begin{figure*}[t]
	\centering
	\begin{subfigure}[t]{0.24\textwidth}
		\centering
		\scalebox{.57}{ 
            \begin{tikzpicture}
    \begin{axis}[
        width=2\textwidth,
        height=2\textwidth,
        xtick={100,200},
        ytick={50,100,150},
        xlabel={\bf Episodes (ISR)},
        xmin=0,
        xmax=200,
        xticklabel style={},
        ylabel={\bf  Total Reach},
        yticklabel style={},
        ymin=0,
        unbounded coords=discard 
    ]
    

    \addplot[color=red!50,  thick ] table [x=episode, y=mean_done, col sep=comma] {figs/isr_base.csv};

        \addplot[color=purple!50,  thick ] table [x=episode, y=total_done, col sep=comma] {figs/isr_base_1.csv};
    
    \addplot[color=teal!50,  thick ] table [x=episode, y=total_done, col sep=comma] {figs/isr_base_2.csv};
    
    \addplot[color=olive!50,  thick ] table [x=episode, y=total_done, col sep=comma] {figs/isr_base_3.csv};
    
    \addplot[color=orange!50,  thick ] table [x=episode, y=total_done, col sep=comma] {figs/isr_base_4.csv};

    \end{axis}
    \end{tikzpicture}
		}
	\end{subfigure}
       \hspace{0.5em}
	\begin{subfigure}[t]{0.24\textwidth}
		\centering
		\scalebox{.57}{
            \begin{tikzpicture}
    \begin{axis}[
        width=2\textwidth,
        height=2\textwidth,
        legend style={at={(0.8,0.02)},anchor=south,draw=none},
        xtick={100,200},
        ytick={20,40,60},
        ylabel={},
        xlabel={\bf Episodes (PENTAGON)},
        xmin=0,
        xmax=200,
        xticklabel style={},
        yticklabel style={},
        ymin=0,
        unbounded coords=discard 
        ]
    
    
    \addplot[color=red!50,  thick ] table [x=episode, y=mean_done, col sep=comma] {figs/pent_base.csv};
    
    \addplot[color=purple!50,  thick ] table [x=episode, y=total_done, col sep=comma] {figs/pent_base_1.csv};
    
    \addplot[color=teal!50,  thick ] table [x=episode, y=total_done, col sep=comma] {figs/pent_base_2.csv};
    
    \addplot[color=olive!50,  thick ] table [x=episode, y=total_done, col sep=comma] {figs/pent_base_3.csv};
    
    \addplot[color=orange!50,  thick ] table [x=episode, y=total_done, col sep=comma] {figs/pent_base_4.csv};

    \end{axis}
    \end{tikzpicture}
    
		}
	\end{subfigure}
        \hspace{-2mm}
	\begin{subfigure}[t]{0.24\textwidth}
		\centering
		\scalebox{.57}{ 
            \begin{tikzpicture}
    \begin{axis}[
        width=2\textwidth,
        height=2\textwidth,
        xtick={100,200},
        ytick={50,100,150},
        xmin=0,
        xmax=200,
        xlabel={\bf Episodes (SUNY)},
        xticklabel style={},
        yticklabel style={},
        ymin=0,
        unbounded coords=discard 
    ]

    \addplot[color=red!50 ] table [x=episode, y=mean_done, col sep=comma] {figs/suny_base.csv};
    
    \addplot[color=purple!50,  thick ] table [x=episode, y=total_done, col sep=comma] {figs/suny_base_1.csv};
    
    \addplot[color=teal!50,  thick ] table [x=episode, y=total_done, col sep=comma] {figs/suny_base_2.csv};
    
    \addplot[color=olive!50,  thick ] table [x=episode, y=total_done, col sep=comma] {figs/suny_base_3.csv};
    
    \addplot[color=orange!50,  thick ] table [x=episode, y=total_done, col sep=comma] {figs/suny_base_4.csv};
    
    \end{axis}
    \end{tikzpicture}
		}
	\end{subfigure}
        \hspace{-2mm}
	\begin{subfigure}[t]{0.24\textwidth}
		\centering
		\scalebox{.57}{ 
			\begin{tikzpicture}
    \begin{axis}[
        width=2\textwidth,
        height=2\textwidth,
        legend style={at={(0.8,0.02)},anchor=south,draw=none},
        xtick={150,300},
        ytick={5,10,15},
        ylabel={},
        xlabel={\bf Episodes (MIT)},
        xmin=0,
        xmax=300,
    	xticklabel style={},
    yticklabel style={},
        ymin=0,
        unbounded coords=discard 
    ]
    
    
    \addplot[color=red!50,  thick ] table [x=episode, y=mean_done, col sep=comma] {figs/mit_base.csv};
    
    \addplot[color=purple!50,  thick ] table [x=episode, y=total_done, col sep=comma] {figs/mit_base_1.csv};

\addplot[color=teal!50,  thick ] table [x=episode, y=total_done, col sep=comma] {figs/mit_base_2.csv};

\addplot[color=olive!50,  thick ] table [x=episode, y=total_done, col sep=comma] {figs/mit_base_3.csv};

\addplot[color=orange!50,  thick ] table [x=episode, y=total_done, col sep=comma] {figs/mit_base_4.csv};

    \end{axis}
    \end{tikzpicture}
    
		}
		
	\end{subfigure}
	\vspace{0.5cm}
	\begin{subfigure}[t]{0.24\linewidth}
		\centering
		\scalebox{.57}{ 
            \begin{tikzpicture}
    \begin{axis}[
        width=2\textwidth,
        height=2\textwidth,
        xtick={100,200},
        ytick={0,5,10},
         ylabel={\bf Collisions},
        xlabel = {\bf Episodes (ISR)},
        xmin=0,
        xmax=200,
        xticklabel style={},
        yticklabel style={},
        ymin=0,
        unbounded coords=discard 
    ]
    


        \addplot[color=purple!50,  thick ] table [x=episode, y=total_col, col sep=comma] {figs/isr_base_1.csv};
    
    \addplot[color=teal!50,  thick ] table [x=episode, y=total_col, col sep=comma] {figs/isr_base_2.csv};
    
    \addplot[color=olive!50,  thick ] table [x=episode, y=total_col, col sep=comma] {figs/isr_base_3.csv};
    

    \end{axis}
    \end{tikzpicture}
		}
	\end{subfigure}
       \hspace{3mm}
	\begin{subfigure}[t]{0.24\linewidth}
		\centering
		\scalebox{.57}{ 
            \begin{tikzpicture}
    \begin{axis}[
        width=2\textwidth,
        height=2\textwidth,
        xtick={100,200},
        ytick={0,5,10},
         ylabel={},
        xlabel = {\bf Episodes (PENTAGON)},
        xmin=0,
        xmax=200,
        xticklabel style={},
        yticklabel style={},
        ymin=0,
        unbounded coords=discard 
    ]
    
    
    
    \addplot[color=purple!50,  thick ] table [x=episode, y=total_col, col sep=comma] {figs/pent_base_1.csv};
    
    \addplot[color=teal!50,  thick ] table [x=episode, y=total_col, col sep=comma] {figs/pent_base_2.csv};
    
    \addplot[color=olive!50,  thick ] table [x=episode, y=total_col, col sep=comma] {figs/pent_base_3.csv};
    

    \end{axis}
    \end{tikzpicture}
		}
	\end{subfigure}
        \hspace{-2mm}
	\begin{subfigure}[t]{0.24\linewidth}
		\centering
		\scalebox{.57}{ 
            \begin{tikzpicture}
    \begin{axis}[
        width=2\textwidth,
        height=2\textwidth,
        xlabel = {\bf Episodes (SUNY)},
        xtick={100,200},
        ytick={0,5,10},
        xmin=0,
        xmax=200,
        xticklabel style={},
        yticklabel style={},
        ymin=0,
        unbounded coords=discard 
    ]
    
    
    
    \addplot[color=purple!50,  thick ] table [x=episode, y=total_col, col sep=comma] {figs/suny_base_1.csv};\label{line:reward1}

    \addplot[color=teal!50,  thick ] table [x=episode, y=total_col, col sep=comma] {figs/suny_base_2.csv};\label{line:reward2}
    
    \addplot[color=olive!50,  thick ] table [x=episode, y=total_col, col sep=comma] {figs/suny_base_3.csv};\label{line:reward3}
    
    
    \end{axis}
    \end{tikzpicture}
		}
	\end{subfigure}
        \hspace{-2mm}
	\begin{subfigure}[t]{0.24\linewidth}
		\centering
		\scalebox{.57}{ 
			\begin{tikzpicture}
    \begin{axis}[
        width=2\textwidth,
        height=2\textwidth,
        xtick={150,300},
        ytick={0,5,10},
         ylabel={},
        xlabel = {\bf Episodes (MIT)},
        xmin=0,
        xmax=300,
        xticklabel style={},
    yticklabel style={},
        ymin=0,
        unbounded coords=discard 
    ]
    
    

    \addplot[color=purple!50,  thick ] table [x=episode, y=total_col, col sep=comma] {figs/mit_base_1.csv};

\addplot[color=teal!50,  thick ] table [x=episode, y=total_col, col sep=comma] {figs/mit_base_2.csv};

\addplot[color=olive!50,  thick ] table [x=episode, y=total_col, col sep=comma] {figs/mit_base_3.csv};


    \end{axis}
    \end{tikzpicture}
		}
	\end{subfigure}
	\begin{subfigure}[t]{0.24\linewidth}
		\centering
		\scalebox{.57}{ 
            \begin{tikzpicture}
    \begin{axis}[
        width=2\textwidth,
        height=2\textwidth,
        xtick={100,200},
        ytick={0,5,10},
         ylabel={\bf Collisions},
        xlabel = {\bf Episodes (ISR)},
        xmin=0,
        xmax=200,
        xticklabel style={},
        yticklabel style={},
        ymin=0,
        unbounded coords=discard 
    ]
    

    \addplot[color=red!50, thick ] table [x=episode, y=mean_col, col sep=comma] {figs/isr_base.csv};

    
    
    
    \addplot[color=orange!50,  thick ] table [x=episode, y=total_col, col sep=comma] {figs/isr_base_4.csv};

    \end{axis}
    \end{tikzpicture}
		}
	\end{subfigure}
       \hspace{3mm}
	\begin{subfigure}[t]{0.24\linewidth}
		\centering
		\scalebox{.57}{ 
            \begin{tikzpicture}
    \begin{axis}[
        width=2\textwidth,
        height=2\textwidth,
        xtick={100,200},
        ytick={0,5,10},
         ylabel={},
        xlabel = {\bf Episodes (PENTAGON)},
        xmin=0,
        xmax=200,
        xticklabel style={},
        yticklabel style={},
        ymin=0,
        unbounded coords=discard 
    ]
    
    
    \addplot[color=red!50,  thick ] table [x=episode, y=mean_col, col sep=comma] {figs/pent_base.csv};
    
    
    
    
    \addplot[color=orange!50,  thick ] table [x=episode, y=total_col, col sep=comma] {figs/pent_base_4.csv};

    \end{axis}
    \end{tikzpicture}
		}
	\end{subfigure}
        \hspace{-2mm}
	\begin{subfigure}[t]{0.24\linewidth}
		\centering
		\scalebox{.57}{ 
            \begin{tikzpicture}
    \begin{axis}[
        width=2\textwidth,
        height=2\textwidth,
        xlabel = {\bf Episodes (SUNY)},
        xtick={100,200},
        ytick={0,5,10},
        xmin=0,
        xmax=200,
        xticklabel style={},
        yticklabel style={},
        ymin=0,
        unbounded coords=discard 
    ]
    
    
    \addplot[color=red!50,  thick ] table [x=episode, y=mean_col, col sep=comma] {figs/suny_base.csv};\label{line:reward}
    

    
    
    \addplot[color=orange!50,  thick ] table [x=episode, y=total_col, col sep=comma] {figs/suny_base_4.csv};\label{line:reward4}
    
    \end{axis}
    \end{tikzpicture}
		}
	\end{subfigure}
        \hspace{-2mm}
	\begin{subfigure}[t]{0.24\linewidth}
		\centering
		\scalebox{.57}{ 
			\begin{tikzpicture}
    \begin{axis}[
        width=2\textwidth,
        height=2\textwidth,
        xtick={150,300},
        ytick={0,5,10},
         ylabel={},
        xlabel = {\bf Episodes (MIT)},
        xmin=0,
        xmax=300,
        xticklabel style={},
    yticklabel style={},
        ymin=0,
        unbounded coords=discard 
    ]
    
    
    \addplot[color=red!50,  thick ] table [x=episode, y=mean_col, col sep=comma] {figs/mit_base.csv};




\addplot[color=orange!50,  thick ] table [x=episode, y=total_col, col sep=comma] {figs/mit_base_4.csv};

    \end{axis}
    \end{tikzpicture}
		}
	\end{subfigure}
	\caption{Total number of successful goal completions by both agents (top) and number of collisions (bottom). Lines~\ref{line:reward}, \ref{line:reward1}, \ref{line:reward2}, \ref{line:reward3}, and~\ref{line:reward4} correspond to reward functions $R^1$, $R^2$, $R^3$, $R^4$, and $R^5$, respectively. Results are averaged over 10 runs.}
	\label{fig:srl_baseline}
	\vspace{-5mm}
\end{figure*}


\subsection{Deep Sea Treasure (DST)}
\paragraph{Case Study Setup.}This case study (see \Cref{fig:dst_grid}) was originally proposed by~\cite{vptrb2011}. 
We adopt the implementation provided in~\cite{NEURIPS2019_4a46fbfc}. 
The state space is represented as a tuple $\tupleof{x, y, step}$, and the action space is defined as $\action = \{\textit{up}, \textit{down}, \textit{left}, \textit{right} \}$.
The following \HyperLTL formula $\varphi_{\text{DST}}$ specifies the required objectives:
\[
\varphi_{\text{DST}} \triangleq \forall \traceone. \exists \tracetwo.  
        \F (\textit{T1}_\traceone \land 
          \F (\textit{T2}_\traceone \land 
            \F (\textit{T3}_\traceone) \dots))
        ~\land \G~(\textit{step}_{\tracetwo} < \delta) \land \G~(|\textit{pos}_\traceone - \textit{pos}_\tracetwo| < 1)
\]
The original case study is in a single-agent environment with two objectives. 
We modified it into a multi-agent problem by assigning one agent with the objective of maximizing collected treasure, and another agent with the task of keeping the number of steps below a certain threshold. 

\paragraph{Experminemt Setup.}For our experiments, we used PPO and DQN implementations from~\cite{stable-baselines3}. 
For DQN, we set the hyperparameters as follows: discount factor $\gamma = 0.99$, $\epsilon$ decaying from 1.0 to 0.07 over a fraction of 0.2, and a learning rate of $0.0004$.
Since the DQN implementation from~\cite{stable-baselines3} does not directly support multi-discrete actions, we expanded the action spaces of both agents and defined a unified action space compatible with this implementation.
For PPO, we set hyperparameters as following, $\gamma = 0.99$, clipping factor to $0.2$, learning rate to 0.0003, and GAE lambda to 0.98. 
Additionally, to ensure a fair comparison, we added a reward penalty of $-10$ to the baseline reward function whenever the two agents became separated.

\begin{table}[b!]
    \centering
    \caption{DST results on steps required to achieve the desired treasures, showing average and standard error over 10 evaluations.}
    \resizebox{\textwidth}{!}{
    \begin{tabular}{cccccccc}
        \toprule
        & & \multicolumn{6}{c}{Treasures Achieved}\\
        \cmidrule(r){3-8}
        Method & Episodes & 1 & 5 & 10 & 15 & 20 & 30 \\
        \midrule
        PPO & 500 & 199.01$\pm6.65$ & 218.66$\pm6.67$ & 244.87$\pm6.68$ & 247.89$\pm6.67$ & 253.16$\pm6.66$ & 268.01$\pm6.70$ \\
        PPO +\method & 500 & \textbf{13.30}$\pm1.01$ & \textbf{29.72}$\pm1.57$ & \textbf{39.10}$\pm1.81$ & \textbf{49.22}$\pm1.96$ & \textbf{56.59}$\pm2.03$ & \textbf{71.55}$\pm2.26$ \\
        \midrule
        PPO & 1000 & \textbf{5.75}$\pm0.30$ & \textbf{6.75}$\pm0.30$ & \textbf{29.68}$\pm0.76$ & \textbf{32.29}$\pm0.75$ &  \textbf{36.09}$\pm0.74$ & \textbf{39.82}$\pm0.72$ \\
        PPO +\method & 1000 & 55.37$\pm3.66$ & 106.89$\pm4.96$ & 126.25$\pm5.10$ & 141.06$\pm5.27$ & 153.49$\pm5.35$ & 183.51$\pm5.70$ \\
        \midrule
        DQN & 500 & 832.10$\pm9.93$ & 897.54$\pm6.95$ & 916.40$\pm5.63$ & 917.64$\pm5.55$ &917.89$\pm5.54$ & 926.33$\pm5.25$ \\
        DQN +\method & 500 & \textbf{114.60}$\pm2.67$ & \textbf{115.61}$\pm2.56$ & \textbf{116.61}$\pm2.52$ & \textbf{117.61}$\pm2.50$ & \textbf{148.66}$\pm4.34$ & \textbf{155.39}$\pm4.58$ \\
        \midrule
        DQN & 1000 & 893.37$\pm9.67$ & 953.25$\pm6.55$ & 992.10$\pm2.50$ & 992.12$\pm2.55$ &996.05$\pm1.52$ & 997.19$\pm1.14$ \\
        DQN +\method & 1000 & \textbf{303.63}$\pm6.64$ & \textbf{319.09}$\pm6.38$ & \textbf{335.20}$\pm6.21$ & \textbf{335.20}$\pm6.21$ & \textbf{337.21}$\pm6.18$ & \textbf{339.90}$\pm6.22$ \\

        \bottomrule
    \end{tabular}
    }
    \label{tab:dst-step}
\end{table}

\subsection{Additional Experiment on DST}
We conduct an additional experiment on DST,  
where the focus is on minimizing the step constraint in the problem. 
In \Cref{tab:dst-step}, we report the number of steps required for the cumulative treasure achieved by the treasure collector to exceed values of 1, 5, 10, 15, 20, and 30.
The results show that \method consistently and significantly outperforms the baselines in all settings, except when using PPO trained for 1000 episodes.

\begin{figure}[t]
	\begin{minipage}{.6\textwidth}
	\centering
    \resizebox{.62\textwidth}{!}{\input{figs/dst_grid.tex}}
	\caption{DST Grid Map.}
	\label{fig:dst_grid}
\end{minipage}
~
\begin{minipage}{.35\textwidth}
	\centering
\scalebox{.8}{
	\input{figs/fair_grid}
}
\caption{Job Scheduling benchmark.}
\label{fig:fair_grid}
\end{minipage}
\vspace{-5mm}
\end{figure}

\subsubsection{The Post Correspondence Problem (PCP)}
\label{sec:pcp}

\begin{figure}[t]
    \centering
    \includegraphics[width=0.8\textwidth]{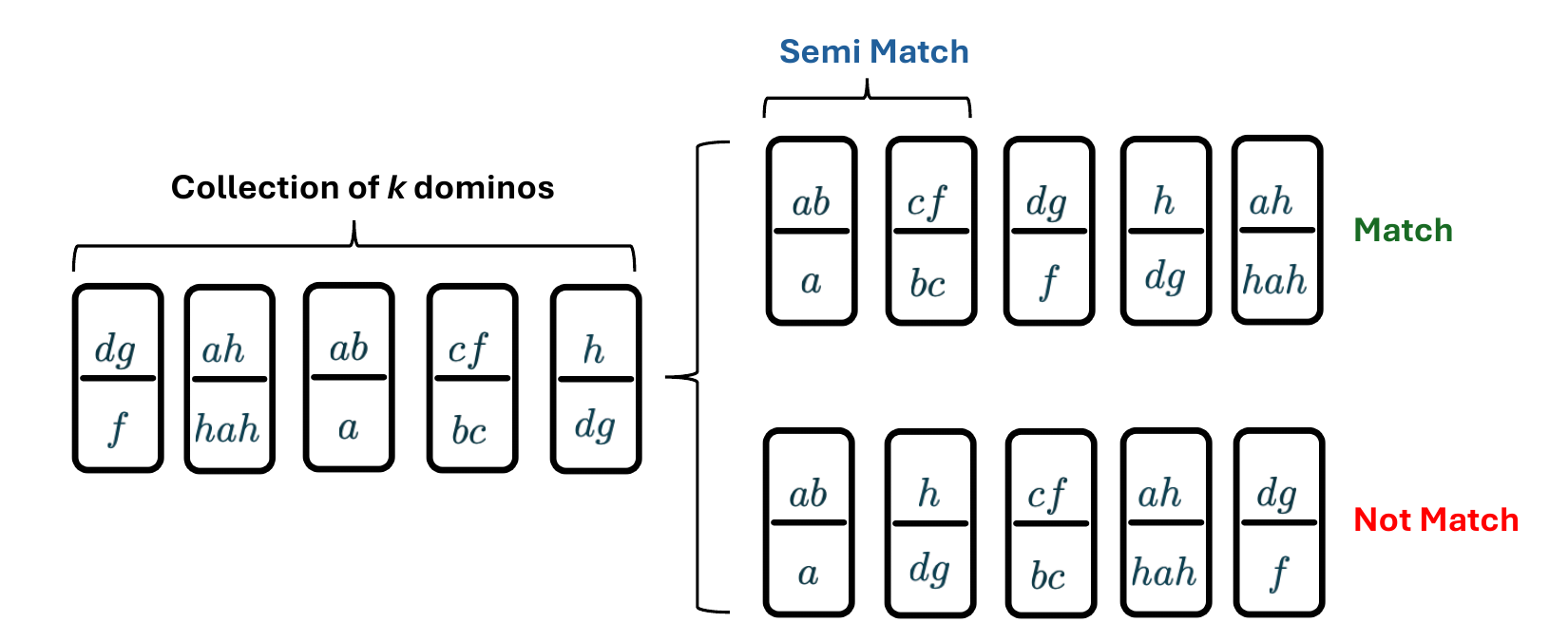}
    \caption{PCP overview}
    \label{fig:pcp_env}
  \end{figure}

\paragraph{Case Study Setup.}
PCP (see \Cref{fig:pcp_env}) consists of a set of $k$ dominos, denoted as 
$\Dominos = \{\dom_0, \dom_1, \dots, \dom_k\}$. 
Each domino $\dom_i$ ($0 \leq i \leq k$) is represented by a pair of nonempty finite words $(\topp_i, 
\bottom_i)$ from a given alphabet $\alphabet$. 
The objective is to find a finite sequence of dominos such that the concatenated words on the top 
match those on the bottom. 
In our case study, the state space is defined as a tuple $\tupleof{\topp, \bottom}$ for each domino. 
The action space consists of selecting a domino from $\Dominos$, resulting in $k$ possible actions $A = \{\dom_0, \dom_1, \dots, \dom_k\}$.
Notably, traces (i.e., agents) are unaware of the context of the dominos and can only identify them by their labels (i.e., $dom_i$). 
The initial state of the MDP encodes empty words on both the top and bottom: $s^0 = 
\tupleof{\varepsilon, \varepsilon}$. 
Subsequently, they choose actions from the action space (choosing a domino) to construct traces that aims to satisfy the PCP objective.

Before encoding PCP in \HyperLTL, we note that in~\cite{fh16,bs23}, the authors reduce PCP to the 
satisfiability problem for the $\forall\exists$ fragment of \HyperLTL and the emptiness problem of 
nondeterministic finite-state hyper automata. 
Part of the encoding is to ensure that only valid dominos are chosen. We do not need those 
constraints in our encoding, as the action space of the MDP enforces choosing valid dominos only.
This is the reason our that encoding is less complex than that of~\cite{fh16,bs23}.

We formalize a valid solution to PCP in \HyperLTL as follows.
%
The set $\AP$ of atomic propositions is defined such that as $\alphabet = 2^\AP \cup \{\#\}$, where 
$\#$ encodes termination. 
Essentially, the \HyperLTL formula requires that for all traces $\traceone$, where top and bottom 
words match up to the end of the shorter trace, there exists a trace $\tracetwo$ such that 
$\traceone$ is a $\semi$ and $\tracetwo$ extends $\traceone$ to complete equal top and 
bottom words:
%
\[
\varphi_{\pcp} \triangleq \forall \traceone. \exists \tracetwo.~ 
\psi_{\semi_{\traceone}} ~\until~
\big( \psi_{\extend_{\traceone, \tracetwo}} \land
\psi_{\match_{\tracetwo}}  \big)
\]
where $\varphi_{\semi}$ means the top and bottom words match up to the length of the shorter 
word:
\[
\psi_{\semi_{\traceone}} \triangleq \Big[\underset{p \in \AP}{\bigwedge} (p_{\topp_{\traceone}} 
\leftrightarrow p_{\bottom_{\traceone}})\Big]~\until (\#_{\topp_{\traceone}} \oplus 
\#_{\bottom_{\traceone}})
\]
where `$\oplus$' is the xor operator. The formula $\varphi_{\match}$ indicates that the word 
constructed on the top and bottom are equal:
\[
\psi_{\match_{\tracetwo}} \triangleq \always \underset{p \in \AP}{\bigwedge} (p_{\topp_{\tracetwo}}  
\leftrightarrow 
p_{\bottom_{\tracetwo}})
\]
Finally, formula $\varphi_{\extend_{\traceone, \tracetwo}}$ encodes that trace $\tracetwo$ is a successor 
trace $\traceone$ as follows:
%
\begin{multline*}
	\varphi_{\extend_{\traceone, \tracetwo}} \triangleq \Big[\underset{p \in \AP}{\bigwedge} \big(  (p_{\topp_{\traceone}} \leftrightarrow p_{\topp_{\tracetwo}})  \land  (p_{\bottom_{\traceone}} \leftrightarrow p_{\bottom_{\tracetwo}})   \big) \Big] ~ \until ~ \\ \big( (\#_{\topp_{\traceone}} \lor \#_{\bottom_{\traceone}} ) \land (\neg \#_{\topp_{\tracetwo}} \land \neg \#_{\bottom_{\tracetwo}})\big)
\end{multline*}


\paragraph{Experminemt Setup.}We employ DQN as our learning algorithm, utilizing a neural network with three layers of 512 nodes and ReLU activation functions. 
We set the discount factor $\gamma$ to 0.99, the learning rate to 0.001, and the initial $\epsilon$ to 1.0 with a decay rate of 0.995 down to a minimum of 0.001. We use the Adam optimizer and train for 1000 episodes.
This experiment is conducted for domino collections containing 5 and 6 dominos.

\paragraph{Baselines Reward Functions.} The manually designed reward functions have the following form: 
\[
R^{\text{PCP}} = \begin{cases}
	a & \text{if the same indexed letters on the top and bottom are equal,} \\
	b & \text{otherwise.} 
\end{cases}
\]

The function $R^{\text{PCP}}$ addresses all objectives and safety constraints of the problem. The manually designed reward function is tested with the following values:
\begin{itemize}
    \item $a=1, b=-1$ ($R_1^{\text{PCP}}$)
    \item $a=5, b=-2$ ($R_2^{\text{PCP}}$)
    \item $a=10, b=-10$ ($R_3^{\text{PCP}}$)
    \item $a=1, b=-5$ ($R_4^{\text{PCP}}$)
\end{itemize}
In the paper, we report results using the baseline reward function with values $a = 5$, $b = -2$ ($R_2^{\text{PCP}}$) for $k = 5$, and $a = 1$, $b = -1$ ($R_1^{\text{PCP}}$) for $k = 6$. \Cref{fig:pcp_base} presents the results corresponding to the different reward functions. 
The baselines $R_2^{\text{PCP}}$, $R_3^{\text{PCP}}$, and $R_4^{\text{PCP}}$ for $k = 6$ did not successfully produce any matches. We visualize this using a value of zero in the log-scale plot shown in \Cref{fig:pcp_base}.

\begin{figure}[t!]
	\centering
	\scalebox{.5}{
	\begin{tikzpicture}
\begin{axis}[
    xlabel={\Large\bf Episodes},
    ylabel={\Large\bf \# of successful match},
    width=1.3\textwidth,
    height=0.8\textwidth,
    legend style={
        at={(0.5,-0.15)},   
        anchor=north,       
        draw=black,
        font=\Large,
        inner sep=10pt
      },
      legend columns=4,
    xtick={200,400,600,800,1000},
    xticklabel style={font=\Large},
    yticklabel style={font=\Large},
    ylabel style={yshift=15pt},
    ytick={1e-1,1e0,1e1,1e2},
    xmin=0,
    ymode=log,
    xmax=1000,
    ymin=1e-4,                    
    extra y ticks={1e-3,1e-2},
    extra y tick labels={0,\vdots},
    extra y tick style={tick pos=left, grid=none},
]

\addplot[color=Blue,  ultra thick, solid ] table [x=episode, y=mean, col sep=comma] 
{figs/dom5_base_1.csv};\addlegendentry{$R_2^{\text{PCP}}$, $k=5$}

\addplot[color=Orange,  ultra thick, solid ] table [x=episode, y=mean_done, col sep=comma] 
{figs/dom_base5.csv};\addlegendentry{$R_1^{\text{PCP}}$, $k=5$}

\addplot[color=Cyan,  ultra thick, solid] table [x=episode, y=mean_done, col sep=comma] 
{figs/dom_base5_2.csv};\addlegendentry{$R_3^{\text{PCP}}$, $k=5$}
\addplot[color=Purple,  ultra thick, solid ] table [x=episode, y=mean_done, col sep=comma] 
{figs/dom_base5_3.csv};\addlegendentry{$R_4^{\text{PCP}}$, $k=5$}

\addplot[color=Maroon,  ultra thick, solid ] table [x=episode, y=tot_done, col sep=comma] 
{figs/dom6_base.csv};\addlegendentry{$R_1^{\text{PCP}}$, $k=6$}

\addplot[color=Red,ultra thick,solid,domain=0:1000,samples=2]
  {0.0012};
\addlegendentry{$R_2^{\text{PCP}}$, $k=6$}

\addplot[color=Yellow,ultra thick,solid,domain=0:1000,samples=2]
  {0.001};
\addlegendentry{$R_3^{\text{PCP}}$, $k=6$}

\addplot[color=Green,ultra thick,solid,domain=0:1000,samples=2]
  {0.0008};
\addlegendentry{$R_4^{\text{PCP}}$, $k=6$}

\end{axis}
\end{tikzpicture}
	}
	\caption{PCP baseline results showing average total matches over 10 runs.    }
	\label{fig:pcp_base}
\end{figure}

\begin{table}[b!]
    \centering
    \caption{Wildfire results, avg. and std. error of 10 trials after 5k episodes.}
    \begin{tabular}{p{3mm}p{21mm}lll}
        \toprule
        Size & Method &  \texttt{Dist} & Steps $O_1$ & Steps $O_2$ \\
        \midrule
        \multirow{2}{*}{$3^2$} & PPO + $R_1^{\text{FAIR}}$ & 2.8$\pm0.01$  & 77.13$\pm5.7$  & T/O \\
        & PPO + $R_2^{\text{FAIR}}$& \textbf{2.5}$\pm0.01$  & \textbf{33.43}$\pm4.1$  & \textbf{787.03}$\pm31.8$ \\
        \midrule
        \multirow{2}{*}{$5^2$} & PPO + $R_1^{\text{FAIR}}$ & 5.6$\pm0.02$ & \textbf{31.2}$\pm4.5$ & \textbf{7057.8}$\pm1047.9$  \\
        & PPO + $R_2^{\text{FAIR}}$ & \textbf{4.2}$\pm0.01$ & 62.7$\pm9.7$ & T/O  \\
        \bottomrule
    \end{tabular}
    \label{tab:wild-base}
\end{table}

\subsubsection{Wild Fire Scenario}
\paragraph{Case Study Setup.} Consider a wildfire scenario in a grid-world environment as we introduced in~\Cref{sec:intro}. 
Three locations are on fire, and two victims need to be rescued. 
Two autonomous agents are deployed from the same position with distinct objectives: the firefighter agent (\droneone) must extinguish all fire zones, and the medical agent (\dronetwo) must rescue all victims. 
The agents must also satisfy two constraints: which requires them to always remain within a 2-cell communication range; and which prohibits \dronetwo from entering any fire zone before \droneone has extinguished the fire in that zone. 
The state space is represented as a tuple $\tupleof{x, y}$, and the action space is defined as $\action = \{ \textit{stay}, \textit{up}, \textit{down}, \textit{left}, \textit{right} \}$.
The goal is to learn optimal policies for \droneone and \dronetwo that maximize the probability of satisfying all these requirements. 
The following \HyperLTL formula captures the objectives of this case study:
%
\begin{center}
	\small
	$
	\varphi_{\textsf{Rescue}}  \triangleq \forall \pathone. \exists \pathtwo.
	(\psi_{\motifputout} \land \psi_{\motifsave} \land \psi_{\motifdist}  \land \psi_{\motifsafe})
	$
			\begin{align*}
			O_1&: 
			~\psi_{\motifputout}  \definedas
			\F ({\locmark i}_{\pathone})  \land 
			\F ({\locmark f}_{\pathone}) \land 
			\F ({\locmark c}_{\pathone})
			~~
			&C_1&: 
			~\psi_{\motifdist}  \definedas
			\G (| \motifloc_{\pathone} - \motifloc_{\pathtwo} | < 3)
			\\
			O_2&:
			~\psi_{\motifsave}  \definedas 
			\F ({\locmark g}_{\pathtwo})
			\land
			\F ({\locmark f}_{\pathtwo})
			&C_2&:
			~\psi_{\motifsafe}  \definedas 
			(\neg {\locmark i}_{\pathtwo} \U {\locmark i}_{\pathone}) \land
			(\neg {\locmark f}_{\pathtwo} \U {\locmark f}_{\pathone}) \land
			(\neg {\locmark c}_{\pathtwo} \U {\locmark c}_{\pathone})
		\end{align*}
\end{center}

\paragraph{Experiment Setup.}
We use the PPO implementation from~\cite{stable-baselines3}, with the following hyperparameters: a learning rate of 0.0003, a clipping parameter of 0.2, a discount factor $\gamma = 0.995$, and a GAE lambda of 0.95.
We conduct our experiments on $3\times3$ and $5\times5$ grid-world environments.
The timeout is set to 1000 steps for the $3\times3$ grid and 10,000 steps for the $5\times5$ grid.

\paragraph{Baselines.}
We evaluated (see \Cref{tab:wild-base}) two baseline reward functions in this experiment:
\begin{itemize}
\item $R_1^{\text{FAIR}}$: extinguishing fire: $+50$, rescuing a victim: $+10$, agent out of range: $-100$, and \dronetwo in a fire zone: $-100$.
\item $R_2^{\text{FAIR}}$: Extinguishing fire: $+10$, rescuing a victim: $+50$, agent out of range: $-100$, and \dronetwo in a fire zone: $-100$.
\end{itemize}
We used $R_2^{\text{FAIR}}$ in the paper, although in $5\times5$ $R_1^{\text{FAIR}}$ performs better than $R_2^{\text{FAIR}}$ (in terms of only Steps $O_1$ and Steps $O_2$). 
However, we chose to include $R_2^{\text{FAIR}}$ in the main paper because its performance in the $3\times3$ grid is consistently better than $R_1^{\text{FAIR}}$ across all metrics. This trend does not hold in the $5\times5$ grid, where $R_1^{\text{FAIR}}$ outperforms $R_2^{\text{FAIR}}$ only in Steps $O_1$ and Steps $O_2$.



\subsection{Additional Experiment}

\paragraph{Case Study Setup.} In this case study (see~\Cref{fig:fair_grid}), a single permanent resource is placed on a grid 
with multiple agents, which must learn to share resources.
The objective here is to maximize the overall utility of all agents while ensuring fair allocation of the 
resource. 
In other words, the goal is not merely to maximize utility by allocating the resource to a single agent 
but to distribute it equitably among all agents.
The action space is defined as $\action = \{ \textit{stay}, \textit{up}, \textit{down}, \textit{left}, \textit{right} \}$.
The state space is extended by $\tupleof{x,y,\textsf{Energy}}$. 
We expresses our allocation and fairness objectives by the following \HyperLTL formula:
%
\[ \varphi_{\text{FAIR}} \triangleq
\forall \traceone. \forall \tracetwo.~ 
\Big(\G\F \ap{Resource}_{\traceone} ~\land~ \G\F \ap{Resource}_{\tracetwo}\Big) \land \Box\big( 
|\textsf{Energy}_{\traceone} - \textsf{Energy}_{\tracetwo}| < \delta \big)
\]
That is, all agents should eventually gain access to the resource at every step, while ensuring that 
the difference in their $\textsf{Energy}$ levels (i.e., allocated resources) remains less than a 
threshold $\delta$, which can be set as a hyperparameter\footnote{We acknowledge that 
$\always\F$ is a B\"{u}chi condition and is strange to be use in finite semantics. Nevertheless, when 
it is interpreted in the context of robustness values, function $\rho$ attempts to maximize the 
occurrence of $\ap{Resource}$, which is the intended objective.}.  In our setup, agents start with $\textsf{Energy} = 0$, and each time an agent reaches the resource 
position, its energy level increments by one while maintaining the same action space.

\paragraph{Experiment Setup.}
We employ PPO as our learning algorithm, using two neural networks: a policy network and a value network. Both networks consist of three hidden layers with 512 nodes and ReLU activation functions. The policy network uses a softmax activation in the output layer, while the value network uses a linear output. We set the learning rate to 0.001, the discount factor $\gamma = 0.95$, and the clipping factor to 0.2. Optimization is performed using the Adam optimizer.
%
%
We also set $\delta=10$.
\paragraph{Baselines Reward Functions.}The manually designed reward functions have the following form: 
\[
R^{\text{FAIR}} = \begin{cases}
	a & \text{if resource allocated to either of agents,} \\
	b & \text{if } |\textsf{Energy}_{\text{Agent 1}} - \textsf{Energy}_{\text{Agent 2}}| > \delta
\end{cases}
\]

The function $R^{\text{FAIR}}$ addresses all objectives and safety constraints of the problem. The manually designed reward function is tested with the following values:
\begin{itemize}
    \item $a=2, b=-1$ ($R_1^{\text{FAIR}}$)
    \item $a=20, b=-5$ ($R_2^{\text{FAIR}}$)
    \item $a=10, b=-5$ ($R_3^{\text{FAIR}}$)
    \item $a=5, b=-5$ ($R_4^{\text{FAIR}}$)
\end{itemize}

\paragraph{Analysis and Results.}The evaluation (see~\Cref{fig:fair}) demonstrates how the learning process maximizes utility for both agents while minimizing the difference in their utilities. 
In \Cref{fig:fair-hyp}, using \method, we observe that the two agents initially start with low resource utilization. Although this issue is addressed over time, there is a noticeable gap in resource allocation between episodes 150 and 270. 
\method gradually reduces this disparity, and after episode 400, it achieves high resource utilization while maintaining fairness. 
This is evident from the fact that the gap between the maximum and minimum utilization becomes nearly invisible between episodes 400 and 500. 
By the end of training, each agent receives approximately 40 to 45 resources per episode (with 100 steps), while the optimal allocation is 50 resources per agent. 
These results highlight the effectiveness of \method in achieving fairness in MARL.

On the other hand, using PPO with the baseline reward functions does not successfully yield a policy that both maximizes the average utility of the agents and minimizes the gap between the minimum and maximum allocated resources. 
For example, $R_1^{\text{FAIR}}$ (see~\Cref{fig:fair-1}) succeeds in maximizing the average resource allocation to the optimal level but fails to minimize the disparity between agents in terms of resource allocation.
The other baseline reward functions $R_2^{\text{FAIR}}$, $R_3^{\text{FAIR}}$, and $R_4^{\text{FAIR}}$ exhibit similar behavior (see~\Cref{fig:fair-2,fig:fair-3,fig:fair-4}, respectively). 
While they perform better than $R_1^{\text{FAIR}}$ in minimizing the min-max range of resource allocation, they still fall short compared to \method in achieving both fairness and optimal utility.

\begin{figure}[t]
    \centering
    \begin{subfigure}[t]{0.95\textwidth}
        \centering
        \begin{tikzpicture}
\begin{axis}[
    xlabel={\bf Episodes},
    ylabel={\bf \# of allocations},
    width=1\textwidth,
    height=0.5\textwidth,
    legend style={at={(0.8,0.02)},anchor=south,draw=none},
    xtick={100,200,300,400,500},
    ytick={20,30,40,50},
    xmin=0,
    xmax=500,
    ymax=55,
    unbounded coords=discard 
]

\addplot[name path=upper, draw=none] table [x=steps, y=max, col sep=comma] {figs/grid_4.csv};

\addplot[name path=lower, draw=none] table [x=steps, y=min, col sep=comma] {figs/grid_4.csv};

\addplot [
    fill=Blue!20,
    opacity=0.8
] fill between[
    of=upper and lower
];\label{line:range_agent}

\addplot[Blue!80, dashed] table [x=steps, y=avg, col sep=comma] {figs/grid_4.csv};\label{line:avg}

\addplot[Blue, solid, thick] table [x=steps, y=trend, col sep=comma] 
{figs/grid_4.csv};\label{line:trend}

\addplot[black, dashdotted] coordinates {(0,50) (500,50)};\label{line:optimal}



\end{axis}
\end{tikzpicture}
        \caption{\method}
        \label{fig:fair-hyp}
    \end{subfigure}
    \begin{subfigure}[t]{0.48\textwidth}
        \centering
        \begin{tikzpicture}
\begin{axis}[
    xlabel={\bf Episodes},
    ylabel={\bf \# of allocations},
    width=1\textwidth,
    height=0.7\textwidth,
    legend style={at={(0.8,0.02)},anchor=south,draw=none},
    xtick={100,200,300,400},
    ytick={20,40, 60, 80},
    xmin=0,
    xmax=500,
    ymax=100,
    unbounded coords=discard 
]

\addplot[name path=upper, draw=none] table [x=steps, y=max, col sep=comma] {figs/fair_base_4.csv};

\addplot[name path=lower, draw=none] table [x=steps, y=min, col sep=comma] {figs/fair_base_4.csv};

\addplot [
    fill=gray!20,
    opacity=0.8
] fill between[
    of=upper and lower
];\label{line:range_agent4}

\addplot[gray, dashed] table [x=steps, y=avg, col sep=comma] {figs/fair_base_4.csv};\label{line:avg4}

\addplot[gray, solid, thick] table [x=steps, y=trend, col sep=comma] {figs/fair_base_4.csv};\label{line:trend4}

\addplot[black, dashdotted] coordinates {(0,50) (500,50)};



\end{axis}
\end{tikzpicture}
        \caption{$R_1^{\text{FAIR}}$ baseline reward function.}
        \label{fig:fair-1}
    \end{subfigure}
    \hfill
    \begin{subfigure}[t]{0.48\textwidth}
        \centering
        \begin{tikzpicture}
\begin{axis}[
    xlabel={\bf Episodes},
    ylabel={\bf \# of allocations},
    width=1\textwidth,
    height=0.7\textwidth,
    legend style={at={(0.8,0.02)},anchor=south,draw=none},
    xtick={100,200,300,400},
    ytick={10,30,50},
    xmin=0,
    xmax=500,
    ymax=70,
    unbounded coords=discard 
]

\addplot[name path=upper, draw=none] table [x=steps, y=max, col sep=comma] {figs/fair_base_1.csv};

\addplot[name path=lower, draw=none] table [x=steps, y=min, col sep=comma] {figs/fair_base_1.csv};

\addplot [
    fill=yellow!30,
    opacity=0.8
] fill between[
    of=upper and lower
];\label{line:range_agent1}

\addplot[orange, dashed] table [x=steps, y=avg, col sep=comma] {figs/fair_base_1.csv};\label{line:avg1}

\addplot[orange, solid, thick] table [x=steps, y=trend, col sep=comma] {figs/fair_base_1.csv};\label{line:trend1}

\addplot[black, dashdotted] coordinates {(0,50) (500,50)};



\end{axis}
\end{tikzpicture}
        \caption{$R_2^{\text{FAIR}}$ baseline reward function.}
        \label{fig:fair-2}
    \end{subfigure}
    
    \vspace{5mm}
    
    \begin{subfigure}[t]{0.48\textwidth}
        \centering
        \begin{tikzpicture}
\begin{axis}[
    xlabel={\bf Episodes},
    ylabel={\bf \# of allocations},
    width=1\textwidth,
    height=0.7\textwidth,
    legend style={at={(0.8,0.02)},anchor=south,draw=none},
    xtick={100,200,300,400},
    ytick={10,30,50},
    xmin=0,
    xmax=500,
    ymax=70,
    unbounded coords=discard 
]

\addplot[name path=upper, draw=none] table [x=steps, y=max, col sep=comma] {figs/fair_base_2.csv};

\addplot[name path=lower, draw=none] table [x=steps, y=min, col sep=comma] {figs/fair_base_2.csv};

\addplot [
    fill=red!20,
    opacity=0.8
] fill between[
    of=upper and lower
];\label{line:range_agent2}

\addplot[red, dashed] table [x=steps, y=avg, col sep=comma] {figs/fair_base_2.csv};\label{line:avg2}

\addplot[red, solid, thick] table [x=steps, y=trend, col sep=comma] {figs/fair_base_2.csv};\label{line:trend2}

\addplot[black, dashdotted] coordinates {(0,50) (500,50)};



\end{axis}
\end{tikzpicture}
        \caption{$R_3^{\text{FAIR}}$baseline reward function.}
        \label{fig:fair-3}
    \end{subfigure}
    \hfill
    \begin{subfigure}[t]{0.48\textwidth}
        \centering
        \begin{tikzpicture}
\begin{axis}[
    xlabel={\bf Episodes},
    ylabel={\bf \# of allocations},
    width=1\textwidth,
    height=0.7\textwidth,
    legend style={at={(0.8,0.02)},anchor=south,draw=none},
    xtick={100,200,300,400},
    ytick={10,30,50},
    xmin=0,
    xmax=500,
    ymax=75,
    unbounded coords=discard 
]

\addplot[name path=upper, draw=none] table [x=steps, y=max, col sep=comma] {figs/fair_base_3.csv};

\addplot[name path=lower, draw=none] table [x=steps, y=min, col sep=comma] {figs/fair_base_3.csv};

\addplot [
    fill=purple!20,
    opacity=0.8
] fill between[
    of=upper and lower
];\label{line:range_agent3}

\addplot[purple, dashed] table [x=steps, y=avg, col sep=comma] {figs/fair_base_3.csv};\label{line:avg3}

\addplot[purple, solid, thick] table [x=steps, y=trend, col sep=comma] {figs/fair_base_3.csv};\label{line:trend3}

\addplot[black, dashdotted] coordinates {(0,50) (500,50)};



\end{axis}
\end{tikzpicture}
        \caption{$R_4^{\text{FAIR}}$ baseline reward function.}
        \label{fig:fair-4}
    \end{subfigure}
    
    \caption{Fair resource allocation results under various baseline reward functions. Boundaries represent the minimum and maximum allocated resources per episode. Dashed lines indicate average usage, dashed-dotted lines show optimal allocations, and solid lines represent polynomial trend fits. Results are based on 10 independent runs in a $4 \times 4$ grid with two agents, each run consisting of 500 episodes with 100 steps per episode.}
    \label{fig:fair}
    \vspace{-5mm}
\end{figure}

\section{Why \method is Important?}
\label{appendix:why}

In this work, we propose a method named \method, which introduces a novel fusion between reinforcement learning and a recently developed class of logics known as hyperproperties. 
Hyperproperties have gained increasing attention in the formal methods community due to their expressive power in capturing system-level behaviors that span across multiple execution traces.
We observe that many objectives and constraints in MARL naturally align with the expressive capabilities of hyperproperties. 
Motivated by this insight, we present a framework that first formally specifies the objectives and constraints of a given RL problem using hyperproperties and then learns policies that aim to maximize the probability of satisfying these specifications.

We believe that our approach provides a solid theoretical foundation for a new direction in reinforcement learning, where users can formally express their high-level requirements using hyperproperties and then synthesize policies that satisfy them. 
Importantly, hyperproperties are not limited to specifying objectives of individual agents or pairwise dependencies. Recent extensions such as HyperATL~\cite{lmcs:9209} allow reasoning about team-level objectives and inter-team relational dependencies, further expanding the scope of formal specification in MARL.



\end{document}